\documentclass[lettersize,journal]{IEEEtran}
\usepackage{amsmath,amsfonts}
\usepackage{algorithmic}
\usepackage{array}
\usepackage[caption=false,font=normalsize,labelfont=sf,textfont=sf]{subfig}
\usepackage{textcomp}
\usepackage{stfloats}
\usepackage{url}
\usepackage{verbatim}
\usepackage{graphicx}
\def\BibTeX{{\rm B\kern-.05em{\sc i\kern-.025em b}\kern-.08em
    T\kern-.1667em\lower.7ex\hbox{E}\kern-.125emX}}
\usepackage{balance}
\usepackage{cite}
\usepackage{booktabs}
\usepackage{color}
\usepackage{url}
\usepackage{bbding}
\usepackage{multirow}
\usepackage{makecell}
\usepackage{newfloat}
\usepackage{listings}
\usepackage{bm}
\usepackage{caption}
\usepackage[dvipsnames]{xcolor}
\usepackage{xcolor,colortbl}
\usepackage{tabularx}
\usepackage[colorlinks,
            linkcolor=blue,
            anchorcolor=blue,
            citecolor=blue]{hyperref}

\colorlet{tabfirst}{Green!25}
\definecolor{tabthird}{rgb}{1, 0.85, 0.7}
\definecolor{tabsecond}{rgb}{1, 0.96, 0.7}

\colorlet{colorFst}{Green!25}       
\colorlet{colorSnd}{SpringGreen!45} 
\colorlet{colorTrd}{Yellow!30}      
\newcommand{\fs}{\cellcolor{colorFst}\bf}   
\newcommand{\nd}{\cellcolor{colorSnd}}      
\newcommand{\rd}{\cellcolor{colorTrd}}      

\begin{document}
\title{NeB-SLAM: Neural Blocks-based Salable RGB-D SLAM for Unknown Scenes}
\author{Lizhi Bai, Chunqi Tian*, Jun Yang, Siyu Zhang, Weijian Liang
  \thanks{Lizhi Bai, Chunqi Tian, Jun Yang, Siyu Zhang and Weijian Liang are with Department
    of Computer Science and Technology, Tongji University, Shanghai, 201804, China
    (e-mail: \{bailizhi, junyang, tianchunqi, 2010149, liangweijian\}@tongji.edu.cn).}
  \thanks{* Corresponding author}}

\maketitle

\begin{abstract}
  Neural implicit representations have recently demonstrated considerable potential
  in the field of visual simultaneous localization and mapping (SLAM). This is due
  to their inherent advantages, including low storage overhead and representation
  continuity. However, these methods necessitate the size of the scene as input,
  which is impractical for unknown scenes. Consequently, we propose NeB-SLAM,
  a neural block-based scalable RGB-D SLAM for unknown scenes.
  Specifically, we first propose a divide-and-conquer mapping strategy that
  represents the entire unknown scene as a set of sub-maps. These sub-maps are
  a set of neural blocks of fixed size. Then, we introduce an adaptive map
  growth strategy to achieve adaptive allocation of neural blocks during
  camera tracking and gradually cover the whole unknown scene. Furthermore,
  the cumulative drift is corrected through global loop closure detection and global
  Bundle Adjustment.
  Finally, extensive evaluations on various datasets demonstrate that our method
  is competitive in both mapping and tracking when targeting unknown environments.

\end{abstract}

\begin{IEEEkeywords}
  Neural RGB-D SLAM, dense mapping.
\end{IEEEkeywords}

\section{Introduction}
\IEEEPARstart{D}{ense} visual simultaneous localization and mapping (SLAM) is an essential
technology in 3D computer vision for a number of applications in autonomous driving,
robotics, mixed reality, and other fields. Considering the complexity of real world
application scenarios, dense vision SLAM is expected to reconstruct high-quality 3D
dense scenes while maintaining real-time performance, which can
be scaled to unknown environments while sensing invisible regions.

As a branch of visual SLAM, RGB-D visual SLAM technology has been fully developed over
the past decade since the pioneering work of KinectFusion\cite{izadi2011kinectfusion,
  newcombe2011kinectfusion}. Traditional RGB-D SLAM methods
\cite{izadi2011kinectfusion,newcombe2011kinectfusion,schops2019bad,whelan2015elasticfusion
  ,thomas2012kintinuous} have the advantage of maintaining an efficient computational
cost while being able to estimate the camera pose with high accuracy and robustness
in a wide range of large-scale unknown scenarios. However, these approaches are
incapable of making reasonable geometric inference for invisible regions, so that
the reconstructed 3D maps have some empty regions, and in addition, dense maps for
large scenes require high storage costs.

Recently, neural implicit 3D scene representation or reconstruction has received a lot of
attentions\cite{mescheder2019occupancy,peng2020convolutional,sun2021neuralrecon,
  murez2020atlas}, especially the advent
of neural radiance fields (NeRF)\cite{mildenhall2021nerf} has brought 3D scene
representation to a new level of excitement. NeRF represents a continuous scene as
an end-to-end learnable neural network.
Specifically, NeRF characterizes the 3D space using a 5D vector-value (3D position and 2D
view direction)
function and fits it through a compact multilayer perceptron (MLP) to map the
corresponding volume densities and colors, optimizing the 3D scene representation by
minimizing the rendering error of the network.
Neural networks come with an inference capability that
can, to some extent, fill in the unobserved areas in 3D space, while having
an inherent advantage in terms of storage requirements.

NeRF's strong 3D spatial representation capability is applied to the field of dense
vision SLAM for the first time by iMAP\cite{2021imap}. For small-sized
rooms, this method demonstrates strong tracking and mapping performance.
However, when scaling up to larger scenes, using a single MLP to represent
the entire scene is clearly limited. The limited number of individual MLP
parameters can lead to catastrophic forgetting of the observed region. This results in
a significant degradation in the approach's tracking and mapping performance.
In light of this, some methods \cite{2022nice,rosinol2023nerf,wang2023co,
  zhu2023nicer,li2023dense} have attempted to use multi-resolution grid features
to obtain a more detailed representation of the scene. This allows for an extension
of their approach to larger scenes. All of these methods have a restriction
that the size of the scene must be known in order to normalize a bounding box.
Therefore, these methods are only applicable in known scenes, not in unknown ones.

\begin{figure*}[t]
  \centering
  \includegraphics[scale=0.62]{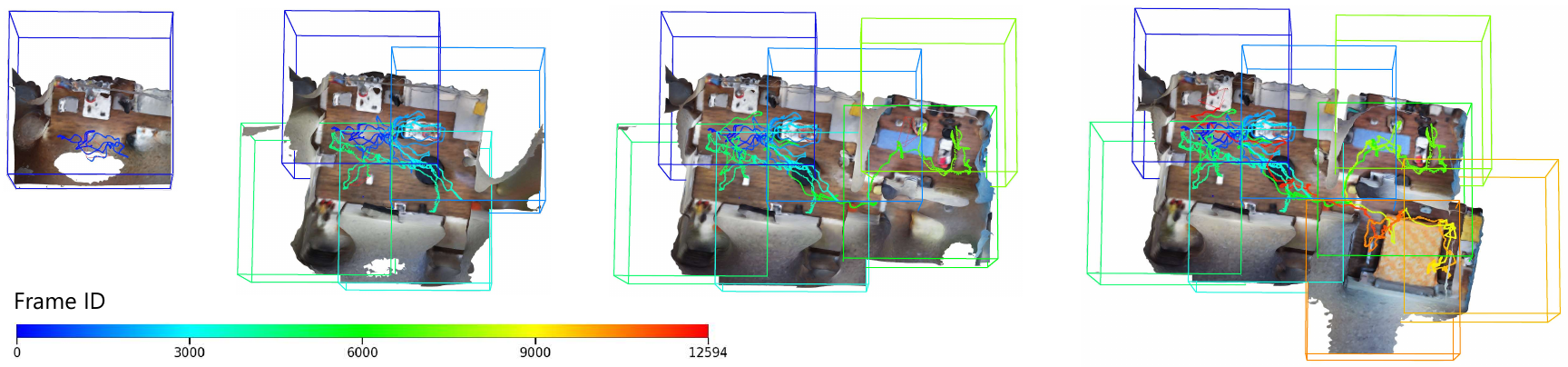}
  \caption{The divide-and-conquer mapping process for unknown scenes.
    NeBs are adaptively allocated with camera tracking to gradually cover
    the entire unknown scene.
  }
  \label{mapping}
\end{figure*}

To realize neural dense visual SLAM that is scalable to unknown scenes, we
propose NeB-SLAM, an end-to-end neural RGB-D visual SLAM system. Our approach
revolves around a divide-and-conquer mapping strategy and an adaptive growth
strategy for mapping as illustrated in Fig. \ref{mapping}. To analyze an unfamiliar scene, we begin by initializing
a neural block (NeB), which is a cube with a fixed size, based on the current camera
pose. Our method then adaptively allocates new NeBs as RGB-D image
sequences are input, eventually partitioning the entire unknown environment
into multiple submaps, also known as NeBs.
In order to identify the global position of each input frame, we employ the
use of the Bag of Words (BoW)\cite{salinas2017dbow3} for the purpose of online
detection of loop closure.
Upon the occurrence of a loop closure, we proceed to undertake a global Bundle
Adjustment (BA)
in order to rectify the cumulative drift that has occurred between the trajectories
of the closed loops.
In our method, each
NeB represents the local scene as a multi-resolution hash
grid\cite{muller2022instant}, which allows for high convergence speed and
the representation of high-frequency local features. Inspired by
Co-SLAM\cite{wang2023co}, each NeB also encodes the coordinates
using One-blob\cite{muller2019neural} to encourage surface coherence.
We conducted a comprehensive assessment of
numerous indoor RGB-D sequences and showcased the scalability and predictive
capabilities of our approach in unfamiliar settings. In summary, our
contributions are as follows:

$\bullet$ We propose NeB-SLAM, an end-to-end dense RGB-D SLAM system
for unknown scenarios that is real-time, scalable, and predictive.

$\bullet$ For unknown scenarios, we propose a divide-and-conquer mapping
strategy and an adaptive map growth strategy to achieve full coverage
of the unknown environment by the sub-map during the tracking process.

$\bullet$ Extensive evaluations on various datasets demonstrate that
our method is competitive in both mapping and tracking when targeting
unknown environments.

\section{Related Work}
\subsection{Dense Visual SLAM}
Thanks to the pioneering work of Klein et al.\cite{klein2009parallel},
most current visual SLAM systems are organized into two parts:
tracking and mapping. DTAM\cite{newcombe2011dtam}
is an early dense SLAM system that tracks the camera using the photometric
consistency of each pixel. It employs multi-view stereo constraints
to update the dense scene model and represent it as a cost-volume.
KinectFusion\cite{izadi2011kinectfusion} takes advantage of RGB-D cameras
to enable real-time camera pose estimation and scene geometry
updating with iterative closest point (ICP) and TSDF-Fusion\cite{curless1996volumetric}.
Many subsequent studies have suggested various efficient data structures
to enable the scalability of SLAM systems, such as VoxelHash\cite{
  niessner2013real,kahler2015very,chen2013scalable} and
Octrees\cite{zeng2013octree,vespa2018efficient}.

With the development of
deep learning, some works have integrated it into SLAM systems to
enhance the robustness and accuracy of conventional methods.
DeepTAM\cite{zhou2018deeptam} is similar to DTAM\cite{newcombe2011dtam},
but it uses convolutional neural networks (CNN) to estimate camera pose
increments and depth maps. Comparable approaches are
Demon\cite{ummenhofer2017demon} and DeepV2D\cite{teed2018deepv2d}.
CodeSLAM\cite{bloesch2018codeslam} employs a variational
auto-encoders\cite{kingma2013auto} to achieve the latent compact
representation of scene geometry, reducing the complexity
of the problem. There are methods, such as SceneCode\cite{zhi2019scenecode}
and NodeSLAM \cite{sucar2020nodeslam}, that optimize the potential
features by decoding them into depth maps.
BA-Net\cite{tang2018ba} and DeepFactors\cite{czarnowski2020deepfactors}
simplify the optimization problem by using networks to generate a set of
basis depth maps and representing the resulting depth maps as a linear
combination of these basis depth maps. Droid-SLAM\cite{teed2021droid}
greatly improves the generalizability of a pre-trained model across
scenarios through the use of dense optical flow estimation\cite{teed2020raft}
and dense bundle adjustment, allowing for competitive
results in a variety of challenging datasets. Tandem\cite{koestler2022tandem}
realizes
a real-time monocular dense SLAM system by performing frame-to-model
photometric tracking to decouple the pose/depth problem using multi-view stereo
network and DSO\cite{engel2017direct}.

In contrast to these approaches, ours is an end-to-end method that
represents the geometric information of the scene as a set of neural
blocks with efficient memory usage and reasonable hole-filling.

\begin{figure*}
  \centering
  \centering
  \centerline{
    \includegraphics[scale=0.62]{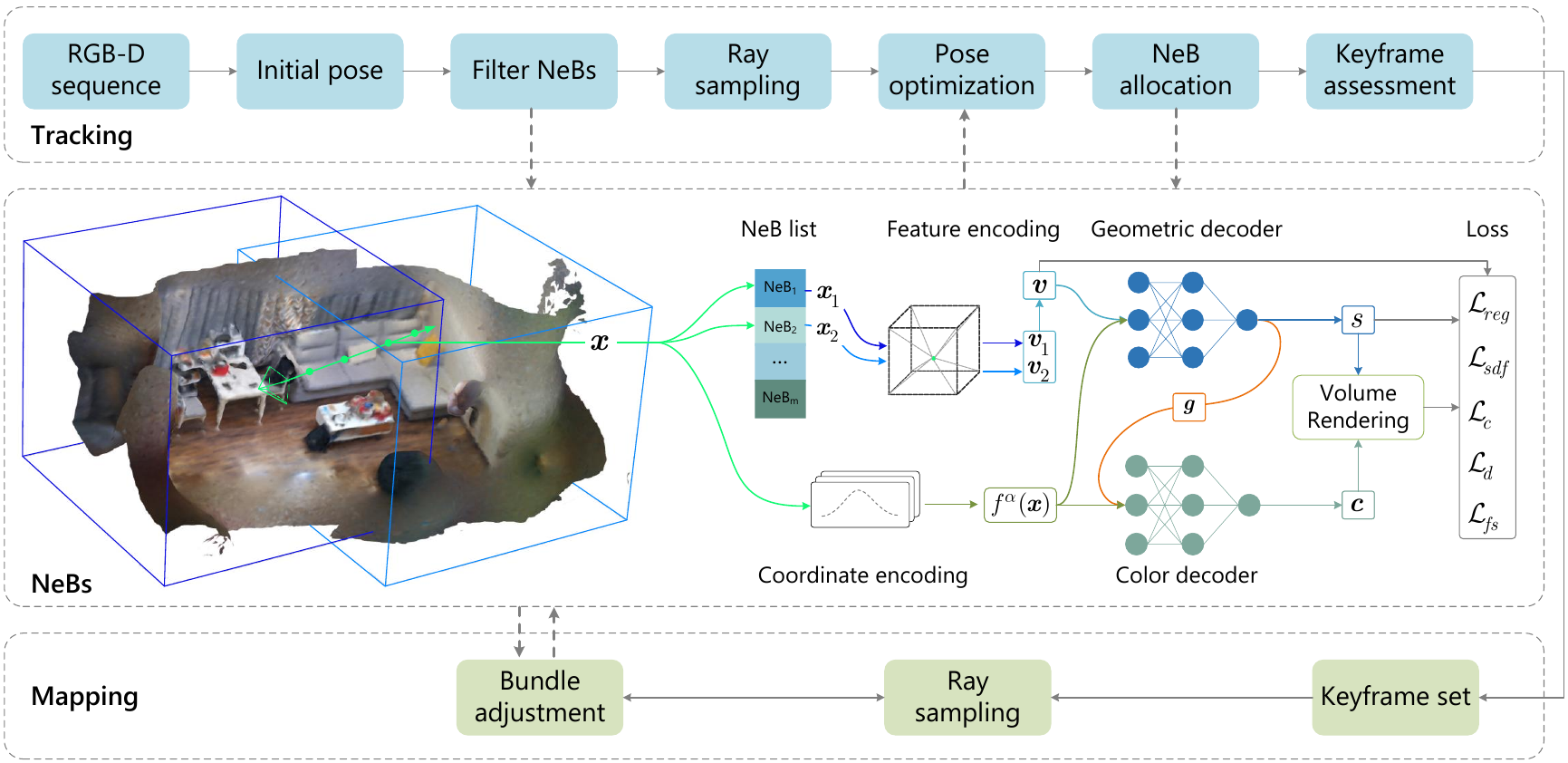}
  }
  \caption{Overview of NeB-SLAM. The scene is represented using
    NeBs, each of which has independent local coordinates and feature
    encoding. For any 3D point, feature encoding and coordinate encoding
    are uesd to estimate the sdf value and color by two compact
    MLPs. Tracking process optimizes the camera pose for each frame, and
    mapping process jointly optimizes the scene representation and the
    poses of all keyframes.}
  \label{overview}
\end{figure*}

\subsection{Neural Implicit Representations}
Recently neural implicit representations have gained significant
attention, which initially encode the geometric and appearance
information of a 3D scene in the parameters of a neural network,
and are characterized by high expressiveness and compactness.
Among these works, NeRF\cite{mildenhall2021nerf} is a valuable
geometric representation for capturing view-dependent accurate
photometric synthesis  while maintaining multi-view consistency.
It has inspired numerous papers that aim to improve 3D reconstruction
and reduce training time. NeRF-W\cite{martin2021nerf} proposes a
method for synthesizing new views of complex scenes using images
captured under natural conditions as datasets. This is achieved
by introducing appearance embedding and static-dynamic scene separation.
Block-NeRF\cite{tancik2022block} chunks the environment and represents
large-scale scenes with multiple NeRFs. By modeling these NeRFs independently,
it improves its ability to represent large scenes and its training
speed. To improve training efficiency, NGLOD\cite{takikawa2021neural}
uses a hierarchical data structure that concatenates features
from each layer to achieve a scene representation at different
levels of detail. A similar approach is NSVF\cite{liu2020neural},
which constructs sparse voxel meshes with geometric features.
Point-NeRF\cite{xu2022point} combines point-cloud and NeRF for
fast convergence and rendering with generalization. For any
3D location, the neural points in its neighborhood are aggregated
using MLP to regress the volume density and view-dependent radiation.
Plenoxels\cite{fridovich2022plenoxels} exploits spherical harmonics
to parameterize the directional coding, bypassing the use of MLP
to improve speed. Instant-NGP\cite{muller2022instant} demonstrates
that neural radiation fields can be trained in real-time using a
hash-based hierarchical volumetric representation of the scene.
Several studies\cite{oechsle2021unisurf,wang2021neus,yariv2021volume,
  yu2022monosdf,azinovic2022neural} have also proposed replacing the density
field with a signed distance function or other representation to
improve 3D reconstruction.

Similar to Co-SLAM\cite{wang2023co}, our approach employs one-blob
to encode spatial coordinates on top of a multi-resolution hash grid\cite{muller2022instant}
representation of the scene to achieve superior model representativeness
and ensemble consistency.

\subsection{Neural Implicit SLAM}
Decoupling the dependence on known camera poses has become another
research topic in the area of neural radiation fields. This
is particularly tempting for NeRF because the acquisition of the
poses of the images usually requires additional preprocessing,
which is usually done with COLMAP\cite{schonberger2016structure}.

iNeRF\cite{2021inerf} was the first system to show that it is possible
to regress the camera pose in the case of a trained NeRF for a given scene.
In actuality, it is not considered a full slam system, but rather a
localization problem under an existing model. Barf\cite{2021barf}
further showed how to fit the NeRF while estimating the camera pose
given an inaccurate initial guess by establishing a theoretical
connection from classical image alignment to joint alignment and
reconstruction with neural radiance field. To be precise, the method
solves the structure from motion (SfM) problem. The aforementioned
methods choose large MLPs as map representations and are therefore
too slow for online inference.

iMAP\cite{2021imap} demonstrates the application of NeRF in
reconstructing precise 3D geometry from RGB-D images without
poses for the first time. iMAP directly uses a single
MLP to approximate a global scene and jointly optimizes the map
and the camera poses. However, the use of a single MLP makes it
difficult to represent geometric details of the scene as well as
scale to larger environments without significantly increasing the
network capacity. NICE-SLAM\cite{2022nice} proposes to tackle the scalability
problem by subdividing the world coordinate frame into uniform grids
in order to make inference faster and more accurate.
NeRF-SLAM\cite{rosinol2023nerf} combines Droid-SLAM\cite{teed2021droid}
with Instant-NGP\cite{muller2022instant}, using Droid-SLAM to
estimate camera poses, dense depth maps and their uncertainties,
and using the above information to optimize the Instant-NGP scene
representations. The GO-SLAM\cite{zhang2023go} improves global
consistency in scene reconstruction by introducing loop closure
and global bundle adjustment. Co-SLAM\cite{wang2023co} combines
the advantages of coordinates and sparse grid encoding to achieve
high quality reconstruction of the scene.

All of the previous methods require the bounding box of the
scene in order to normalize a feature space. In contrast,
our approach is specifically designed for unknown scenes.
This means that it can still work even if the spatial
dimension information of the scene is not known.

\section{Method}
Given a set of RGB-D image sequences $\{I_t,D_t\}_{t=1}^N$
($I$ and $D$ are the color and depeh images, respectively) as input,
our goal is to output a surface reconstruction of the scene, as
well as a trajectory of the 6 DoF camera poses
$\bm{T}_t=[\bm{R}_t|\bm{t}_t]$ ($[\bm{R}|\bm{t}]\in SE(3)$).
Fig. \ref{overview} presents an overview of our approach. The map is composed of
NeBs, each of which is a fixed-size cube defined in a
local coordinate frame with spatial features
encoded by a multi-resolution hash grid\cite{muller2022instant}.
These NeBs are adaptively allocated and progressively cover the
entire unknown scene as the camera tracking.

Similar to other SLAM systems, our approach consists of two distinct
processes: a tracking process that estimates the current camera pose
and a mapping process that optimizes the global map.
At system startup, the global map is initialized through a few mapping
iterations for the first frame. For subsequent frames, the tracking process
uses the NeBs corresponding to the current frame and the differential
volume rendering method to estimate its pose. It determines for each frame
whether it is a keyframe or not and decides if a new NeB should be
allocated or not. The mapping process receives new key frames and optimizes
the map (NeBs) globally.

\subsection{Neural Blocks}
\label{neural-blocks}
We represent the unknown scene with a set of NeBs $\{B_m\}_{m=1}^M$
that are adaptively allocated along the camera trajectory. Each NeB
$B_m=(\bm{C}_m,\mathcal{F}_m,f^\alpha,f^\beta_m,f^\gamma,f^\delta)$ is a
multi-resolution hash grid defined in its respective local coordinate frame.
Here, $\bm{C}_m\in\mathbb{R}^3$ refers to the center coordinate of each neural
block as defined in the world coordinate frame, and $\mathcal{F}_m$ denotes
the sequence of keyframes corresponding to each NeB, with each
keyframe including its pose $\bm{T}_t$ in the world coordinate frame, as
well as a color image and a depth image. $f^\beta_m$ represents the
multi-resolution hash encoding\cite{muller2022instant} for the corresponding
NeB that encodes any local coordinate as a feature vector.
Given a 3D point $\bm{x}\in \mathbb{R}^3$ in the world frame, the encoding
feature vector in $m^{th}$ NeB is
$\bm{v}_m=\{\bm{v}_{ml}\}_{l=1}^L=f^\beta_m(\bm{x}_m)$,
where $\bm{x}_m$ refers to the local coordinate of
$\bm{x}$ in the coordinate frame of $B_m$.
The encoding feature vector $\bm{v}_{ml}$ in the level $l$
with resolution $R_l$ is defined as:
\begin{equation}
  \begin{split}
    &\bm{v}_{ml}=\sum_{i = 1}^{8}w(\bm{x}_{ml}^i)  h(\bm{x}_{ml}^i),\\
    &\lfloor \bm{x}_{ml}\rfloor:=\lfloor \bm{x}_m R_l \rfloor, \lceil \bm{x}_{ml}\rceil:=\lceil\bm{x}_m R_l  \rceil,\\
    &R_l:=\lfloor R_{min} b^l\rfloor, b:=\exp{(\frac{\ln{R_{max}}-\ln{R_{min}}}{L-1})},
  \end{split}
  \label{encoding}
\end{equation}
where $R_{min}$ and $R_{max}$ are the minimum and maximum resolution of
the hash grid, $L$ is the number of levels. $\bm{x}_{ml}^i$ denotes the
neighboring grid point around $\bm{x}_{ml}$ for trilinear interpolation,
and $w(\bm{x}_{ml}^i)$ is the corresponding weight.
$h$ represents the hash function\cite{muller2022instant,teschner2003optimized}
to retrieve the feature vector at $\bm{x}_{ml}^i$. Similar to \cite{wang2023co},
Spatial coordinates in world frame are encoded using One-blob encoding
$f^\alpha $ for coherence and smoothness reconstruction. With the encoded
features above, the geometric decoder $f^\gamma$ predicts the SDF value
$s_m$ and the feature vector $\bm{g}_m$ at $\bm{x}$:
\begin{equation}
  f^\gamma(f^\alpha(\bm{x}),f_m^\beta(\bm{x}_m))\mapsto (s_m,\bm{g}_m).
  \label{geo_dec}
\end{equation}
Then, the color MLP $f^\delta$ predicts the RGB value $\bm{c_m}$:
\begin{equation}
  f^\delta(f^\alpha(\bm{x}),\bm{g}_m)\mapsto \bm{c}_m.
  \label{col_dec}
\end{equation}
Here, the parameters in $f_m^\beta$, $f^\gamma$ and $f^\delta$ are learnable.

\subsection{Rendering for Neural Blocks}
\label{rendering}
Similar to other methods\cite{2021imap,2022nice,wang2023co}, the depth and
color maps are obtained through differentiable volume rendering, which
integrates the SDFs and colors obtained in Sec. \ref{neural-blocks}. Specifically,
Given the camera intrinsic parameters $\bm{K}$ and camera pose
$\bm{T}=[\bm{R}|\bm{t}]$, the ray origin $\bm{o}$ and direction $\bm{r}$
corresponding to each pixel coordinate $[u,v]$ can be obtained:
\begin{equation}
  \begin{split}
    &\bm{o}=\bm{t},\\
    &\bm{r}=\bm{RK}^{-1}[u,v,1]^\top.
  \end{split}
  \label{ray}
\end{equation}
Along this ray, we sample $N_1$ points uniformly between the $near$ and
$far$ bound of the viewing frustum.
Additionally, we further sample $N_2$ points near the surface uniformly
for rays with valid depth values. Thus, a total of $N_p=N_1+N_2$ points
are sampled on each ray. These sampling points can be written as
$\bm{x}_i=\bm{o}+d_i\bm{r}, i\in \{1,\ldots ,N_p \}$, and $d_i$ corresponds
to the depth value of $\bm{x}_i$ along this ray. For each point $\bm{x}_i$ on the ray,
the SDF and color values can be calculated using Eq. (\ref{geo_dec})
and Eq. (\ref{col_dec}), and the corresponding depth and color values
of the ray can be obtained by volume rendering:
\begin{equation}
  \begin{split}
    &\hat{d}=\frac{1}{\sum_{i = 1}^{N_p}w_i} \sum_{i = 1}^{N_p}w_id_i,\\
    &\hat{\bm{c}}=\frac{1}{\sum_{i = 1}^{N_p}w_i} \sum_{i = 1}^{N_p}w_i \bm{c}_i,\\
    &w_i=\sigma(\frac{s_i}{tr})\sigma(-\frac{s_i}{tr}),
  \end{split}
  \label{vol-ren}
\end{equation}
where $\{w_i\}_{i=1}^{N_p}$ are the weights of the corresponding depths
$\{d_i\}_{i=1}^{N_p}$ along the ray and $tr$ is the
truncation distance. Following \cite{azinovic2022neural}, we multiply
the two Sigmoid functions $\sigma(\cdot)$ to compute the weights $w_i$.

As previously stated, a set of adaptively allocated NeBs is
employed to represent the unknown scene. As illustrated in and Fig. \ref{overview},
with respect to a given point $\bm{x}$ on a ray, there may be more than one
NeB $\{B_m\}_{m=1}^M$ to which it is assigned, or there may be only one.
In the former case, the feature vectors are extracted from the NeBs
in separate instances and the mean value is taken as the output of the point
$\bm{v}_{\bm{x}}=\frac{1}{M}\sum_{m = 1}^{M} f_m^\beta(\bm{x}_m)$, where
$\bm{x}_m=\bm{x}-\bm{C}_m$ is the local coordinate in the corresponding NeB.
For the latter, the feature vector of the point is computed in the corresponding
NeB $\bm{v}_x=f_m^\beta(\bm{x}_m)$. Subsequently, Eq. (\ref{geo_dec}) and
Eq. (\ref{col_dec}) are employed to obtain the corresponding SDFs and
colors, and Eq. (\ref{vol-ren}) is used to derive the depths and colors
of the volume rendering. In the aforementioned procedure, we limit our
consideration to candidate NeBs within the viewing frustum.
Additionally, points that are within the viewing frustum but not included
in either NeB are discarded during the sampling process.

\begin{figure}
  \centering
  \centering
  \centerline{
    \includegraphics[scale=0.56]{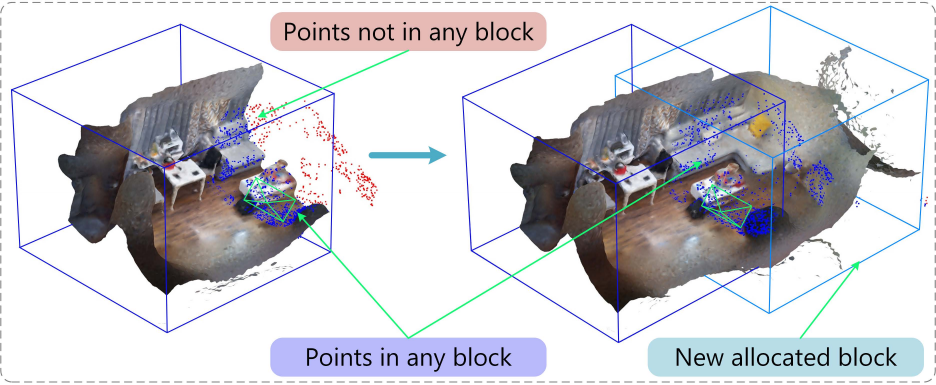}
  }
  \caption{NeB allocation. NeBs are adaptively allocated based on the
    proportion of newly observed scene to the whole scene in the current
    viewing frustum.}
  \label{neb_alloc}
\end{figure}

\subsection{Neural Block Allocation}
\label{neb-alloc}
The NeB is an axis-aligned cube of fixed size ($5\times5\times5$ m$^3$
in our experiments) and we adaptively allocate NeBs along the camera
trajectory for unknown scenes during camera tracking. For each frame with a
pose $\bm{T}$ and intrinsic parameters $\bm{K}$, we determine whether to
allocate a new NeB by a metric $\tau$ that is the ratio of the newly
observed scene to the whole scene in current viewing frustum as shown in Fig.
\ref{neb_alloc}. Specifically,
we first select a random pixels with valid depths $\{D[u_i,v_i]\}$
($[u_i,v_i]$ is the $i^{th}$ pixel coordinate) in current frame and then
back-project them into the world coordinate frame to obtain a 3D point set
$\mathcal{X}=\{\bm{x}_i\}$, here
$\bm{x}_i=D[u_i,v_i]\bm{T}\bm{K}^{-1}[u_i,v_i,1]^\top$. Then, we
compute the ratio $\tau$ as:
\begin{equation}
  \tau=\frac{\left\lvert \mathcal{X}\right\rvert-\sum_{i = 1}^{\left\lvert \mathcal{X}\right\rvert}\mathbb{I}(\bm{x}_i\in \mathcal{B})}{\left\lvert \mathcal{X}\right\rvert},
  \label{neb_allocation}
\end{equation}
where $\mathcal{B}$ denotes the set of all the 3D points in NeBs within
the current viewing frustum. $\mathbb{I}$ is an indicator function that
results in 1 if $\bm{x}_i$ belongs to the set $\mathcal{B}$ and 0 otherwise.
If $\tau$ is greater than a threshold
$\tau_{th}$, a new NeB is allocated with center $\bm{C}$:
\begin{equation}
  \bm{C}=\frac{1}{\left\lvert \mathcal{X} \setminus \mathcal{B}\right\rvert}\sum_{\bm{x}\in\mathcal{X} \setminus \mathcal{B}}\bm{x}.
  \label{neb_center}
\end{equation}

\begin{figure*}
  \centering
  \scriptsize
  \setlength{\tabcolsep}{0.5pt}
  \newcommand{\sz}{0.245}  %
  \begin{tabular}{lccccc}
                                                                                 & \tt \footnotesize{office-2} & \tt \footnotesize{office-3} & {\tt \footnotesize{room-0}} & {\tt \footnotesize{room-1}} & {\tt \footnotesize{room-2}} \\
    \makecell{\rotatebox{90}{iMAP$^*$~\cite{2021imap}}}                          &
    \makecell{\includegraphics[height=\sz\columnwidth]{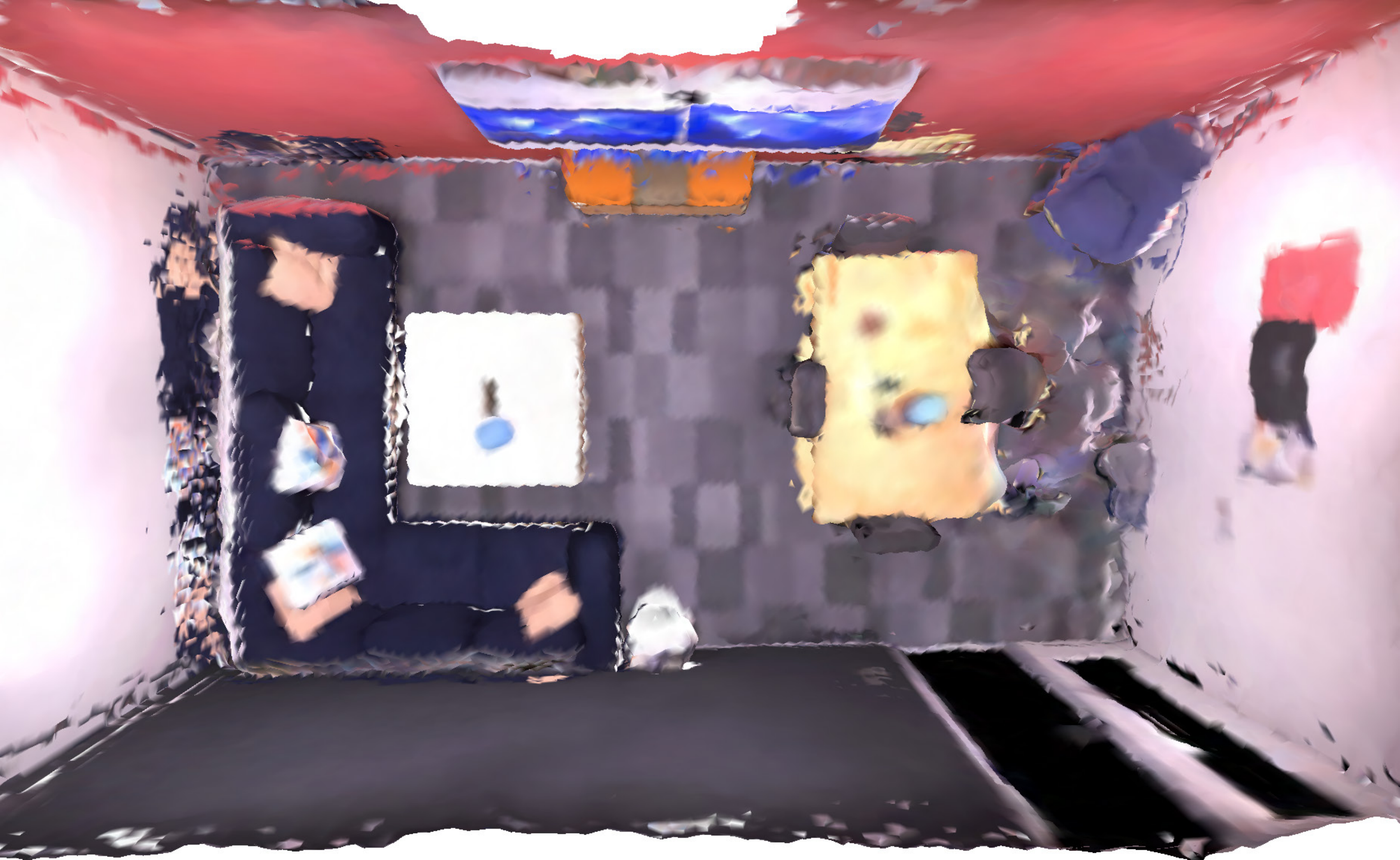}} &
    \makecell{\includegraphics[height=\sz\columnwidth]{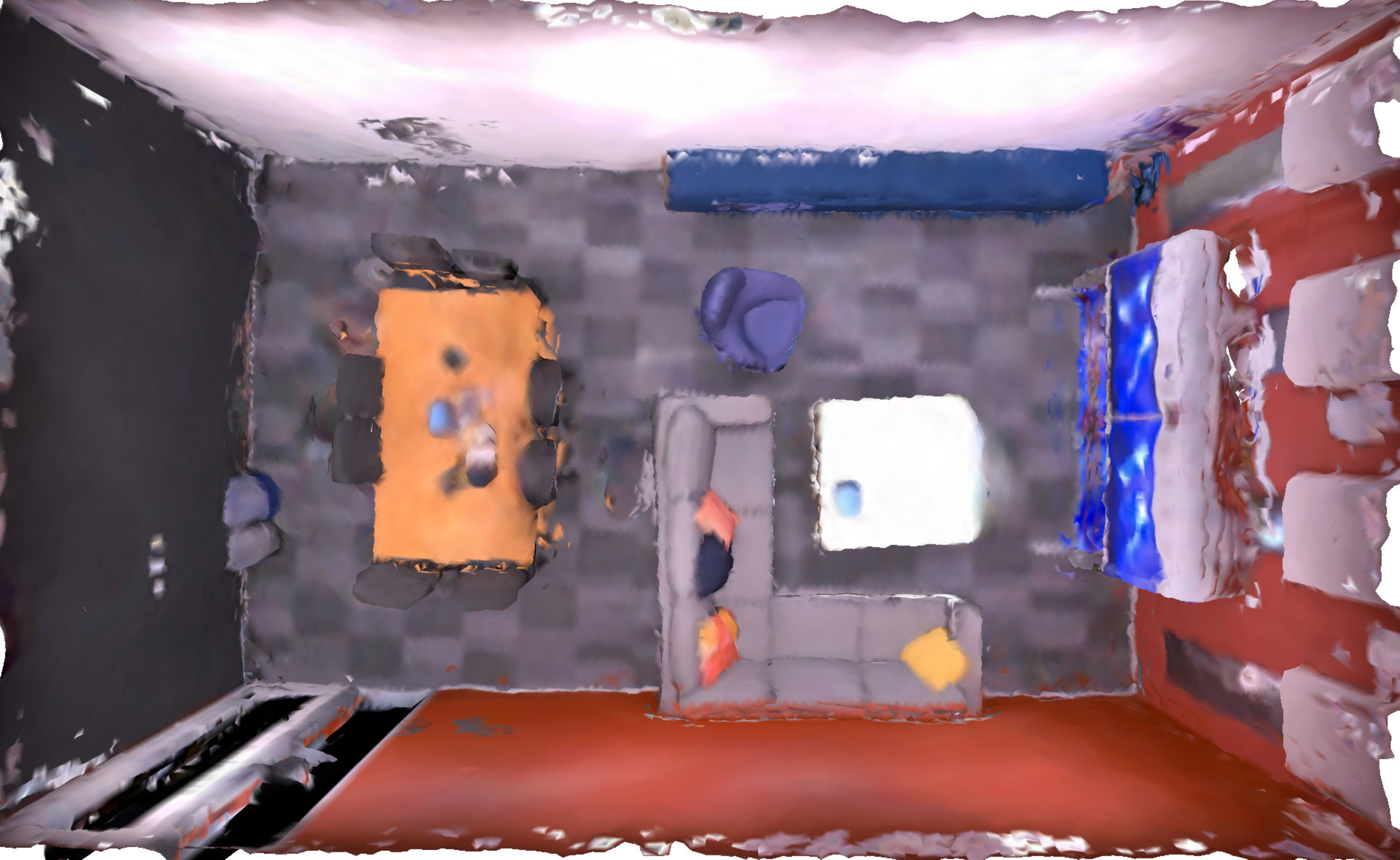}} &
    \makecell{\includegraphics[height=\sz\columnwidth]{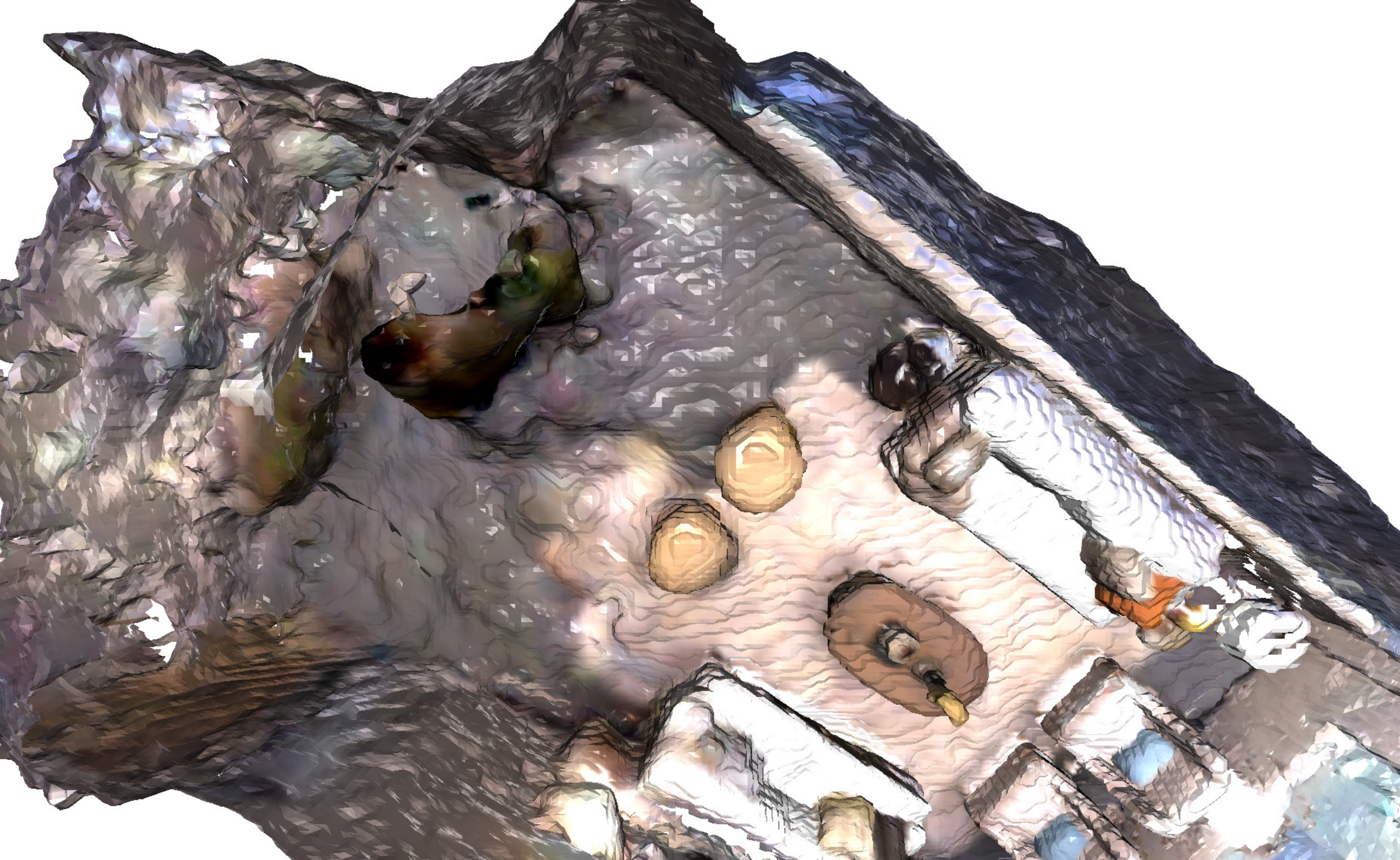}} &
    \makecell{\includegraphics[height=\sz\columnwidth]{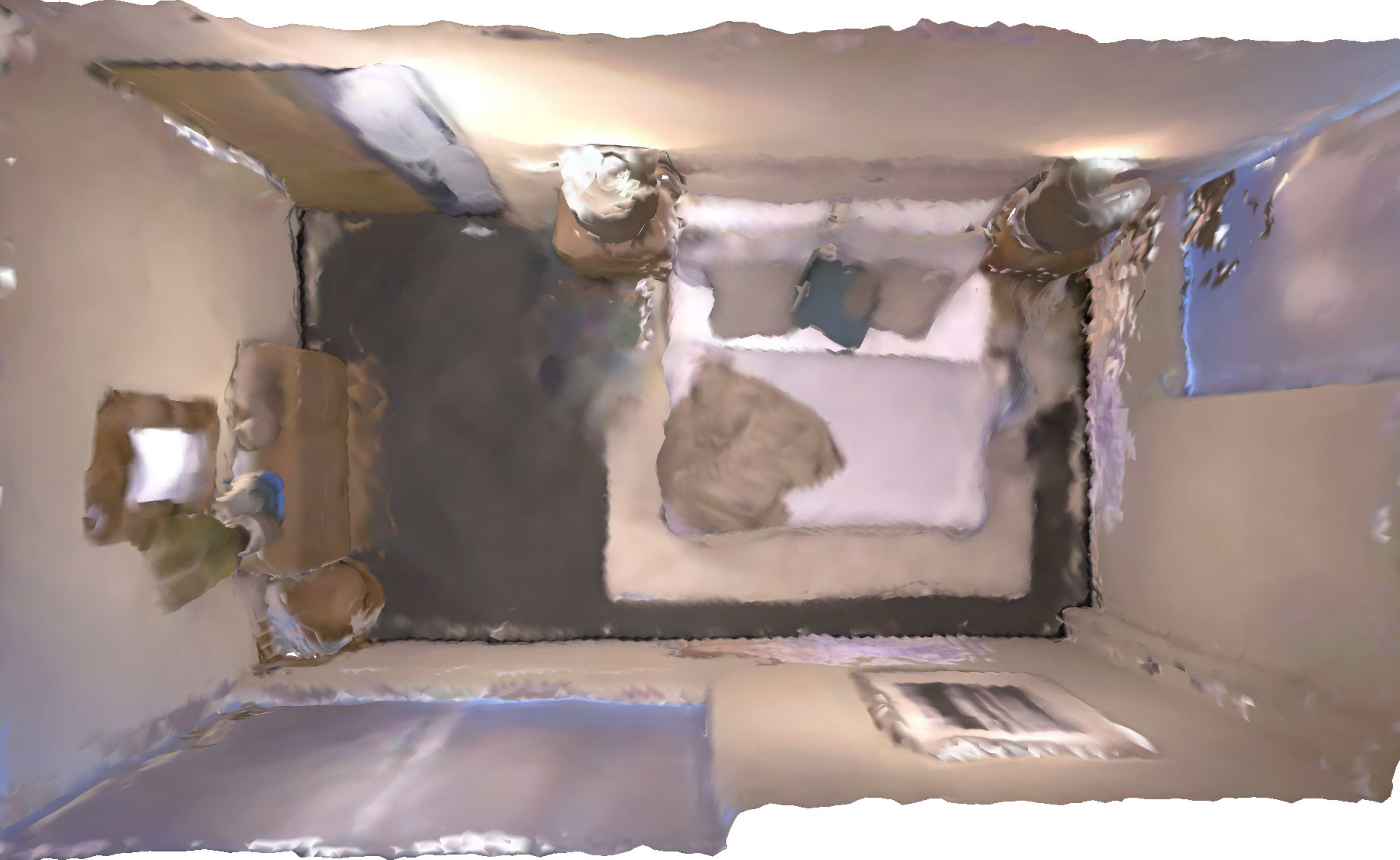}} &
    \makecell{\includegraphics[height=\sz\columnwidth]{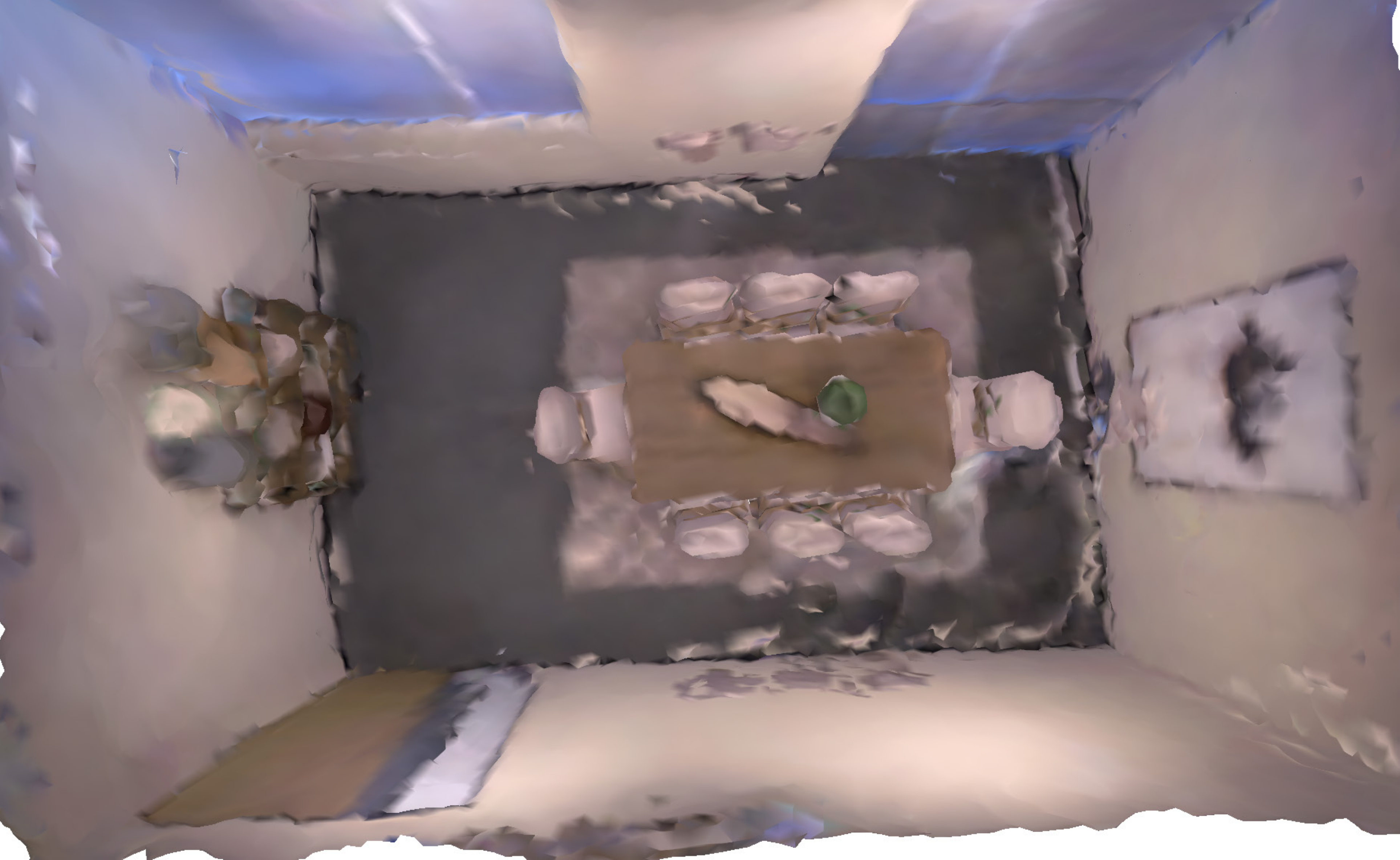}}                                                                                                                                                       \\

    \makecell{\rotatebox{90}{NICE-SLAM~\cite{2022nice}}}                         &
    \makecell{\includegraphics[height=\sz\columnwidth]{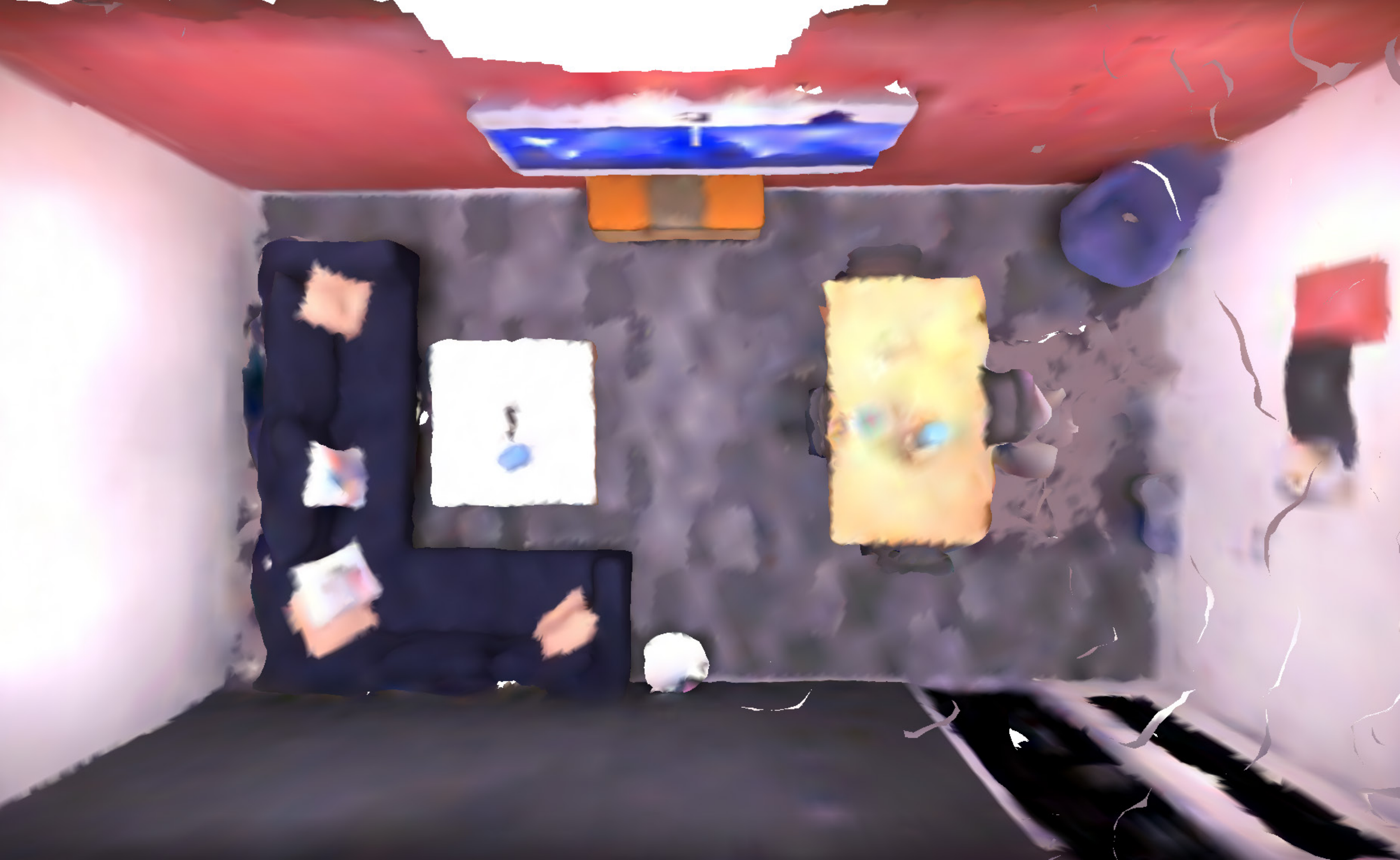}} &
    \makecell{\includegraphics[height=\sz\columnwidth]{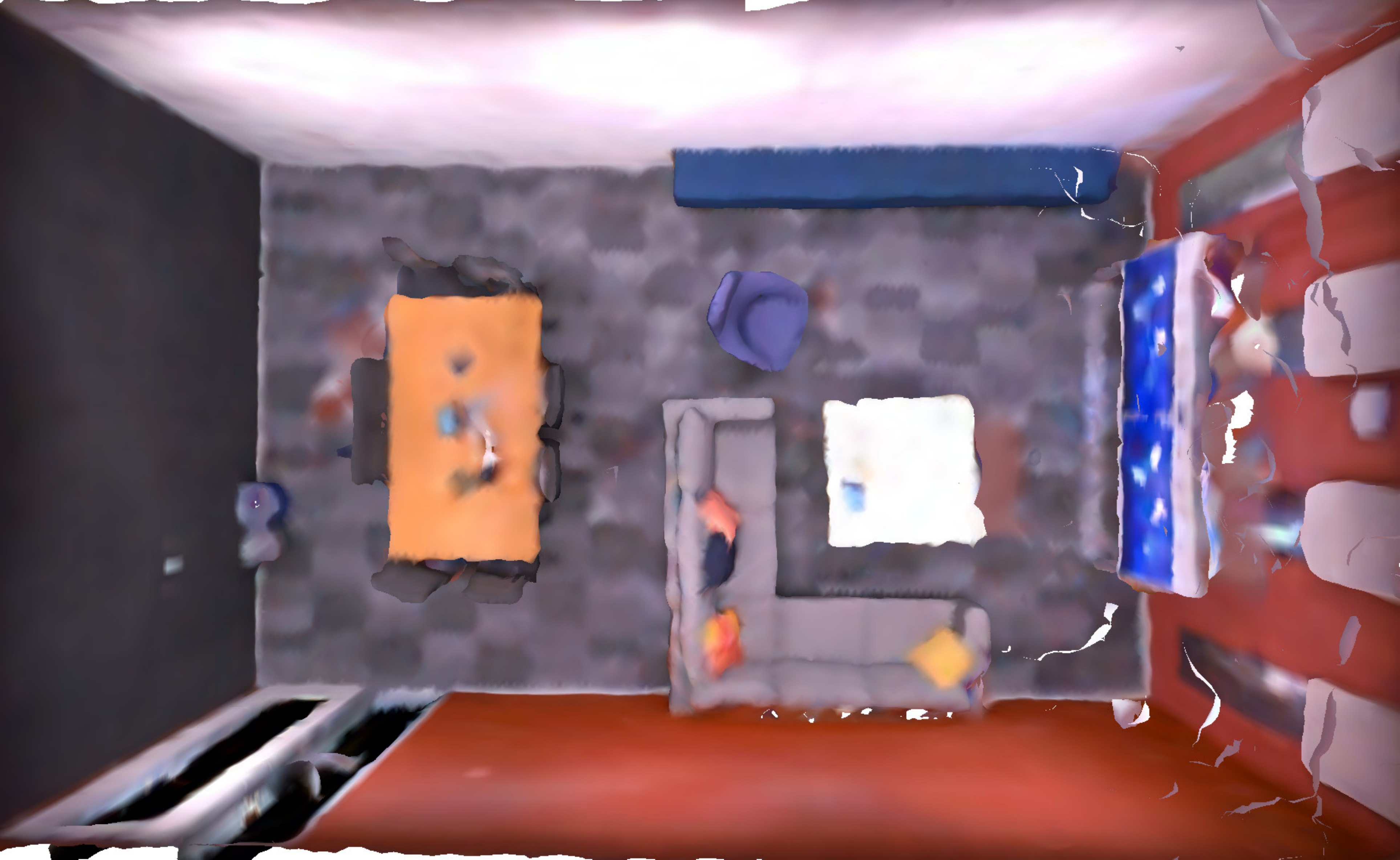}} &
    \makecell{\includegraphics[height=\sz\columnwidth]{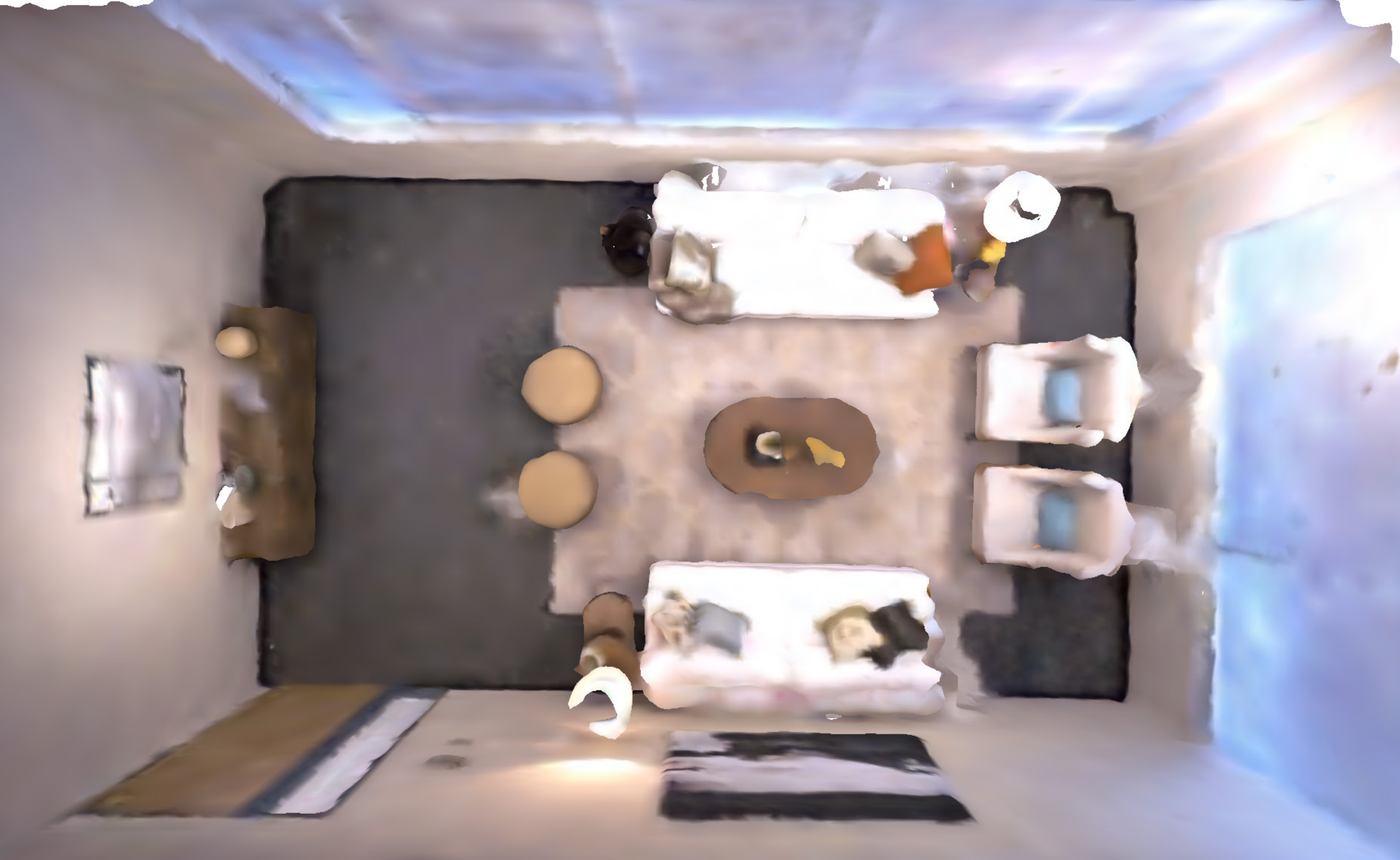}} &
    \makecell{\includegraphics[height=\sz\columnwidth]{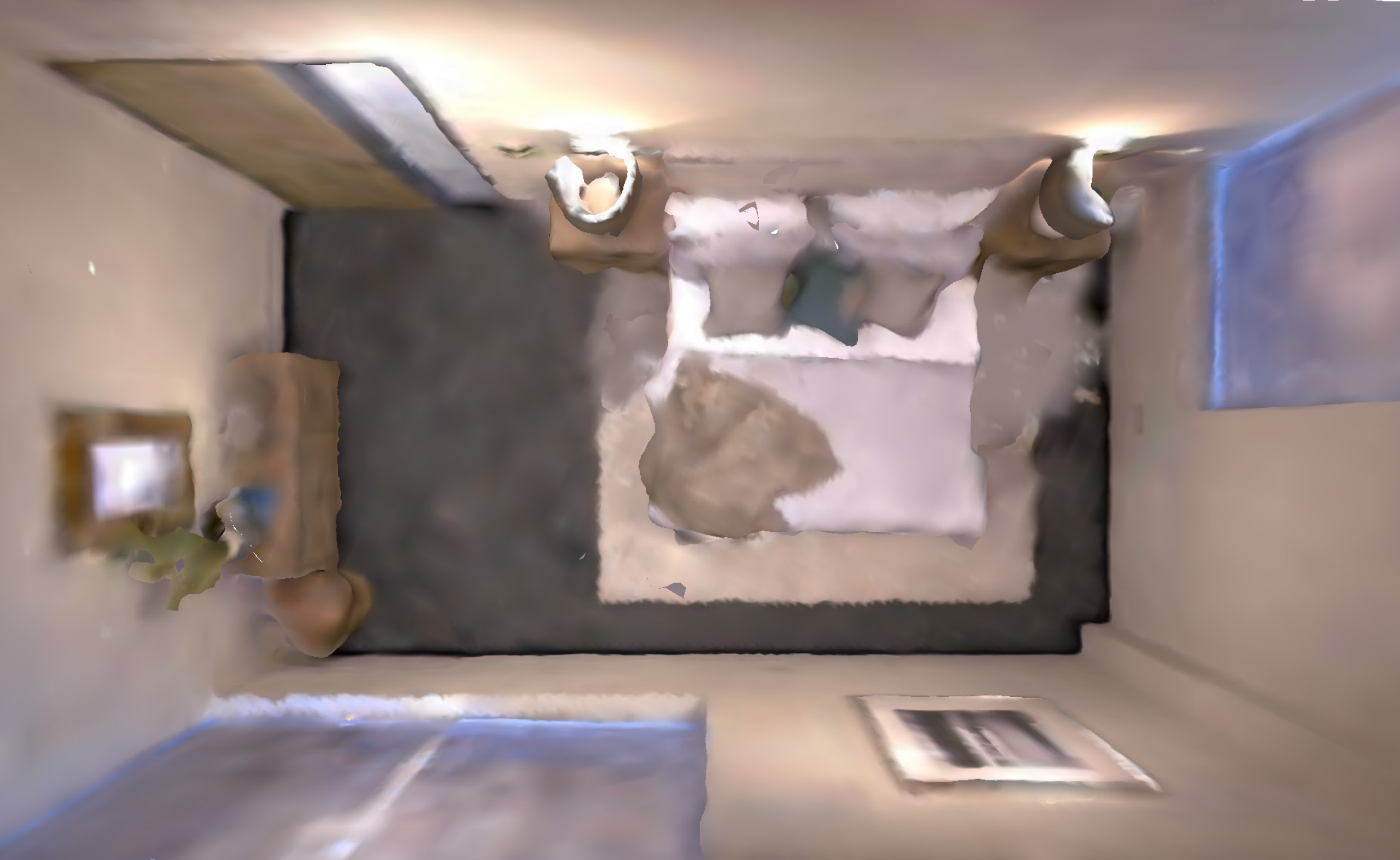}} &
    \makecell{\includegraphics[height=\sz\columnwidth]{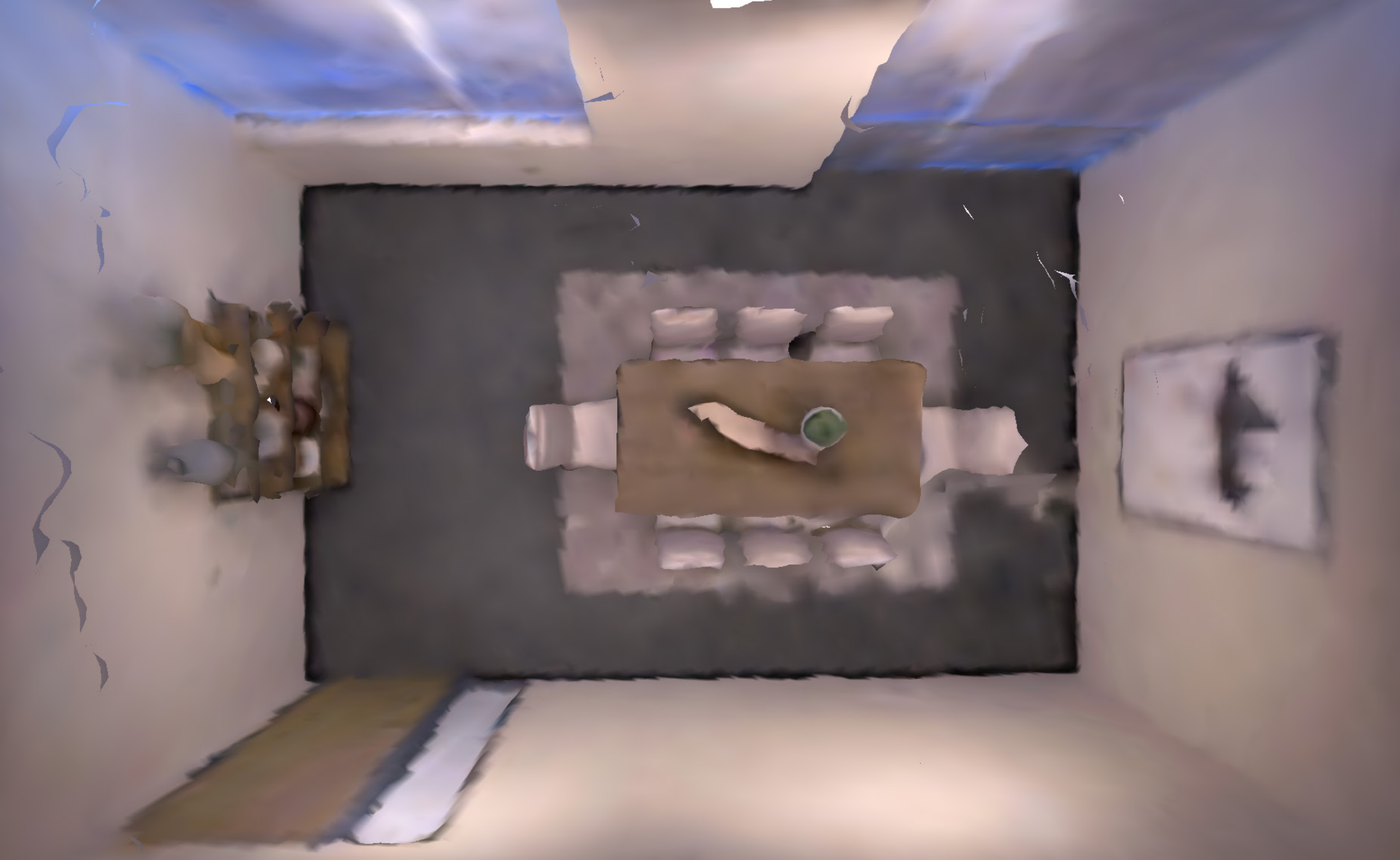}}                                                                                                                                                       \\

    \makecell{\rotatebox{90}{Co-SLAM~\cite{wang2023co}}}                         &
    \makecell{\includegraphics[height=\sz\columnwidth]{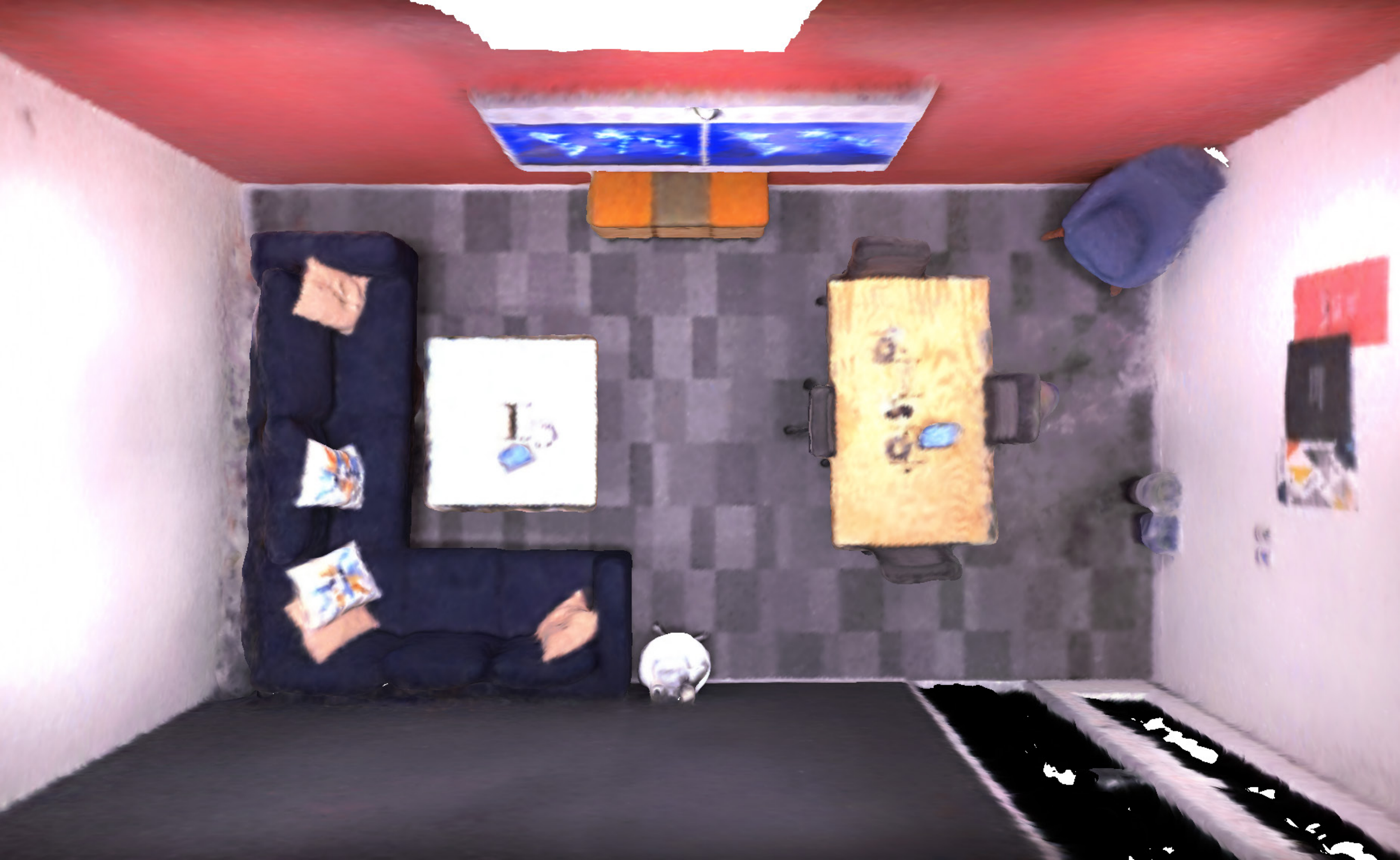}}   &
    \makecell{\includegraphics[height=\sz\columnwidth]{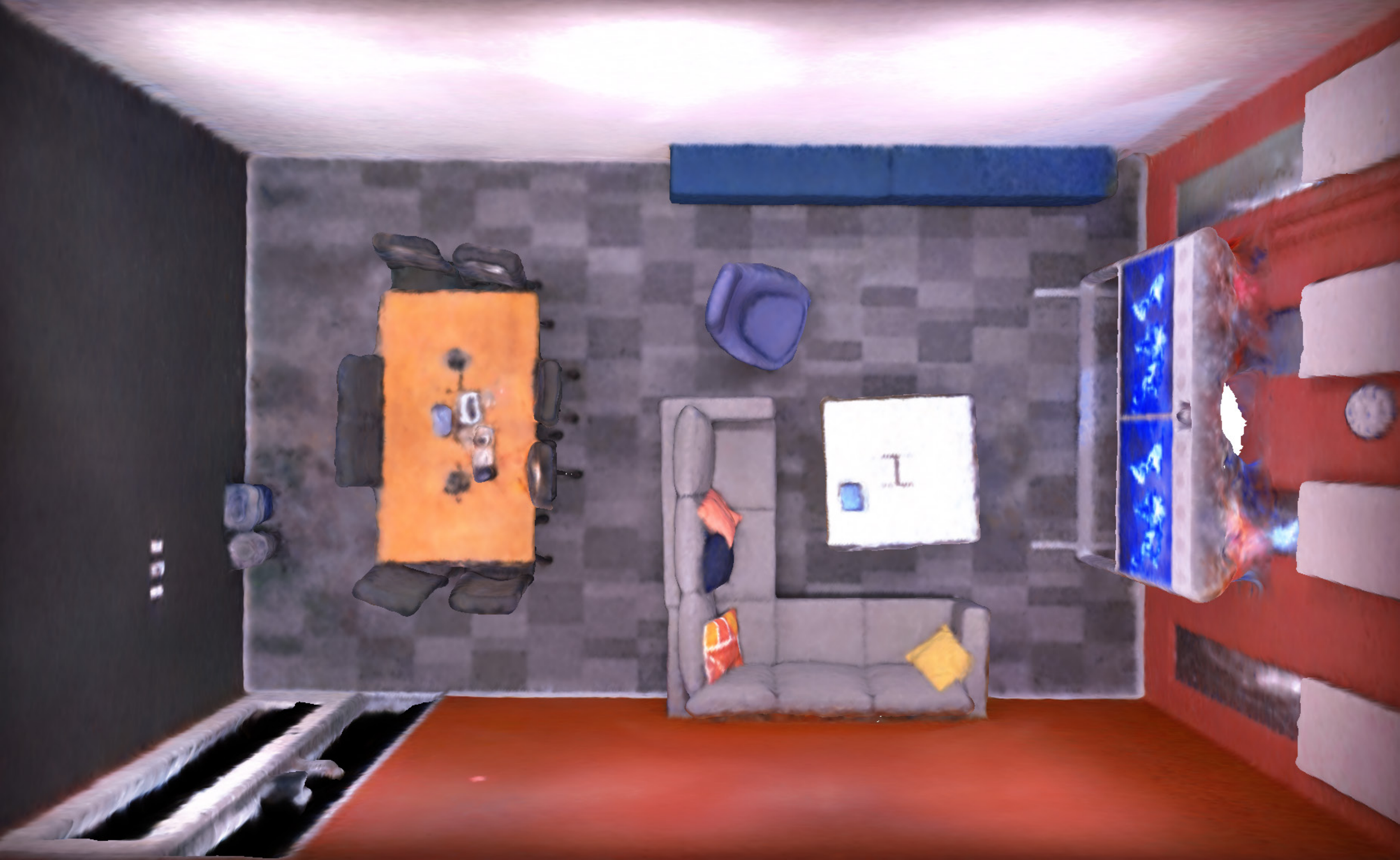}}   &
    \makecell{\includegraphics[height=\sz\columnwidth]{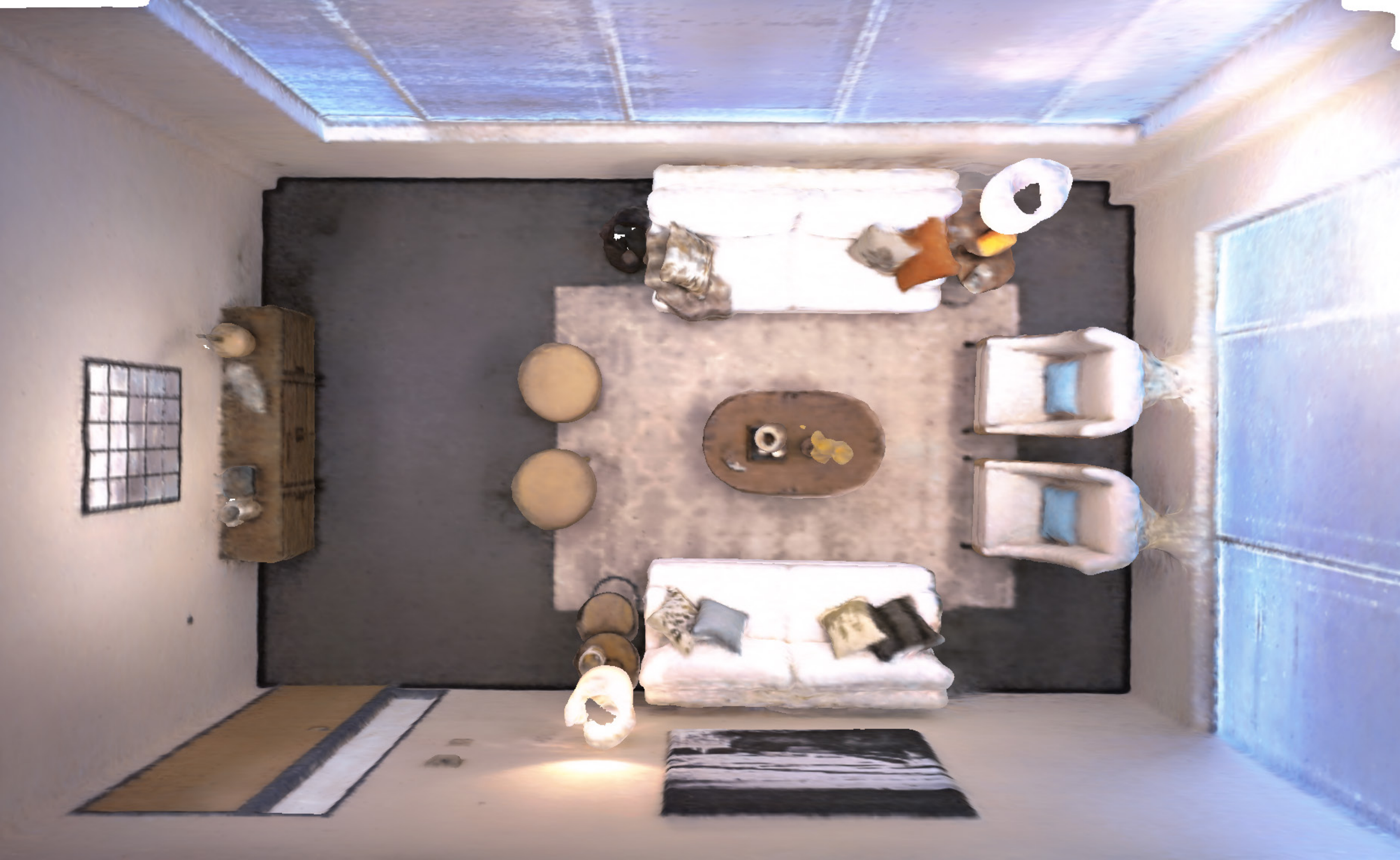}}   &
    \makecell{\includegraphics[height=\sz\columnwidth]{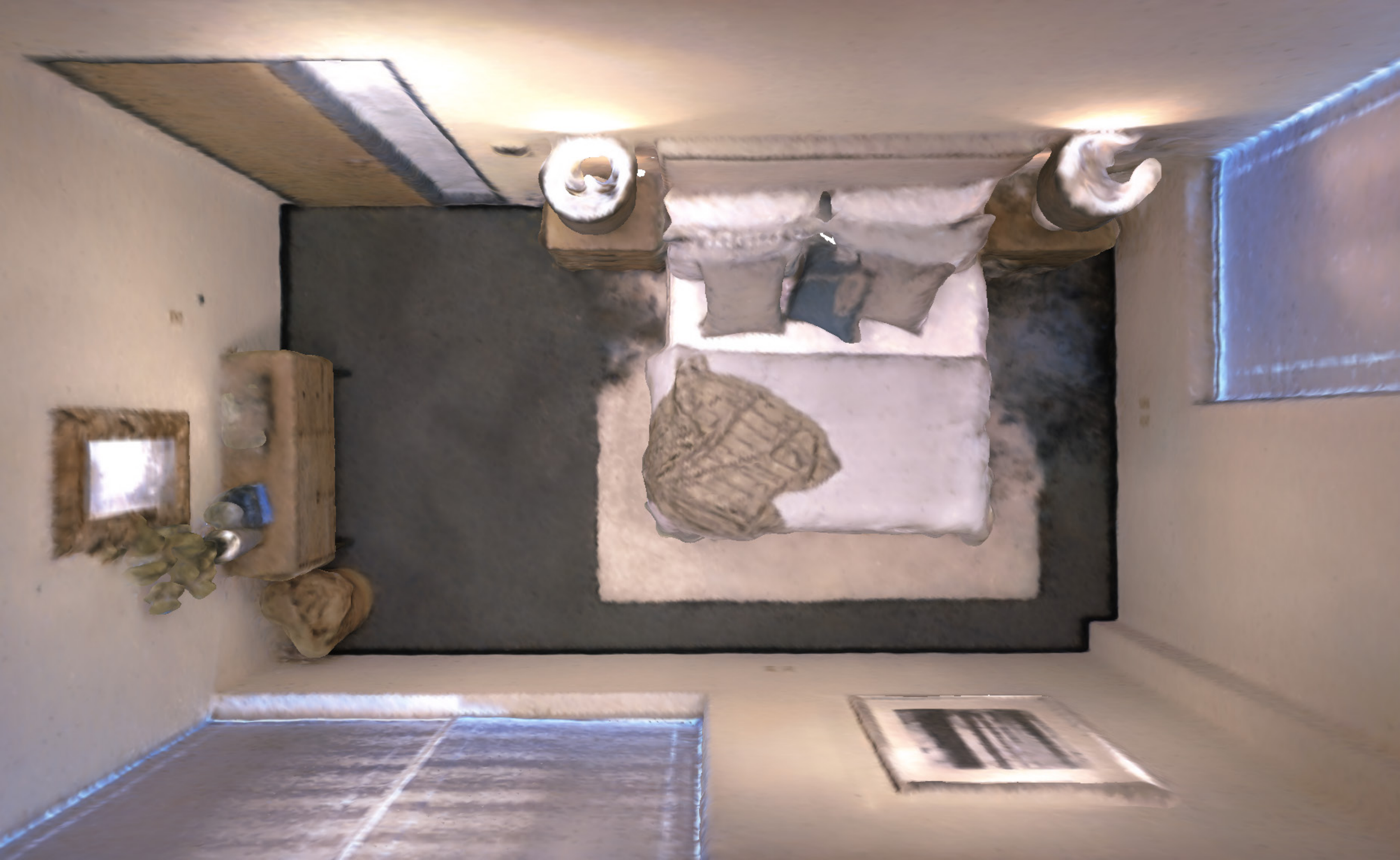}}   &
    \makecell{\includegraphics[height=\sz\columnwidth]{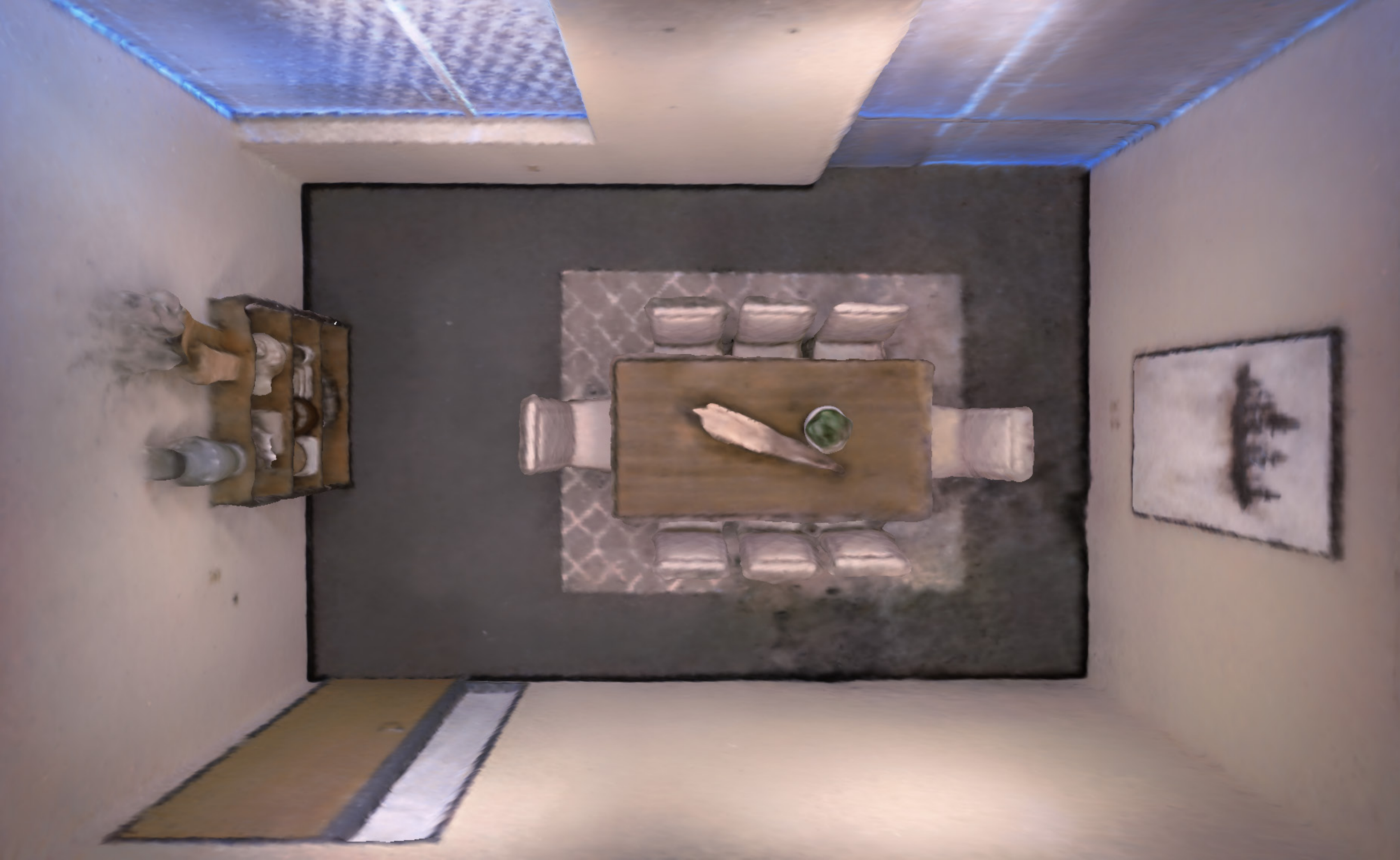}}                                                                                                                                                         \\

    \makecell{\rotatebox{90}{NeB-SLAM}}                                          &
    \makecell{\includegraphics[height=\sz\columnwidth]{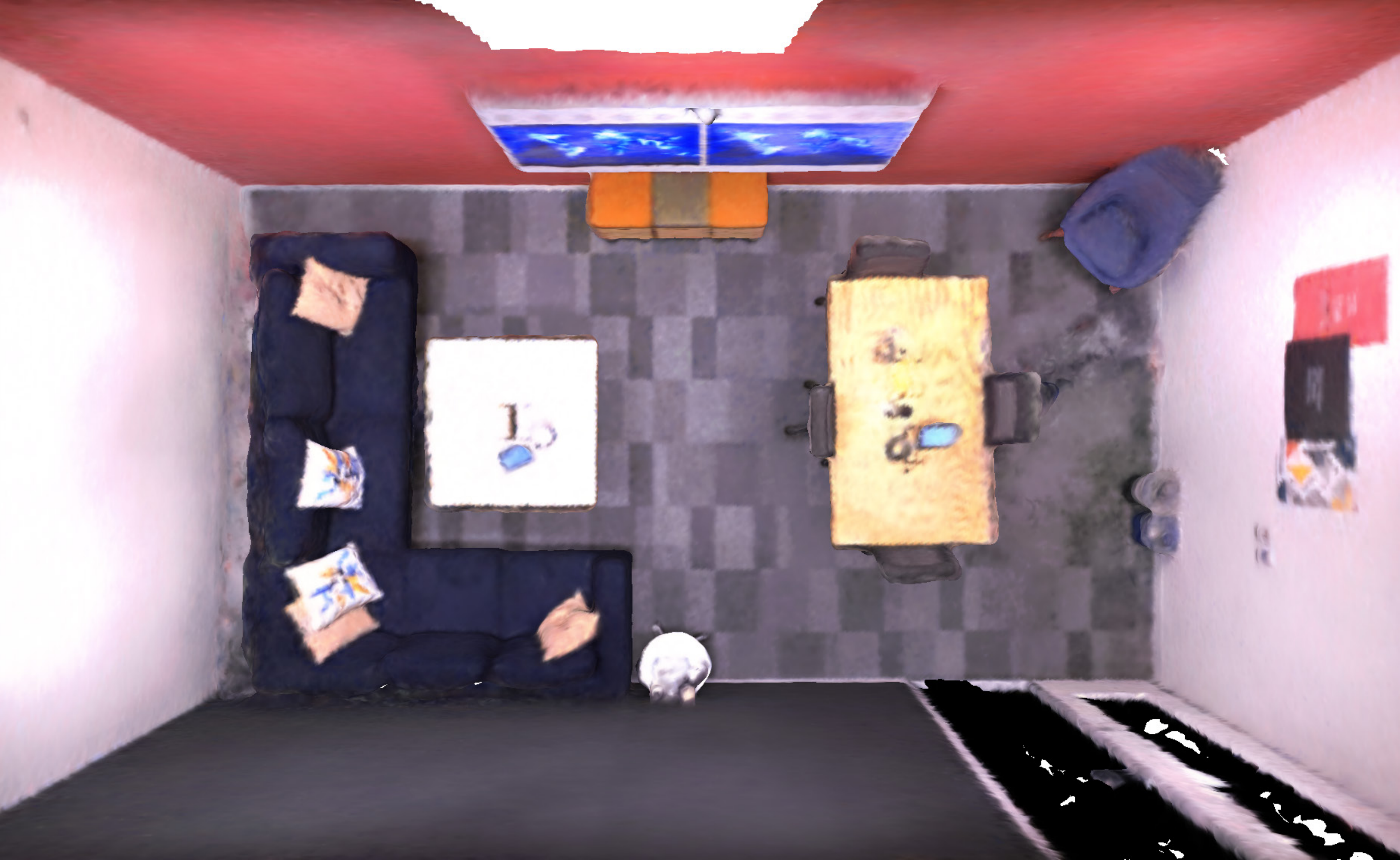}}  &
    \makecell{\includegraphics[height=\sz\columnwidth]{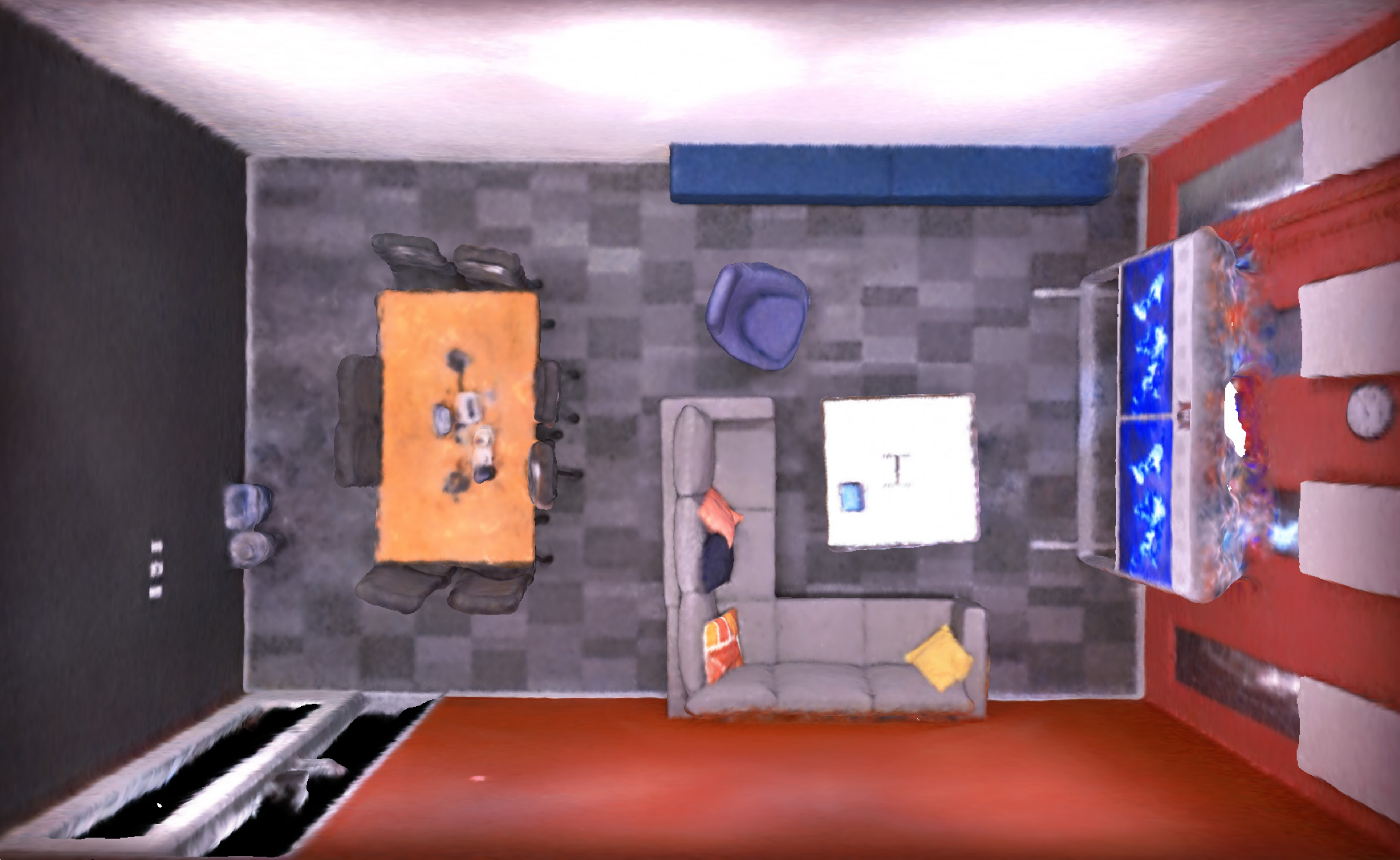}}  &
    \makecell{\includegraphics[height=\sz\columnwidth]{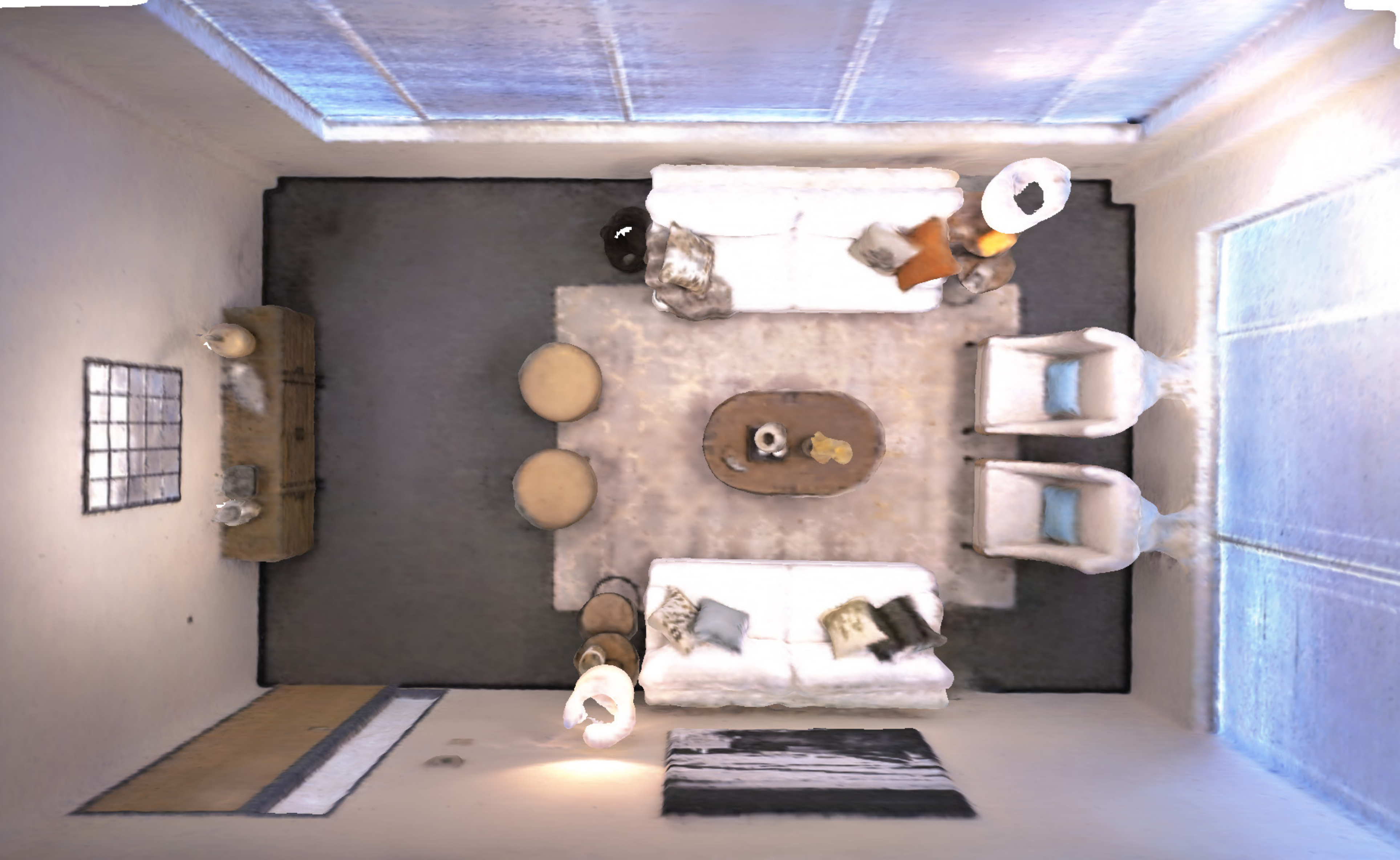}}  &
    \makecell{\includegraphics[height=\sz\columnwidth]{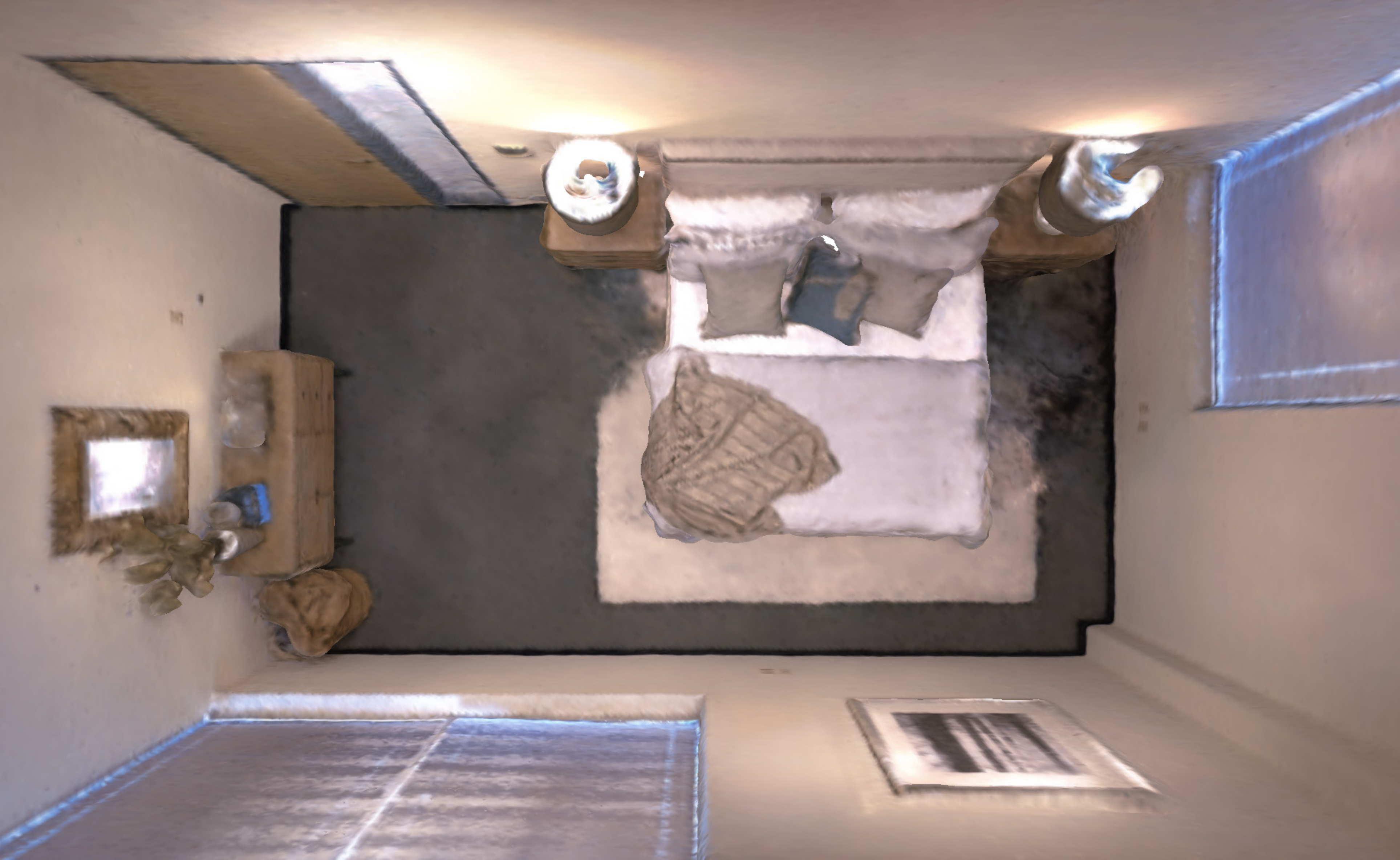}}  &
    \makecell{\includegraphics[height=\sz\columnwidth]{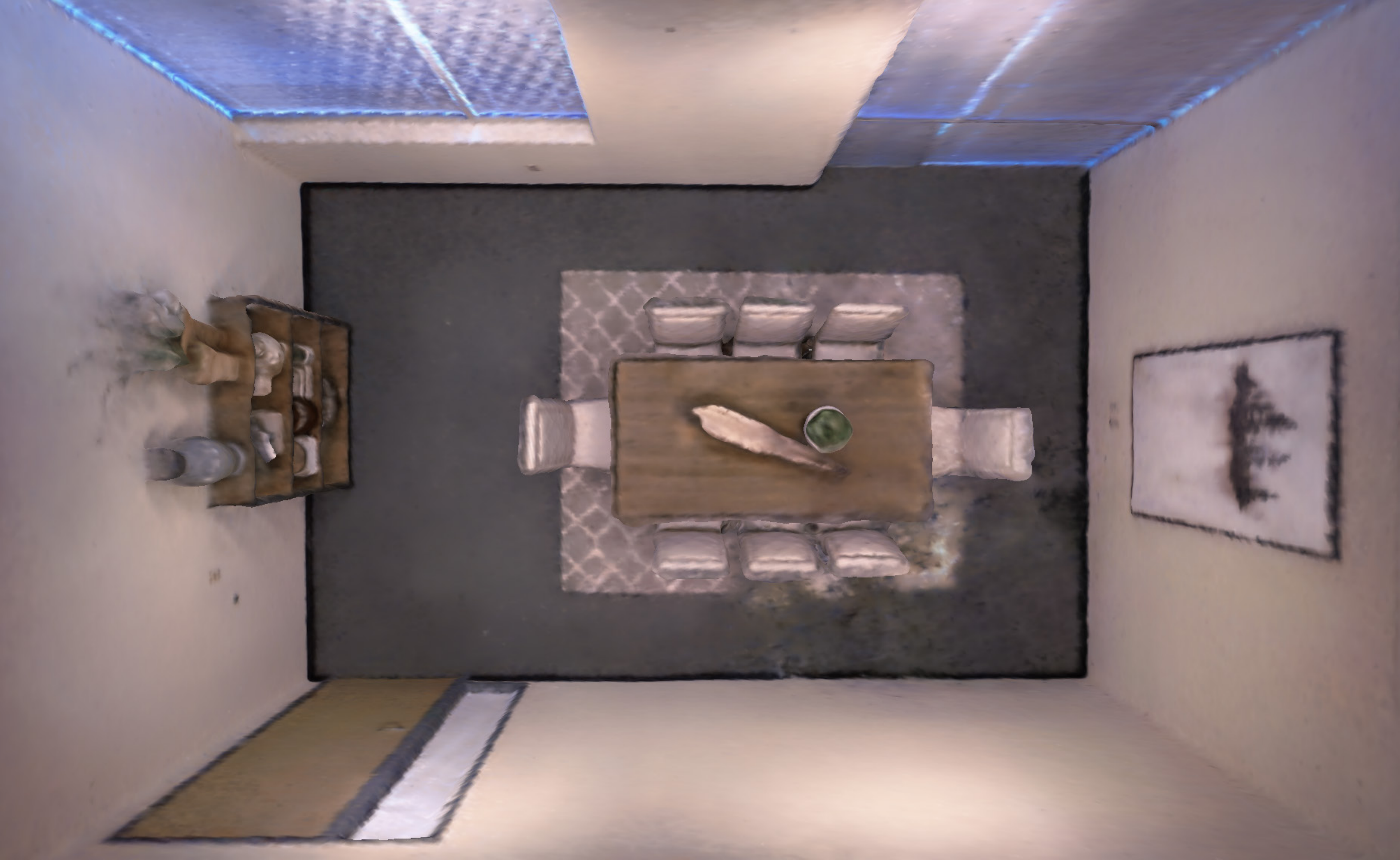}}                                                                                                                                                        \\

    \makecell{\rotatebox{90}{GT}}                                                &
    \makecell{\includegraphics[height=\sz\columnwidth]{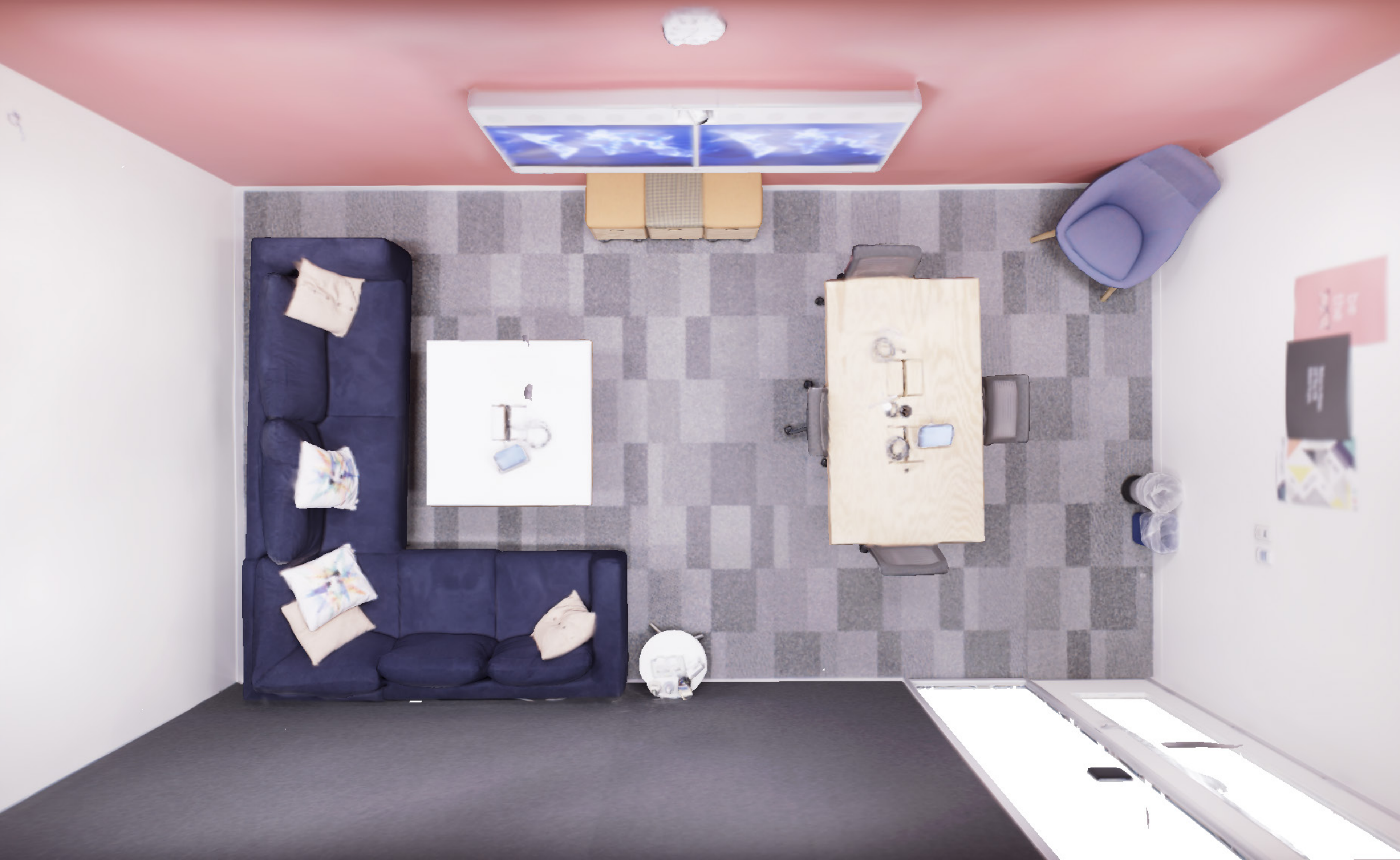}}   &
    \makecell{\includegraphics[height=\sz\columnwidth]{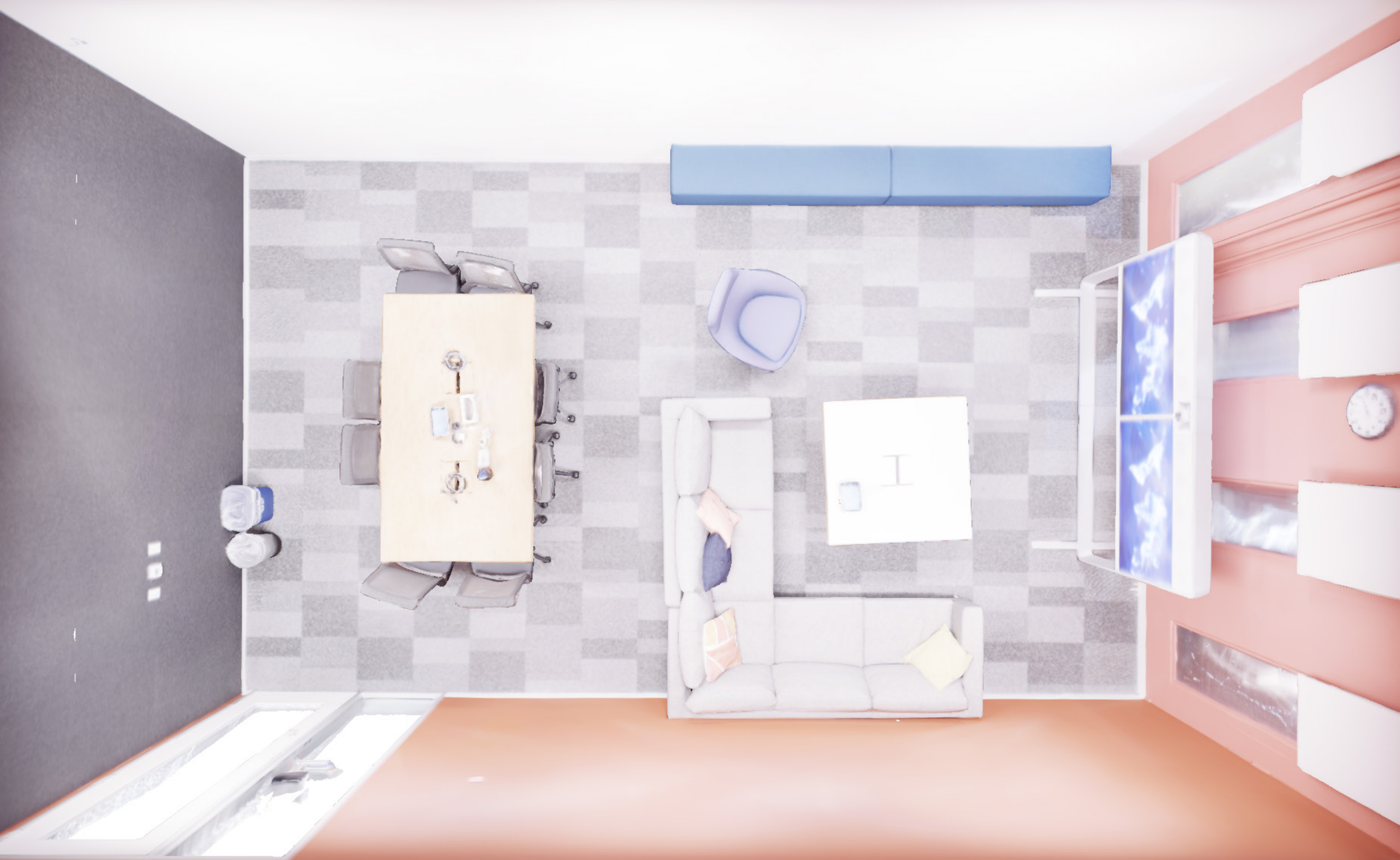}}   &
    \makecell{\includegraphics[height=\sz\columnwidth]{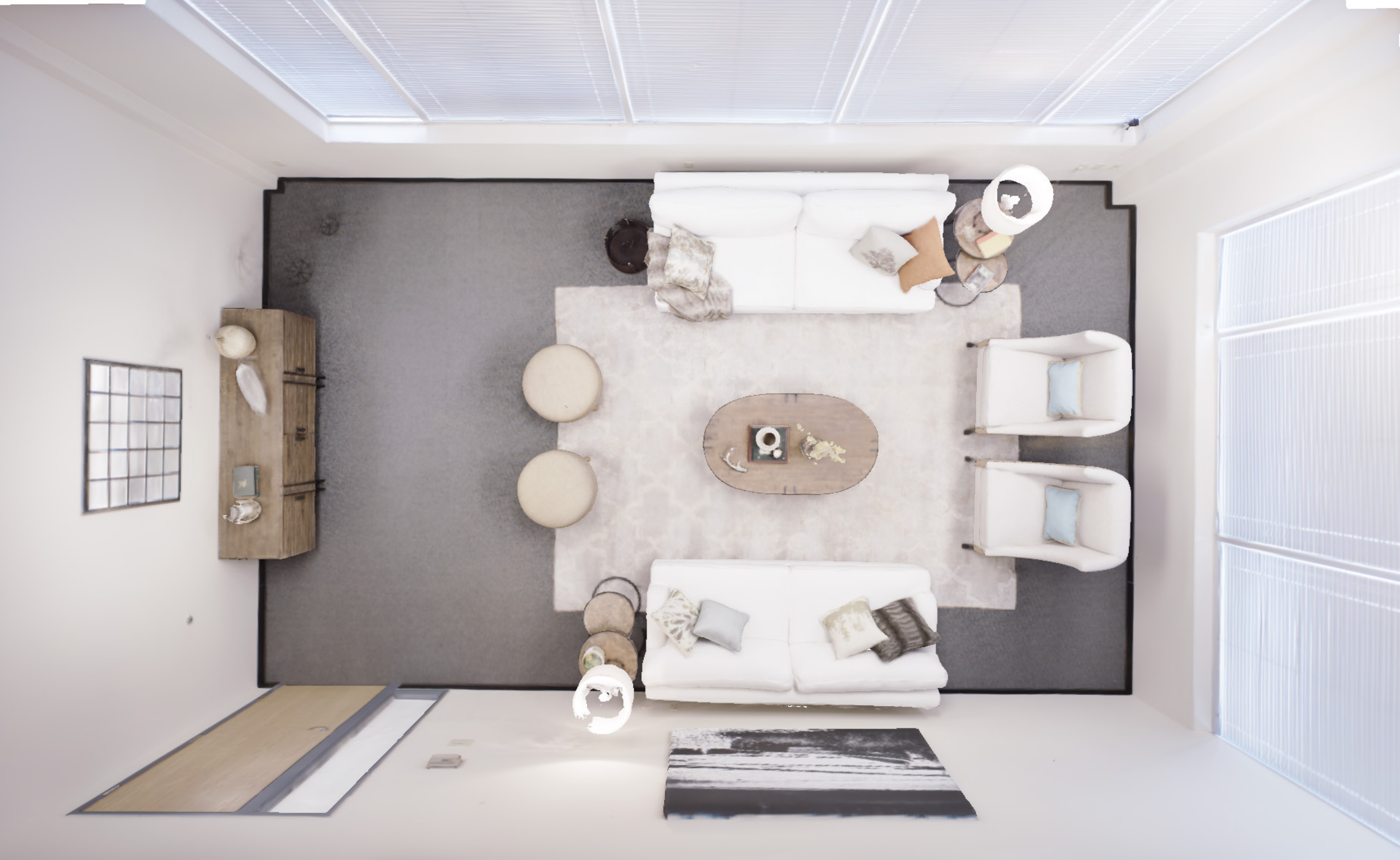}}   &
    \makecell{\includegraphics[height=\sz\columnwidth]{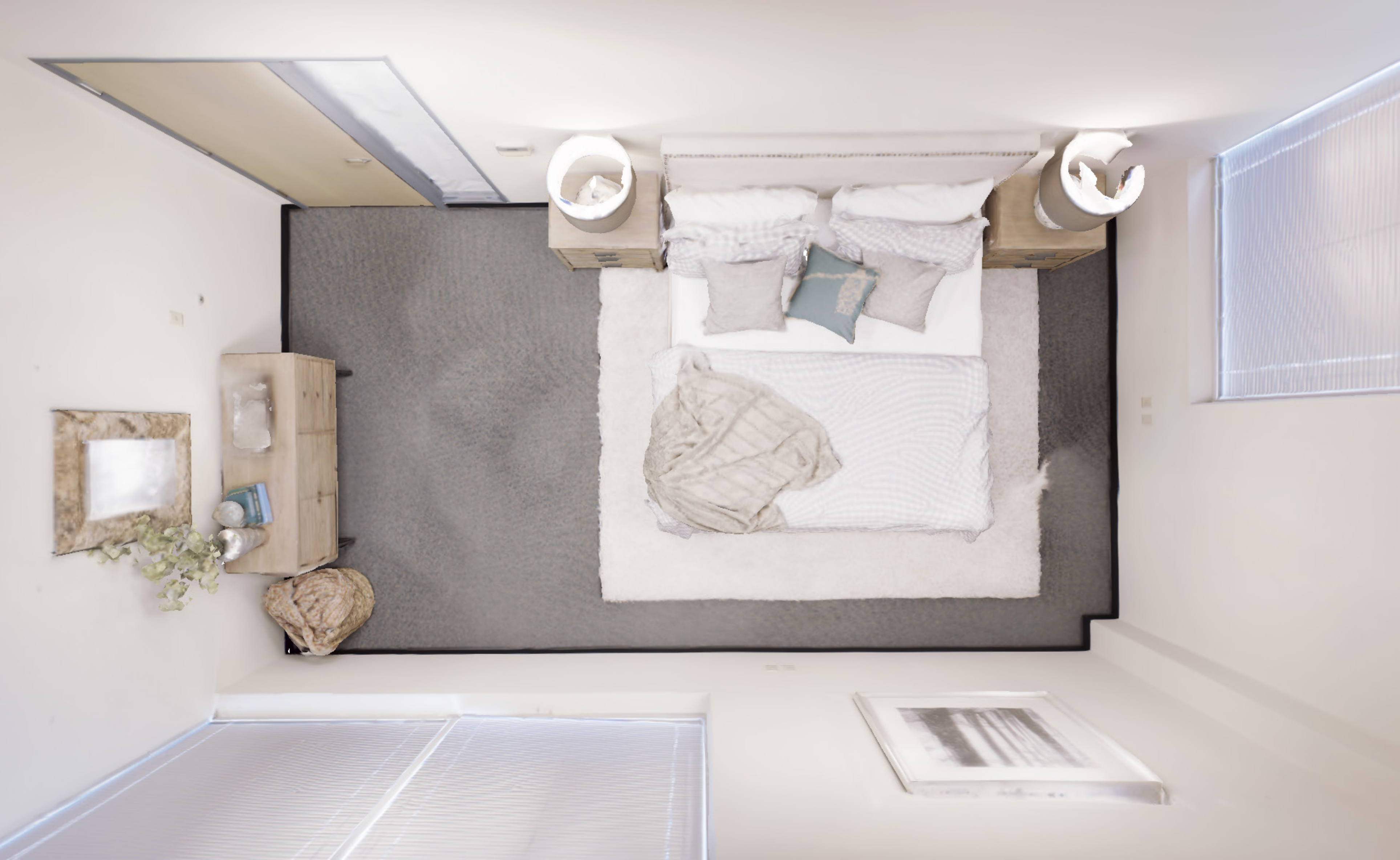}}   &
    \makecell{\includegraphics[height=\sz\columnwidth]{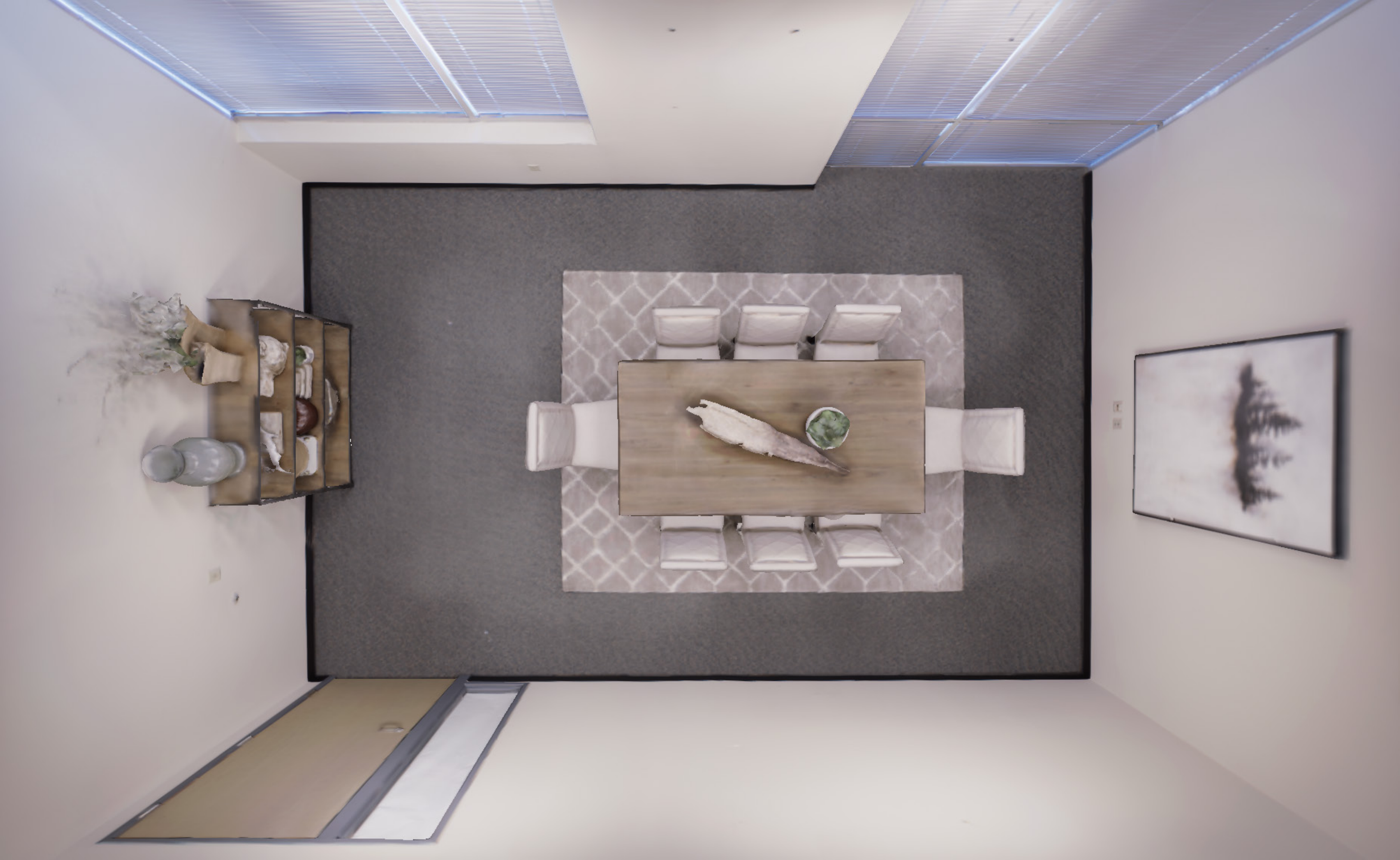}}                                                                                                                                                         \\
  \end{tabular}
  \vspace{-1mm}
  \caption{Partial reconstruction results on Replica dataset\cite{straub2019replica}. All
    baseline methods are based on known scene size. Our method, however, is capable of
    obtaining complete, accurate, and high-quality reconstruction results without the
    need for scene size.}
  \label{render_replica}
  \vspace{-5pt}
\end{figure*}

\begin{table*}[t]
  \centering
  \caption{Reconstruction results on Replica\cite{straub2019replica} and Synthetic
    RGBD\cite{azinovic2022neural} datasets. \textbf{NeB-SLAM} and \textbf{NeB-SLAM}$^{\dagger}$
    denote our methods with hash sizes of 15 and 14, respectively.
    Best results are highlighted as
    \colorbox{colorFst}{\bf first}, \colorbox{colorSnd}{second}, and \colorbox{colorTrd}{third}.
    Our method exhibits more favorable results.}
  \footnotesize
  \setlength{\tabcolsep}{0.27em}
  \begin{tabular}{l l *{9}{c} | *{8}{c}}

    \toprule
    \multirow{2}{*}{Methods} & \multirow{2}{*}{Metrics}                        & \multicolumn{9}{c}{Replica} & \multicolumn{8}{c}{Synthetic}                                                                                                                                                                                                                  \\
    \cmidrule(lr){3-11} \cmidrule(lr){12-19}
                             &                                                 & \texttt{r0}                 & \texttt{r1}                   & \texttt{r2} & \texttt{o0} & \texttt{o1} & \texttt{o2} & \texttt{o3} & \texttt{o4} & Avg.       & \texttt{BR} & \texttt{CK} & \texttt{GR} & \texttt{GWR} & \texttt{MA} & \texttt{TG} & \texttt{WR} & Avg.       \\

    \midrule
    \multirow{4}{*}{iMap$^*$}
                             & \textbf{Depth L1}[cm]$\downarrow$               & 6.85                        & 5.83                          & 6.31        & 6.37        & 4.22        & 6.20        & 8.57        & 6.64        & 6.37       & 23.80       & 63.11       & 30.32       & 35.93        & 56.61       & 19.88       & 65.83       & 42.21      \\
                             & \textbf{Acc.}[cm]$\downarrow $                  & 5.72                        & 4.02                          & 5.39        & 4.16        & 6.38        & 5.95        & 5.35        & 5.38        & 5.29       & 10.29       & 30.16       & 13.16       & 21.68        & 15.13       & 13.97       & 37.85       & 20.32      \\
                             & \textbf{Comp.}[cm]$\downarrow  $                & 5.33                        & 5.70                          & 5.49        & 4.15        & 5.02        & 6.74        & 5.44        & 6.39        & 5.53       & 13.42       & 37.79       & 16.51       & 26.13        & 43.18       & 15.25       & 31.54       & 26.26      \\
                             & \textbf{Comp. Ratio}[\textless 5cm\%]$\uparrow$ & 77.93                       & 76.82                         & 79.08       & 81.14       & 79.74       & 75.08       & 72.06       & 73.02       & 76.86      & 38.05       & 13.84       & 32.47       & 17.96        & 12.58       & 28.32       & 12.37       & 22.23      \\
    \midrule
    \multirow{4}{*}{NICE-SLAM}
                             & \textbf{Depth L1}[cm]$\downarrow$               & 2.63                        & 1.43                          & 2.22        & 1.94        & 4.95        & 2.78        & 2.64        & 2.16        & 2.59       & 5.27        & 15.30       & 3.00        & 2.50         & 2.21        & 9.93        & 6.59        & 6.40       \\
                             & \textbf{Acc.}[cm]$\downarrow $                  & 2.38                        & 2.03                          & 2.18        & 1.79        & 1.78        & 3.09        & 2.99        & 2.68        & 2.37       & 2.53        & 11.08       & 2.20        & 2.69         & 1.78        & 5.10        & 7.03        & 4.63       \\
                             & \textbf{Comp.}[cm]$\downarrow  $                & 3.01                        & 2.31                          & 2.71        & 2.44        & 2.34        & 3.04        & 3.26        & 3.73        & 2.86       & 5.03        & 14.49       & 4.37        & 3.11         & 3.53        & 7.30        & 5.68        & 6.22       \\
                             & \textbf{Comp. Ratio}[\textless 5cm\%]$\uparrow$ & 90.74                       & 93.08                         & 90.76       & 92.56       & 92.13       & 87.75       & 86.54       & 86.24       & 89.98      & 84.44       & 51.70       & 85.69       & 88.25        & 82.27       & 59.53       & 70.61       & 74.64      \\
    \midrule
    \multirow{4}{*}{Co-SLAM}
                             & \textbf{Depth L1}[cm]$\downarrow$               & \nd{0.99}                   & \rd{0.89}                     & \nd{2.28}   & \rd{1.21}   & \nd{1.45}   & \nd{1.78}   & \rd{1.60}   & \nd{1.45}   & \nd{1.46}  & \rd{3.43}   & \rd{6.55}   & \fs{1.96}   & \rd{1.36}    & \nd{1.33}   & \rd{4.85}   & \nd{3.04}   & \rd{3.22}  \\
                             & \textbf{Acc.}[cm]$\downarrow $                  & \rd{2.01}                   & \nd{1.60}                     & \nd{1.90}   & \nd{1.54}   & \fs{1.27}   & \nd{2.69}   & \rd{2.74}   & \rd{2.31}   & \rd{2.01}  & \rd{2.04}   & \nd{4.49}   & \rd{1.96}   & \nd{1.99}    & \rd{1.60}   & \rd{5.61}   & \nd{6.14}   & \rd{3.40}  \\
                             & \textbf{Comp.}[cm]$\downarrow  $                & \rd{2.17}                   & \rd{1.84}                     & \nd{1.94}   & \nd{1.53}   & \nd{1.64}   & \rd{2.47}   & \nd{2.69}   & \nd{2.49}   & \rd{2.10}  & \nd{1.93}   & \fs{5.05}   & \fs{2.73}   & \rd{2.32}    & \rd{2.66}   & \rd{2.67}   & \nd{3.46}   & \nd{2.97}  \\
                             & \textbf{Comp. Ratio}[\textless 5cm\%]$\uparrow$ & \rd{94.35}                  & \rd{95.20}                    & \nd{93.72}  & \rd{96.36}  & \rd{94.48}  & \nd{91.87}  & \rd{91.15}  & \rd{90.95}  & \rd{93.51} & \nd{95.21}  & \nd{67.45}  & \fs{91.90}  & \rd{94.15}   & \nd{87.03}  & \rd{86.82}  & \rd{83.61}  & \nd{86.60} \\
    \midrule
    \multirow{4}{*}{\textbf{NeB-SLAM}}
                             & \textbf{Depth L1}[cm]$\downarrow$               & \fs{0.95}                   & \fs{0.75}                     & \fs{2.25}   & \fs{1.12}   & \fs{1.42}   & \fs{1.73}   & \fs{1.35}   & \fs{1.44}   & \fs{1.38}  & \fs{3.13}   & \fs{5.02}   & \rd{2.06}   & \fs{1.21}    & \fs{1.28}   & \fs{4.04}   & \fs{2.89}   & \fs{2.80}  \\
                             & \textbf{Acc.}[cm]$\downarrow $                  & \fs{1.85}                   & \fs{1.54}                     & \fs{1.76}   & \fs{1.45}   & \nd{1.31}   & \fs{2.56}   & \fs{2.59}   & \nd{2.28}   & \fs{1.92}  & \fs{1.87}   & \fs{4.11}   & \fs{1.69}   & \fs{1.76}    & \nd{1.55}   & \fs{2.77}   & \rd{6.24}   & \fs{2.86}  \\
                             & \textbf{Comp.}[cm]$\downarrow  $                & \fs{2.03}                   & \fs{1.76}                     & \fs{1.57}   & \fs{1.36}   & \rd{1.67}   & \fs{2.31}   & \fs{2.37}   & \fs{2.48}   & \fs{1.94}  & \fs{1.88}   & \rd{5.35}   & \nd{3.07}   & \fs{2.20}    & \fs{2.51}   & \fs{2.37}   & \fs{3.25}   & \fs{2.95}  \\
                             & \textbf{Comp. Ratio}[\textless 5cm\%]$\uparrow$ & \nd{94.78}                  & \fs{95.56}                    & \fs{93.89}  & \fs{97.32}  & \nd{94.67}  & \fs{92.15}  & \fs{91.24}  & \fs{91.18}  & \fs{93.85} & \fs{95.52}  & \fs{67.63}  & \rd{89.80}  & \fs{94.67}   & \rd{86.98}  & \fs{87.92}  & \fs{84.13}  & \fs{86.66} \\
    \midrule
    \multirow{4}{*}{\textbf{NeB-SLAM}$^{\dagger}$}
                             & \textbf{Depth L1}[cm]$\downarrow$               & \rd{1.03}                   & \nd{0.77}                     & \rd{2.34}   & \nd{1.18}   & \nd{1.45}   & \rd{1.89}   & \nd{1.55}   & \fs{1.44}   & \nd{1.46}  & \nd{3.27}   & \nd{5.13}   & \nd{1.99}   & \nd{1.33}    & \nd{1.33}   & \nd{4.38}   & \rd{3.12}   & \nd{2.94}  \\
                             & \textbf{Acc.}[cm]$\downarrow $                  & \nd{1.98}                   & \nd{1.60}                     & \rd{1.91}   & \rd{1.55}   & \rd{1.37}   & \nd{2.69}   & \nd{2.65}   & \fs{2.27}   & \nd{2.00}  & \nd{2.02}   & \rd{4.56}   & \nd{1.87}   & \rd{2.11}    & \fs{1.51}   & \fs{2.77}   & \fs{5.47}   & \nd{2.90}  \\
                             & \textbf{Comp.}[cm]$\downarrow  $                & \nd{2.05}                   & \fs{1.76}                     & \nd{1.94}   & \rd{1.55}   & \fs{1.58}   & \nd{2.45}   & \rd{2.70}   & \rd{2.52}   & \nd{2.07}  & \rd{2.16}   & \nd{5.09}   & \rd{3.53}   & \nd{2.31}    & \nd{2.61}   & \nd{2.46}   & \rd{3.78}   & \rd{3.13}  \\
                             & \textbf{Comp. Ratio}[\textless 5cm\%]$\uparrow$ & \fs{94.79}                  & \nd{95.39}                    & \rd{93.40}  & \nd{96.78}  & \fs{94.84}  & \rd{91.67}  & \nd{91.21}  & \nd{90.98}  & \nd{93.63} & \rd{95.13}  & \rd{67.33}  & \nd{89.89}  & \nd{94.22}   & \fs{87.18}  & \nd{87.33}  & \nd{83.66}  & \rd{86.39} \\
    \bottomrule
  \end{tabular}
  \label{recon_results}
\end{table*}

\subsection{Tracking and Mapping}
\label{tracking}
During the tracking and mapping, we optimize the camera poses
$\{\bm{T}_t\}$ and the scene geometry parameters $\{f_m^\beta\}$,
as well as the network parameters $f^\gamma$ and $f^\delta$, by minimizing
the objective function. For the sampled set of pixels $\mathcal{P}=\{[u_i,v_i]\}$
with corresponding colors $\{c_i\}$ and depths $\{d_i\}$, a ray is generated
for each pixel using the corresponding camera pose via Eq. (\ref{ray}) and
$N_p$ points are sampled on each ray. The depth loss $\mathcal{L}_d$
and color loss $\mathcal{L}_c$ are defined as the
$L_2$ losses between the observation and the results rendered by
Eq. (\ref{vol-ren}):
\begin{equation}
  \begin{split}
    &\mathcal{L}_c=\frac{1}{\left\lvert \mathcal{P}\right\rvert}\sum_{i = 1}^{\left\lvert \mathcal{P}\right\rvert}(\bm{c}_i-\hat{\bm{c}})^2,\\
    &\mathcal{L}_d=\frac{1}{\left\lvert \mathcal{R}\right\rvert}\sum_{r\in \mathcal{R}}(d_r-\hat{d}_r)^2,
  \end{split}
  \label{loss}
\end{equation}
where $\mathcal{R}$ denotes the set of pixels with valid depth measurements
in $\mathcal{P}$. For points within the truncation region
$\mathcal{X}^{tr}=\{\bm{x}\mid |d_i-d_{\bm{x}}|\leq tr\}$, we calculate the SDF loss:
\begin{equation}
  \mathcal{L}_{sdf}=\frac{1}{\left\lvert\mathcal{R}\right\rvert}
  \sum_{r\in\mathcal{R}}\frac{1}{\left\lvert \mathcal{X}_r^{tr}\right\rvert}
  \sum_{\bm{x}\in \mathcal{X}_r^{tr}}(s_{\bm{x}}-\hat{s_{\bm{x}}})^2,
  \label{loss_sdf}
\end{equation}
where $\hat{s_{\bm{x}}}$ is the predicted SDF value of the point $\bm{x}$ and
$s_{\bm{x}}=d_i-d_{\bm{x}}$ is the observed SDF value. For points not in the
truncation region $\mathcal{X}^{fs}=\{\bm{x}\mid |d_i-d_{\bm{x}}|>tr\}$, similar
to \cite{azinovic2022neural,wang2023co}, a free-space loss is applied, which
forces the SDF prediction to be the truncated distance $tr$:
\begin{equation}
  \mathcal{L}_{fs}=\frac{1}{\left\lvert\mathcal{R}\right\rvert}
  \sum_{r\in\mathcal{R}}\frac{1}{\left\lvert \mathcal{X}_r^{fs}\right\rvert}
  \sum_{\bm{x}\in \mathcal{X}_r^{fs}}(\hat{s_{\bm{x}}}-tr)^2,
  \label{loss_fs}
\end{equation}
Furthermore, to prevent the occurrence of noisy reconstruction due to hash
collisions in unobserved free-space regions, following \cite{wang2023co},
we randomly select a set of points $\mathcal{X}^g$ and perform regularization
on the corresponding interpolated features
$v_{\bm{x}}=\frac{1}{M}\sum_{m=1}^{M}f_m^\beta(\bm{x}-\bm{C}_m)$:
\begin{equation}
  \mathcal{L}_{reg}=\frac{1}{\left\lvert\mathcal{X}^g\right\rvert}
  \sum_{\bm{x}\in\mathcal{X}^g}(\Delta_{x}^2+\Delta_{y}^2+\Delta_{z}^2),
  \label{loss_reg}
\end{equation}
where $\Delta_{x,y,z}=v_{\bm{x}+\epsilon_{x,y,z}}-v_{\bm{x}}$ denotes
the feature-metric difference between adjacent sampled vertices on the
hash-grid along the three dimensions. In summary, the objective of our
optimization process is to minimize a combination of the aforementioned
losses:
\begin{equation}
  \mathcal{L}=\lambda_1\mathcal{L}_c+\lambda_2\mathcal{L}_d+
  \lambda_3\mathcal{L}_{sdf}+\lambda_4\mathcal{L}_{fs}+
  \lambda_5\mathcal{L}_{reg},
  \label{loss_total}
\end{equation}
here $\lambda_{1,2,3,4,5}$ are the corresponding weights.

\textbf{Tracking.}
During tracking, we estimate the camera pose $\bm{T}_t$ for each
frame. When a new frame comes in, the pose of the current frame
$t$ is initialized using the constant velocity assumption:
\begin{equation}
  \bm{T}_t=\bm{T}_{t-1}{\bm{T}_{t-2}^{-1}\bm{T}_{t-1}},
  \label{pose_update}
\end{equation}
Then, a random selection of pixels $N_t$ is made, and sample points are
generated. The corresponding estimates are rendered
by the method described in Sec. \ref{rendering}. Finally, the camera
pose is optimized iteratively by minimizing the objective function
Eq. (\ref{loss_total}). For each frame, following the optimization
of its pose, a determination is made as to whether a new NeB should
be allocated by the method described in Sec. \ref{neb-alloc}. In the
pose optimization process, only regions covered by NeBs are sampled,
and thus regions that are currently not covered by NeBs will not affect
the optimization.

In the context of keyframe selection, a fixed number of frames is
employed in a manner analogous to other methods. However, when a new
NeB is allocated at current frame, this frame is designated as a
keyframe without the need for further selection.

\textbf{Mapping.}
During the mapping process, we employ the keyframe data management
strategy in \cite{wang2023co}. Instead of storing the complete keyframe
image, only a subset of pixels is stored to represent each keyframe.
This approach enables more frequent insertion of new keyframes and
maintains a larger keyframe database. For joint optimization, we
randomly sample $N_m$ rays from the global keyframe list to optimize
our scene representation $\{f_m^\beta\}$, MLPs $f^\gamma$, $f^\delta$
and camera poses $\{T_t\}$. The rendering approach and the optimization
objective function are identical to those employed in the tracking process.
Furthermore, in the event that the number of NeBs exceeds one, an
additional $N_a$ rays are collected in the keyframe
sequences corresponding to the most recent NeB to be included in the
joint optimization process. This is done in order to accelerate the
convergence of the most recent NeB.

\subsection{Loop Closure}
To address arbitrary drift, a BoW\cite{salinas2017dbow3} model for global
position identification is utilized, wherein each global keyframe is incorporated.
Upon the generation of a global keyframe, it is appended to the aforementioned
database. This methodology contrasts with that of MIPSFusion\cite{tang2023mips},
which employs
submap overlap for the detection of loop closures and is constrained to the
correction of smaller drifts.

For each new keyframe with camera pose $\bm{T}_c=[\bm{R}|\bm{t}]$, a loop
closure is identified by querying the BoW database. The first $K$ keyframes
from BoW that are not adjacent to the current keyframe are queried, and the
two frames $\bm{T}_{r1}$ and $\bm{T}_{r2}$ with the highest
visual similarity score and the greatest timestamp distance are identified
as the closed-loop keyframes. The threshold is defined as the minimum
similarity score between the current keyframe and its neighboring keyframes.
For each closed-loop keyframe, the reprojection error between it and the
current keyframe is calculated in order to optimize the pose of the latter:
\begin{equation}
  \begin{split}
    &\bm{T}_{c}=\arg\min_{\bm{T}_{c}}\sum_{i=1}^{2}\sum_{j=1}^{N}\|\bm{u}_{ij}^{\prime}-\bm{u}_{j}\|_{2}^{2}.
  \end{split}
  \label{loop_closure}
\end{equation}
The process commences with the back-projection of the pixels $\bm{u}_j$
belonging to the current keyframe into the world coordinate system
$\bm{P}_{j}=\bm{T}_c\pi^{-1}(\bm{u}_{j})$. Subsequently, the point is
projected onto the corresponding closed-loop keyframe
$\bm{u}_{ij}=\pi(\bm{T}_{ri}^{-1}\bm{P}_{j})$, and the corresponding
depth value is obtained by interpolating the corresponding pixel
coordinates with the depth map. Subsequently, the depth value is
utilized to backproject $\bm{u}_{ij}$ to the world coordinate system
and project it to the current keyframe
$\bm{u}_{ij}^{\prime}=\pi(\bm{T}_c^{-1}\bm{T}_{ri}\pi^{\prime-1}(\bm{u}_{ij}))$.
Here, $\pi$ is used to perform dehomogenization and perspective projection
and $\pi^{\prime}$ employs the depth data obtained through interpolation.
Subsequently, the optimized current pose $\bm{T}_c$ and relative poses
between frames are employed to adjust the keyframe poses between closed
loops. Ultimately, the camera poses and map between the closed loops are
optimized in two stages using the aforementioned method in Sec.
\ref{tracking}. Initially, the camera poses are fixed in order
to optimize the map. Thereafter, the camera poses and maps are optimized concurrently.


\begin{figure}
  \centering
  \scriptsize
  \setlength{\tabcolsep}{0.5pt}
  \newcommand{\sz}{0.32}  %
  \begin{tabular}{lccc}
    \makecell{\rotatebox{90}{iMAP$^*$~\cite{2021imap}}}                                &
    \makecell{\includegraphics[width=\sz\linewidth]{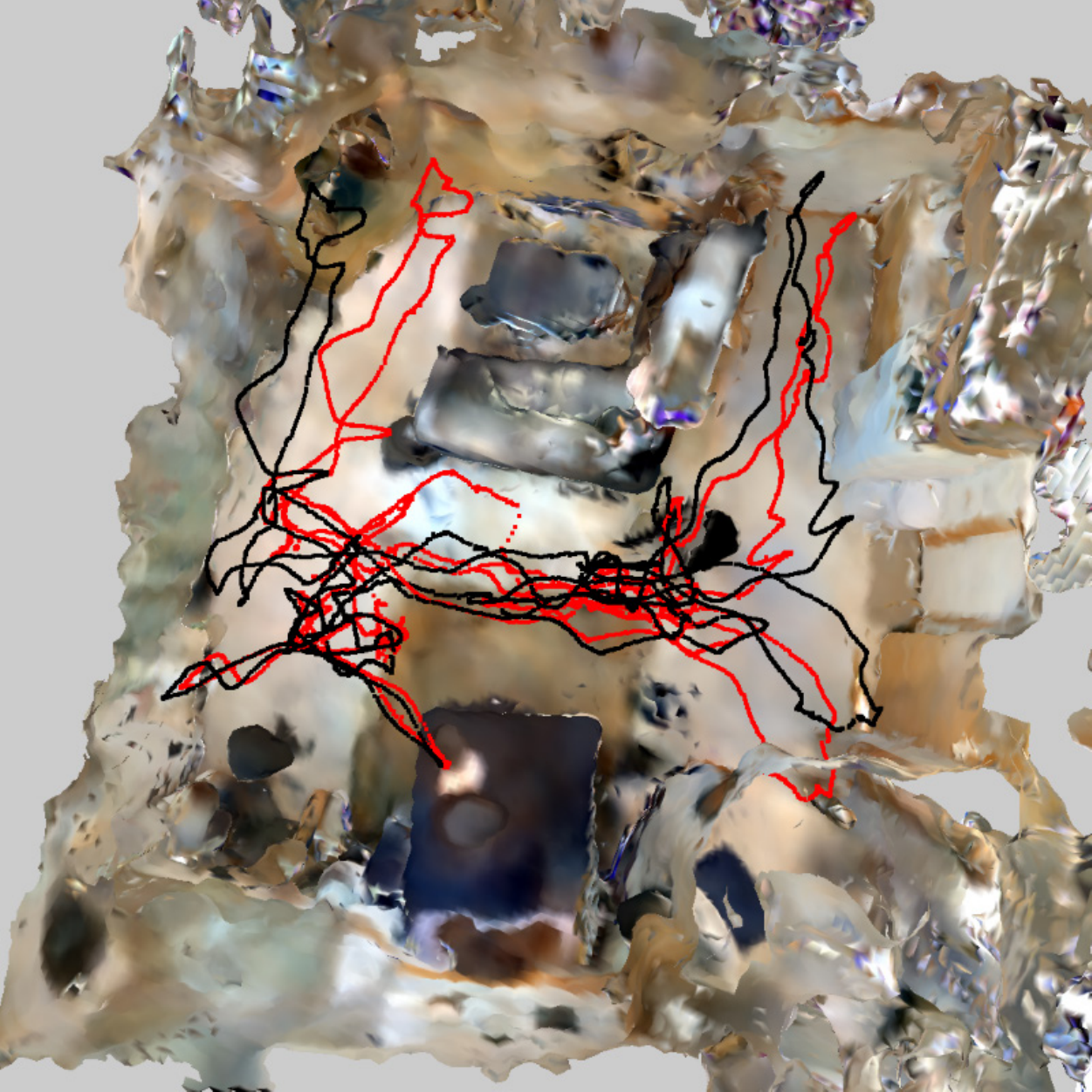}}        &
    \makecell{\includegraphics[width=\sz\linewidth]{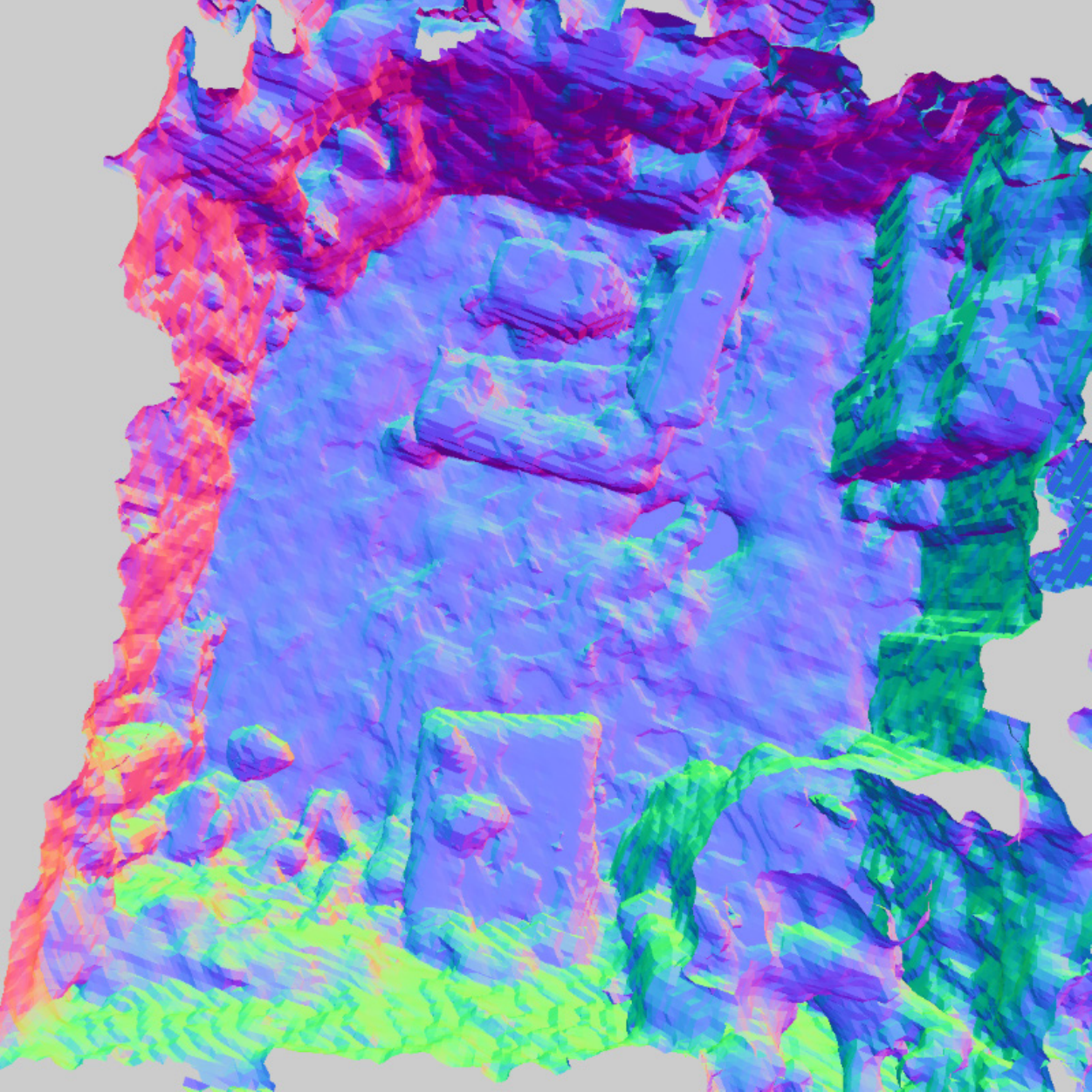}} &
    \makecell{\includegraphics[width=\sz\linewidth]{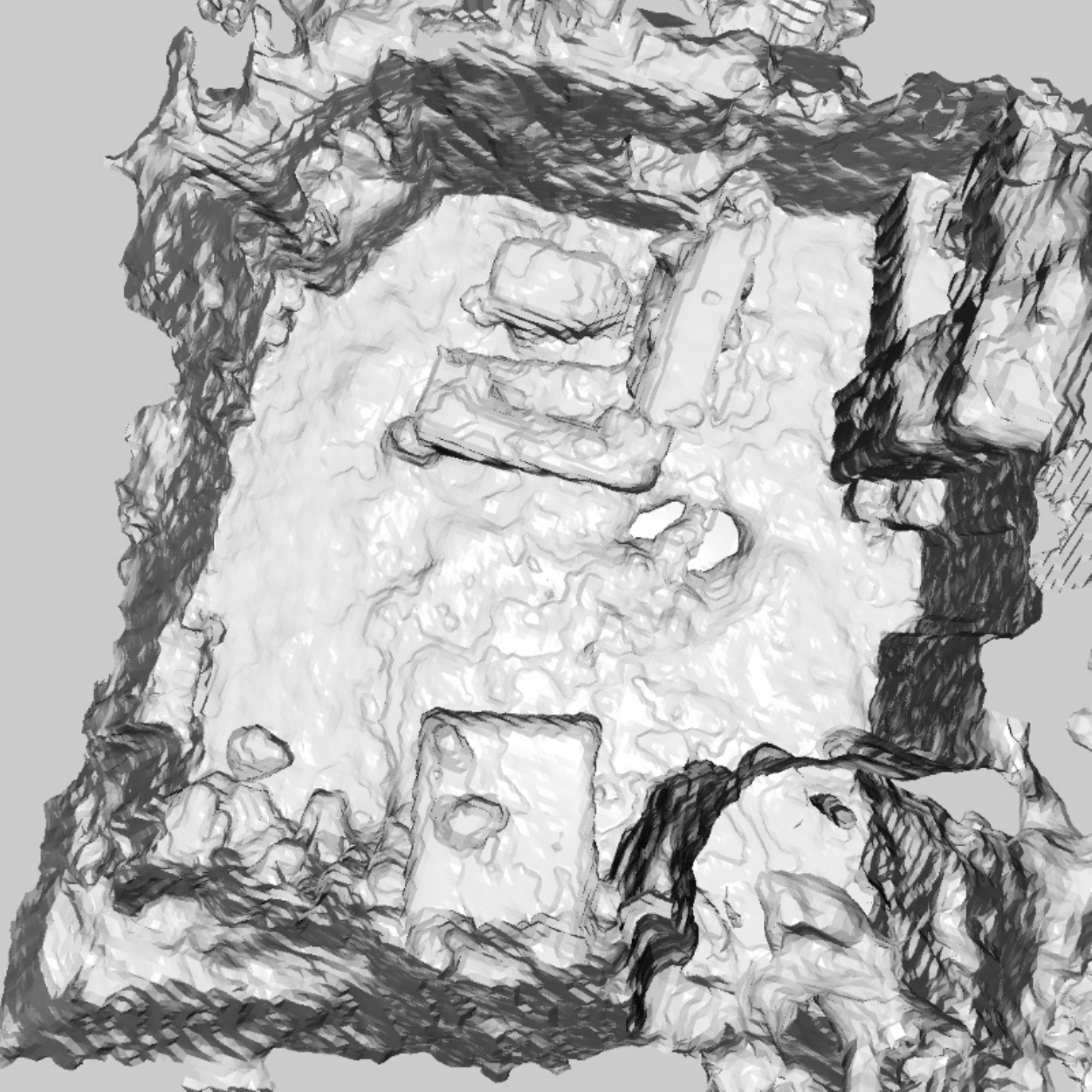}}     \\

    \makecell{\rotatebox{90}{NICE-SLAM~\cite{2022nice}}}                               &
    \makecell{\includegraphics[width=\sz\linewidth]{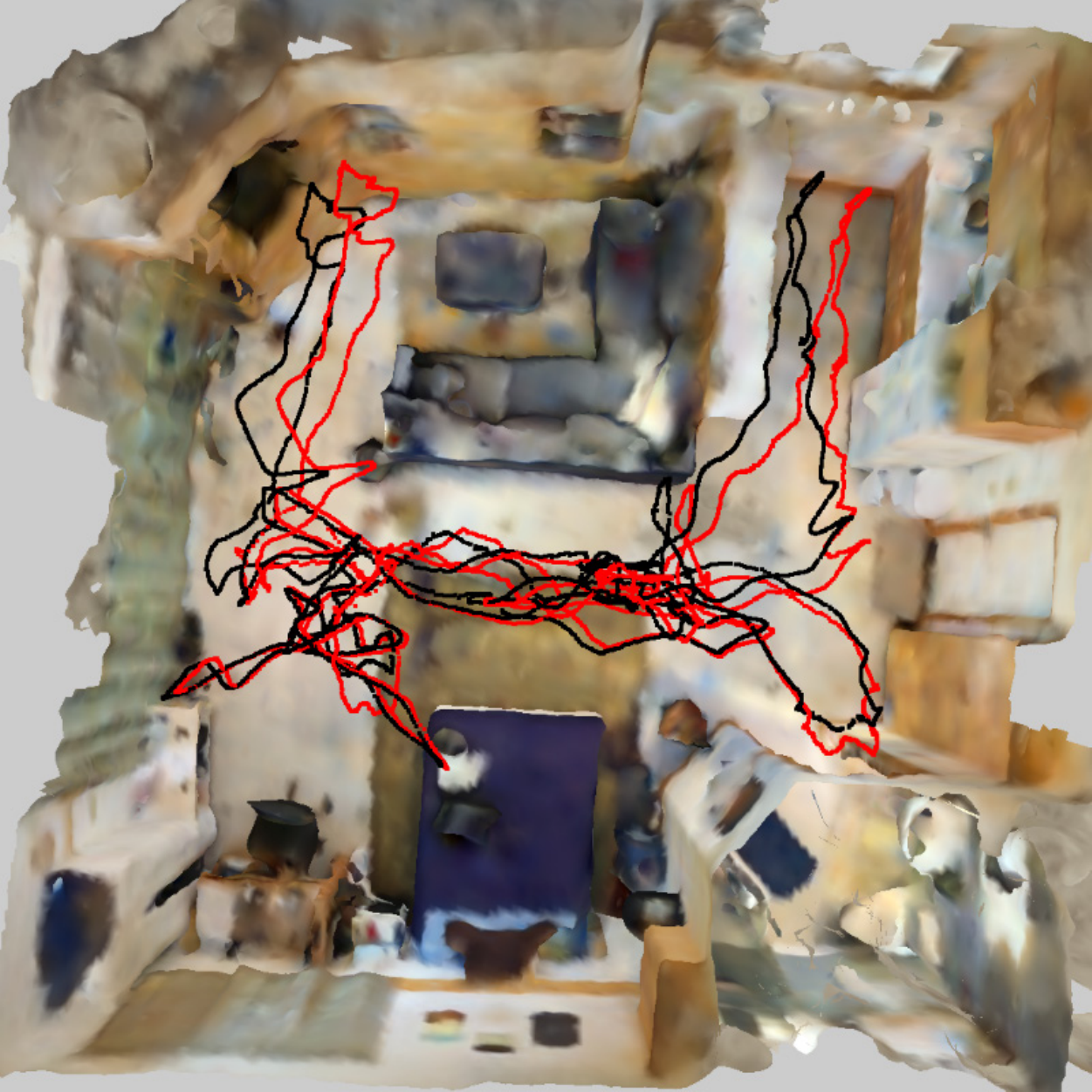}}        &
    \makecell{\includegraphics[width=\sz\linewidth]{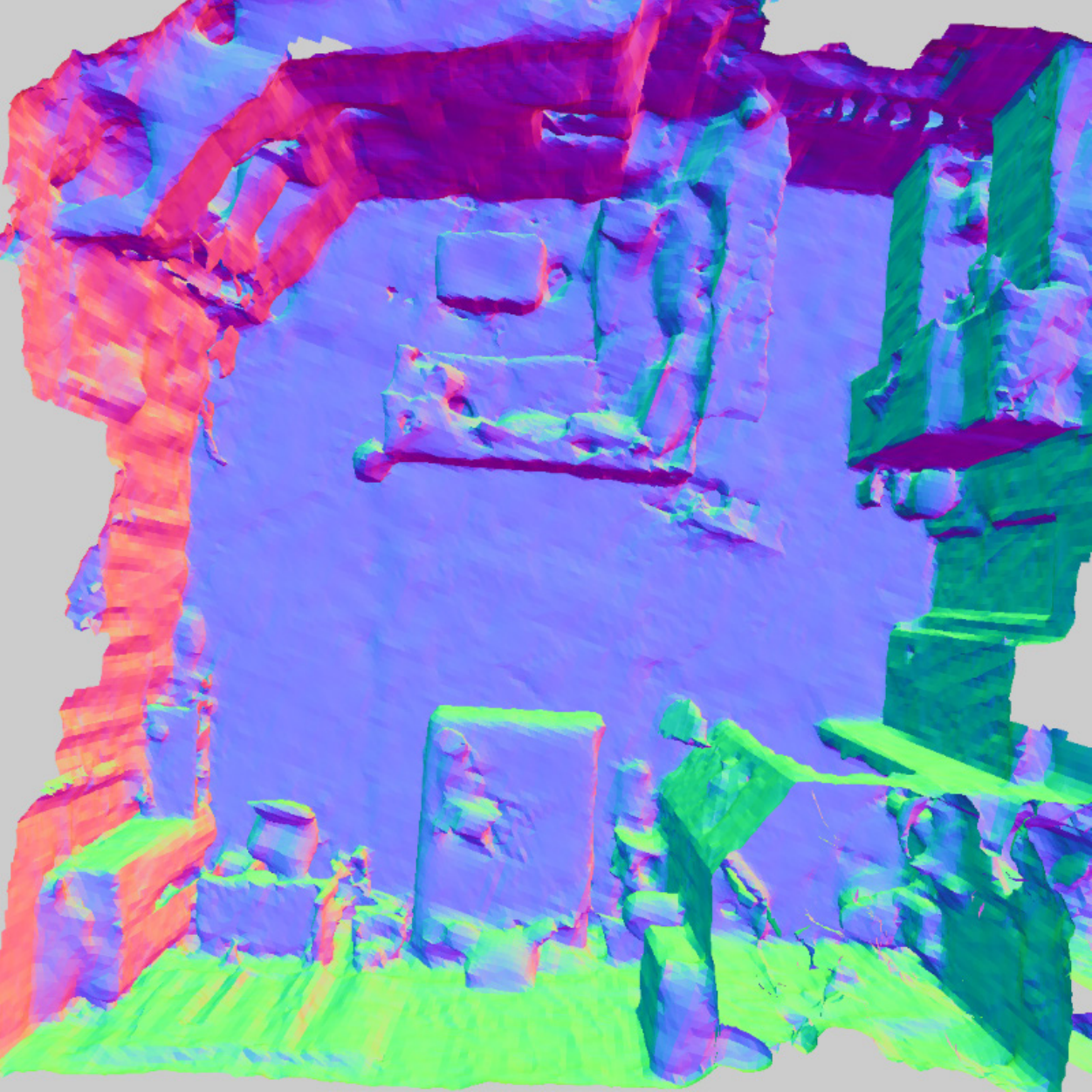}} &
    \makecell{\includegraphics[width=\sz\linewidth]{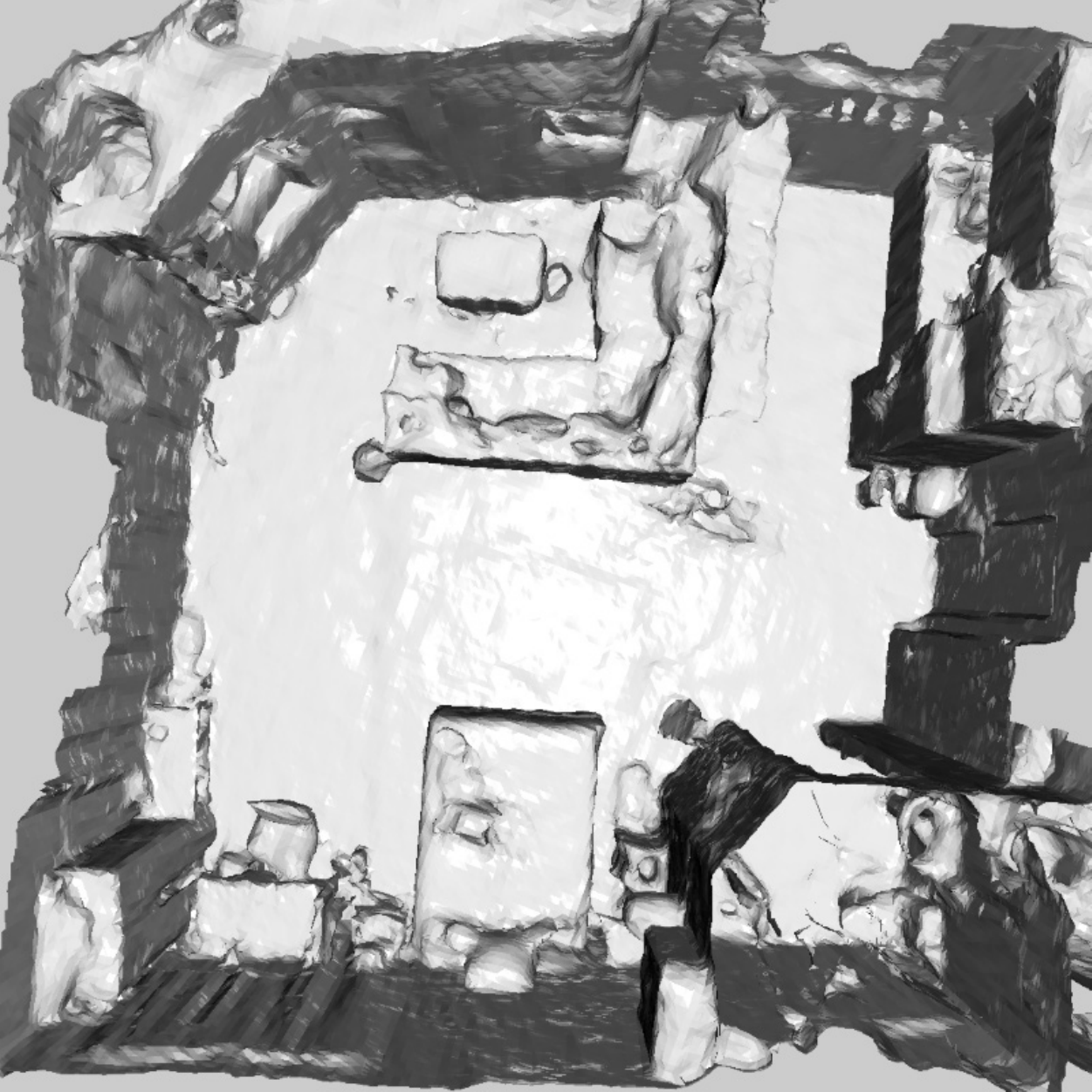}}     \\

    \makecell{\rotatebox{90}{Co-SLAM~\cite{wang2023co}}}                               &
    \makecell{\includegraphics[width=\sz\linewidth]{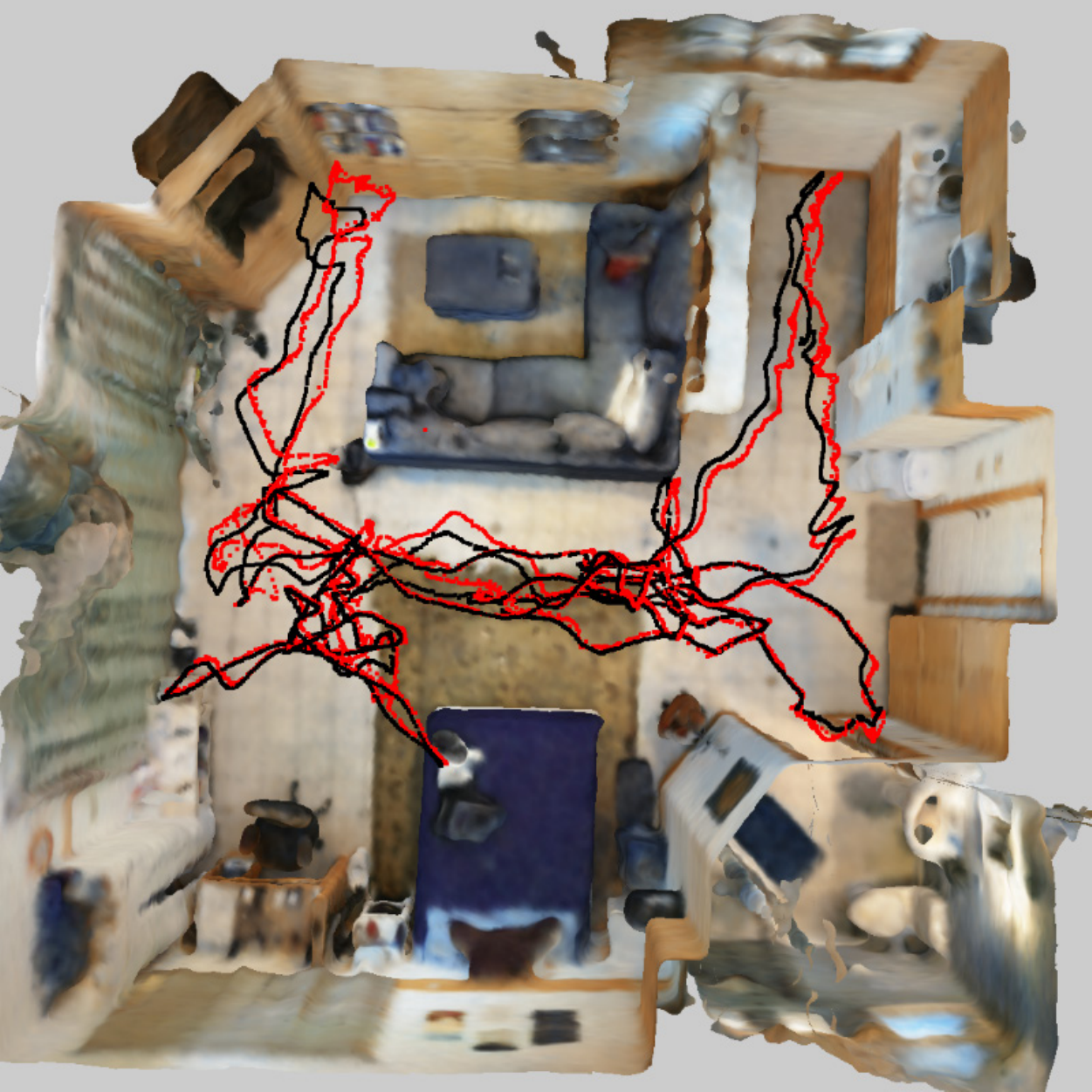}}          &
    \makecell{\includegraphics[width=\sz\linewidth]{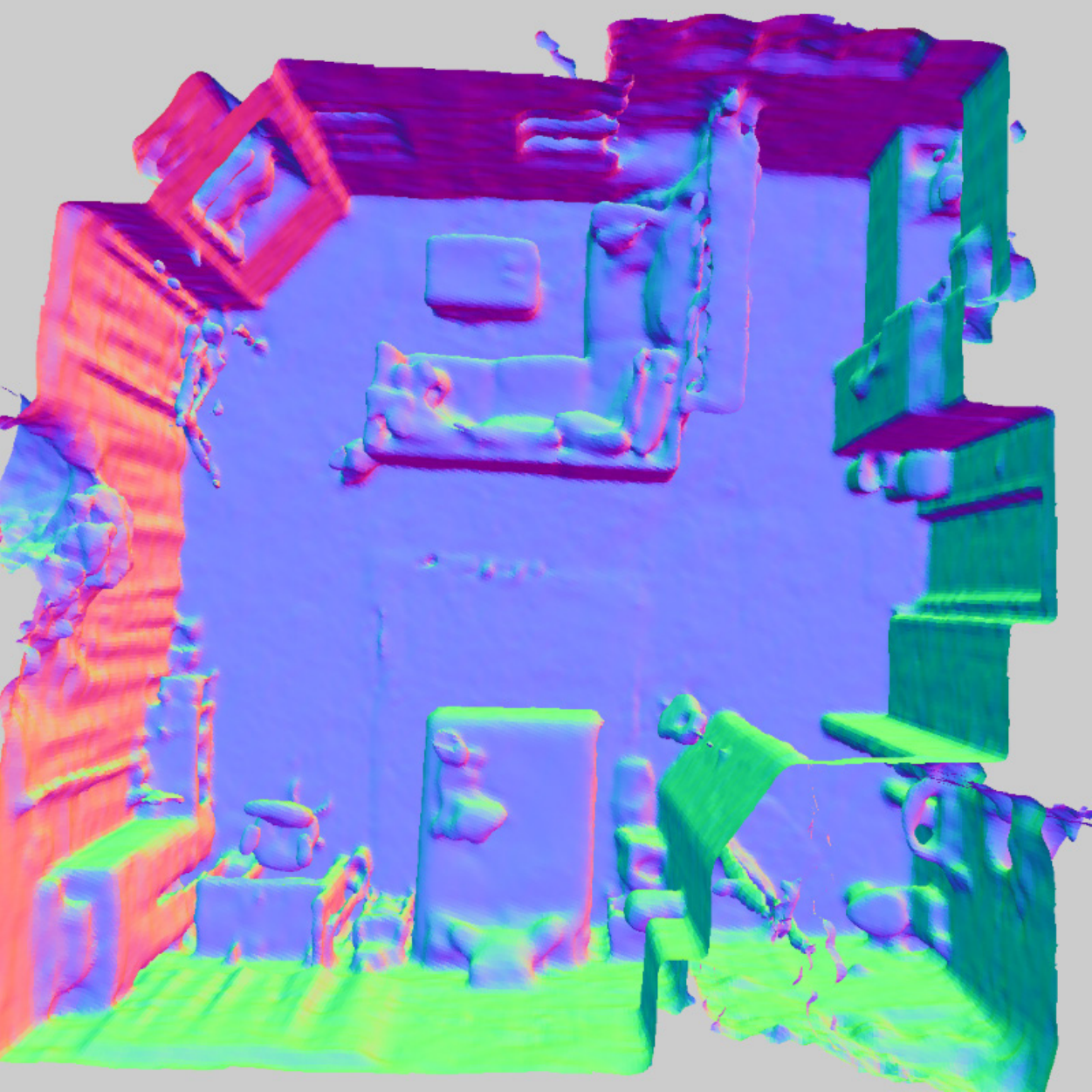}}   &
    \makecell{\includegraphics[width=\sz\linewidth]{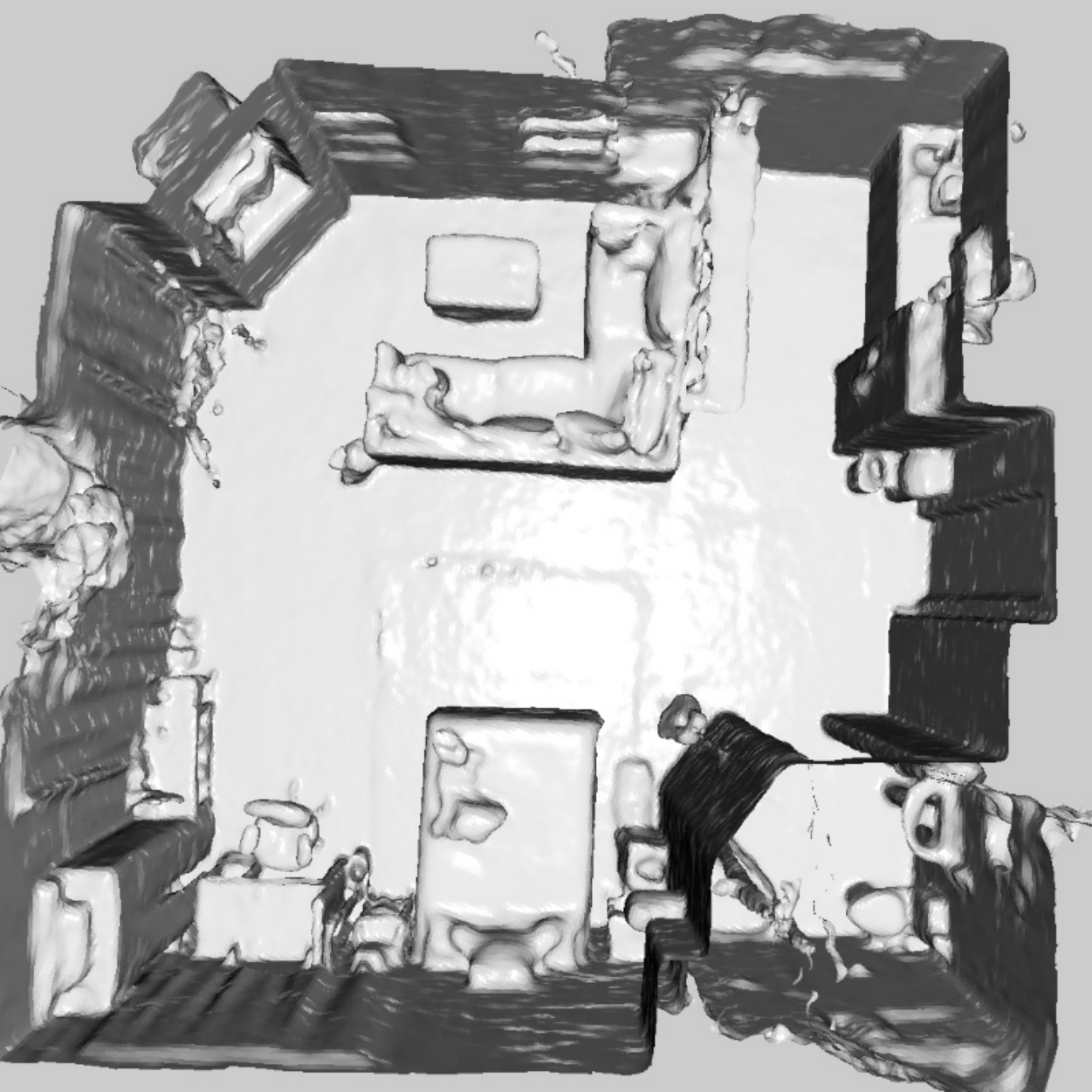}}       \\

    \makecell{\rotatebox{90}{NeB-SLAM}}                                                &
    \makecell{\includegraphics[width=\sz\linewidth]{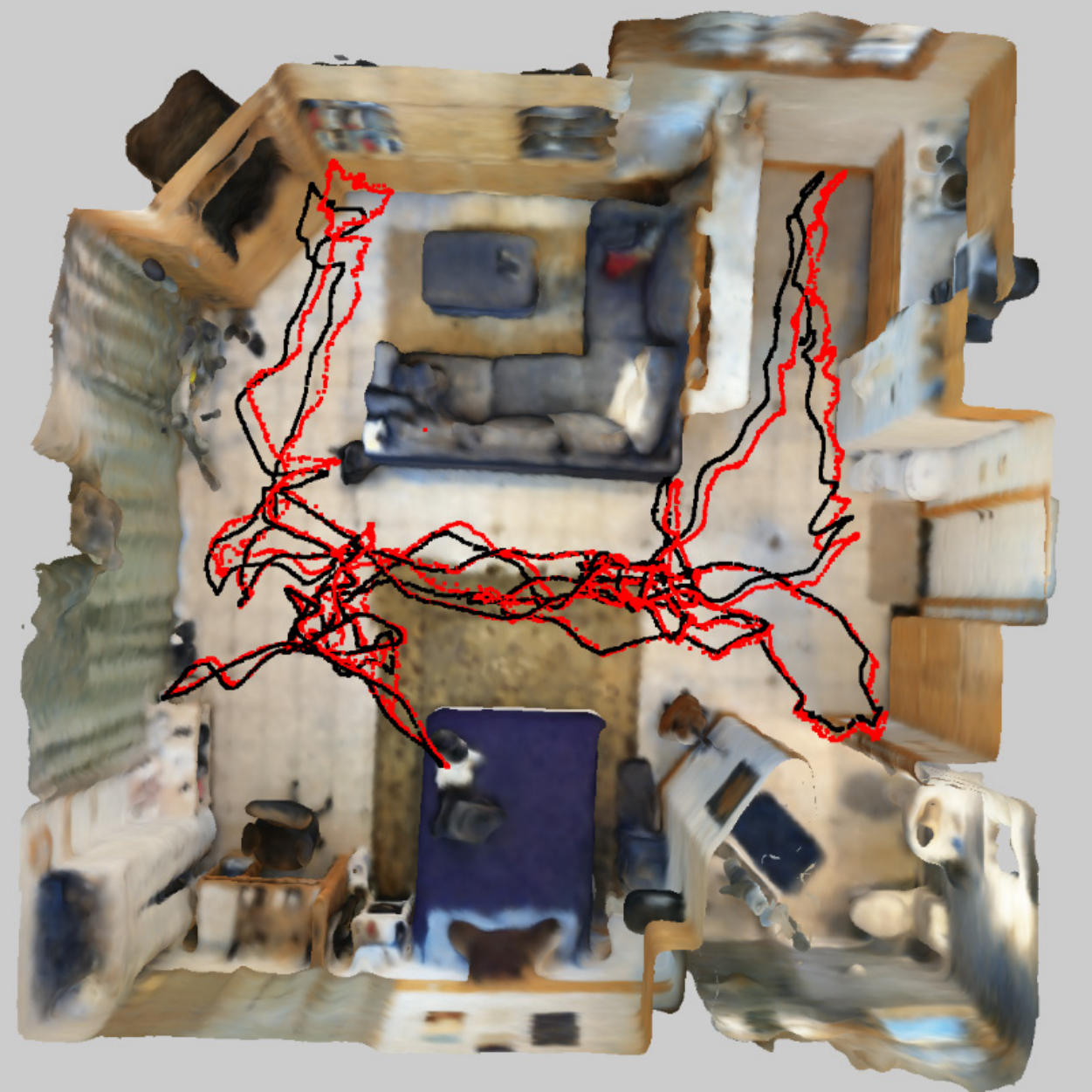}}         &
    \makecell{\includegraphics[width=\sz\linewidth]{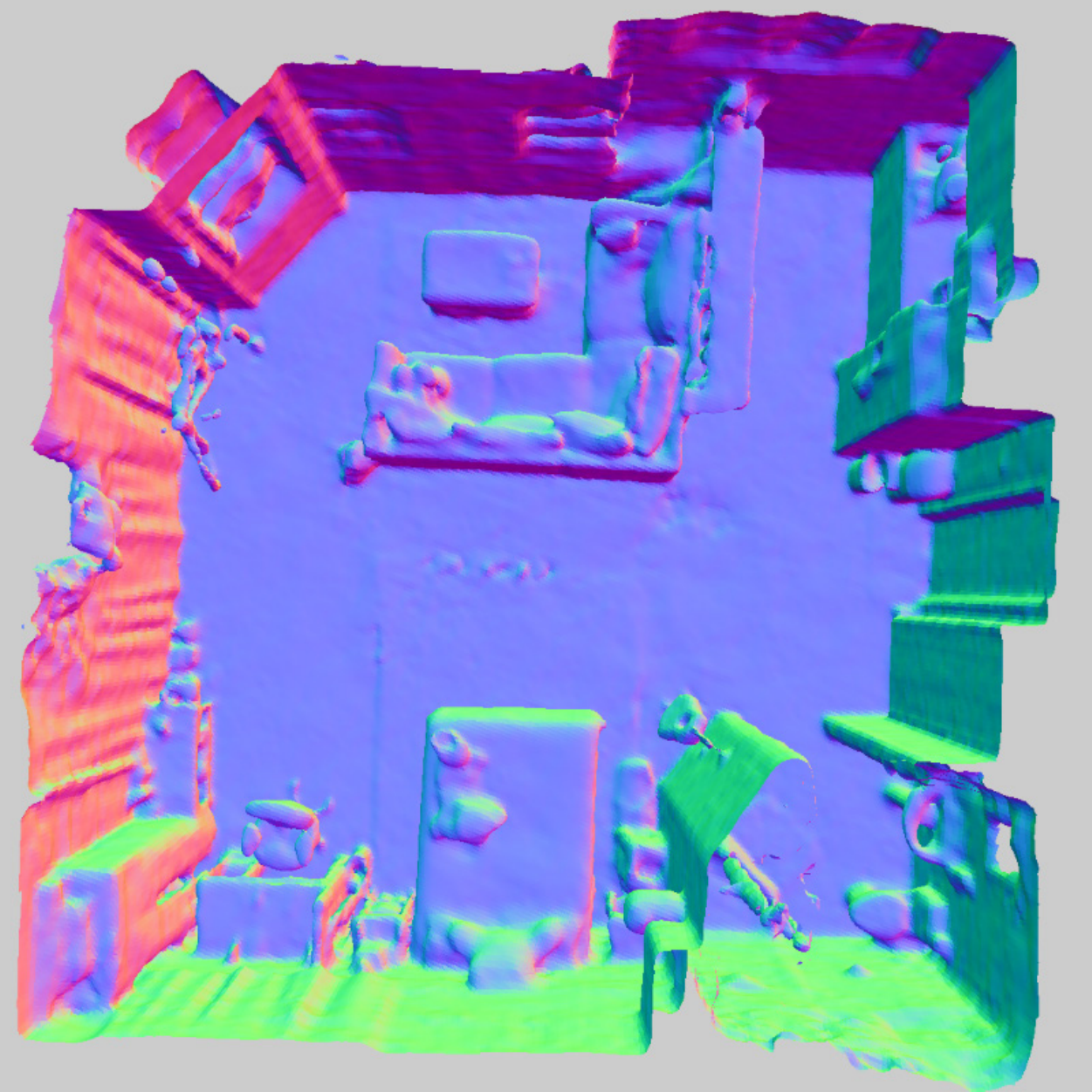}}  &
    \makecell{\includegraphics[width=\sz\linewidth]{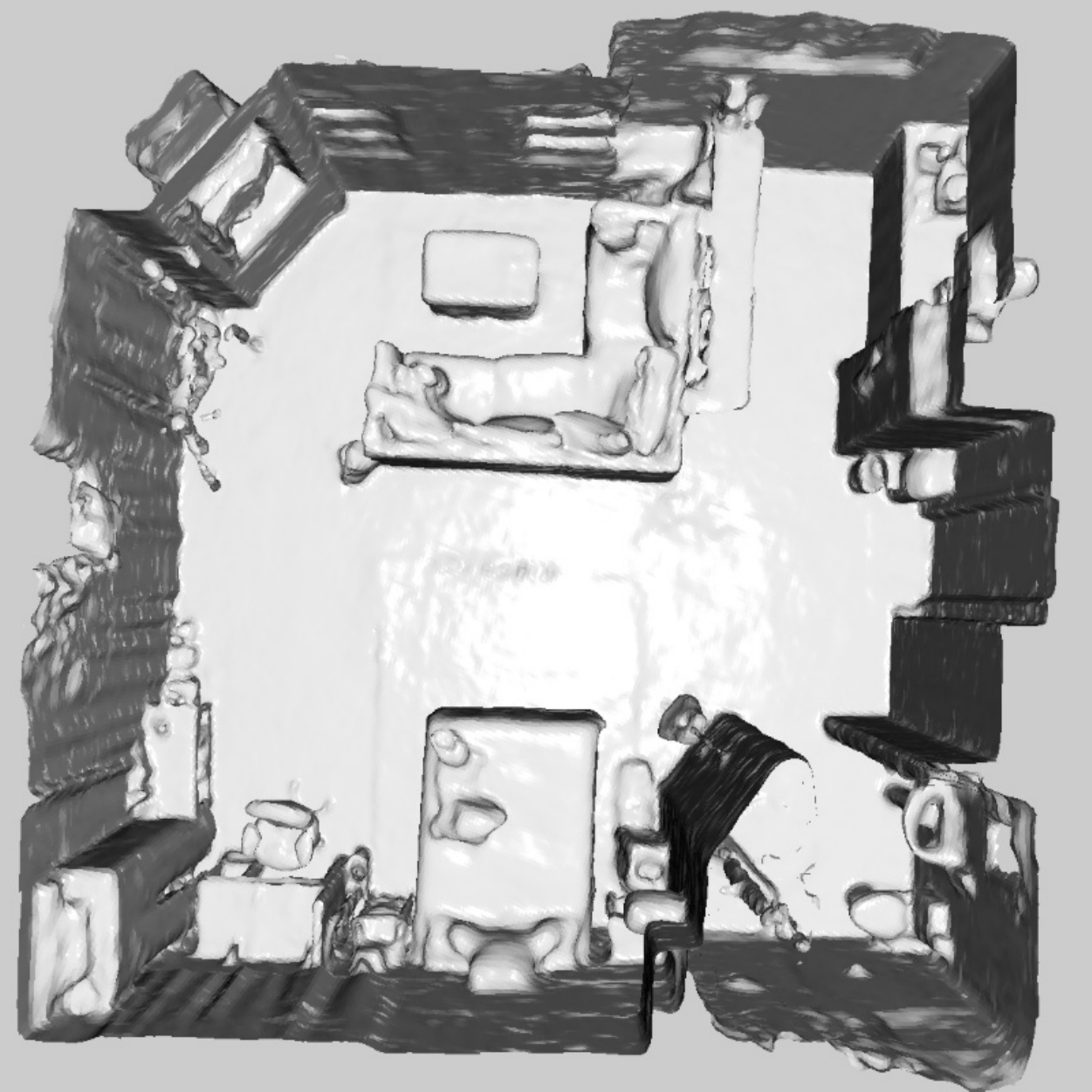}}      \\

    \makecell{\rotatebox{90}{GT}}                                                      &
    \makecell{\includegraphics[width=\sz\linewidth]{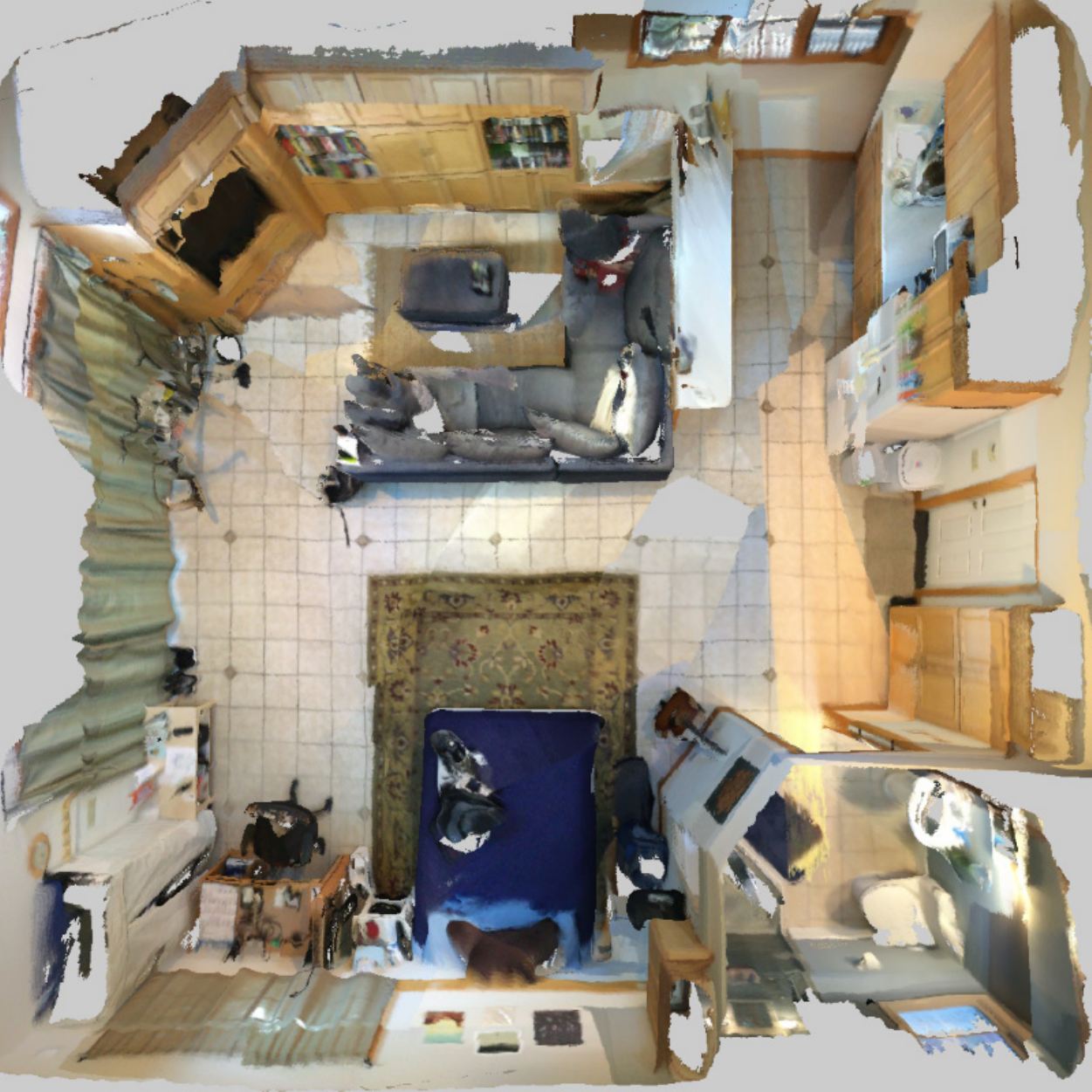}}          &
    \makecell{\includegraphics[width=\sz\linewidth]{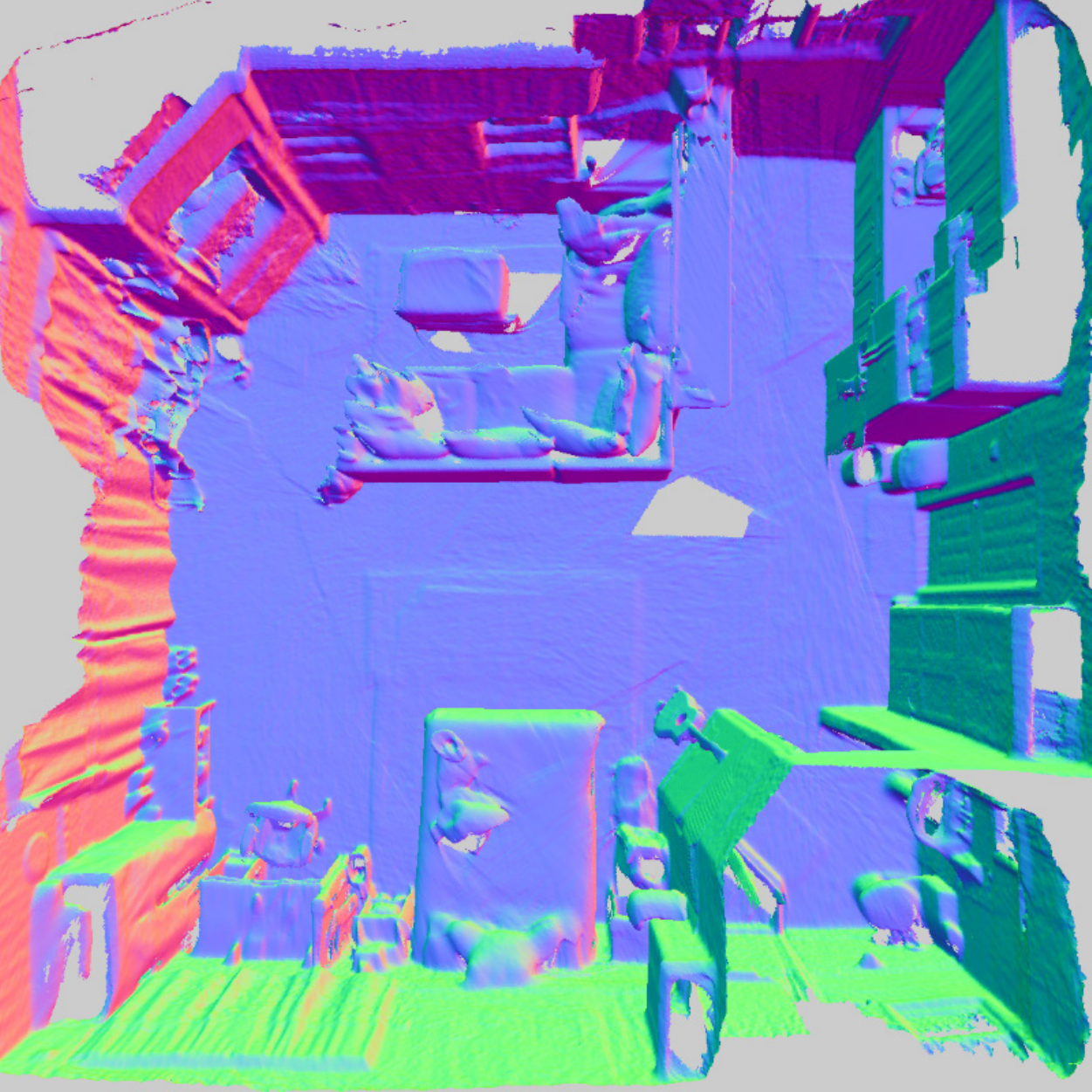}}   &
    \makecell{\includegraphics[width=\sz\linewidth]{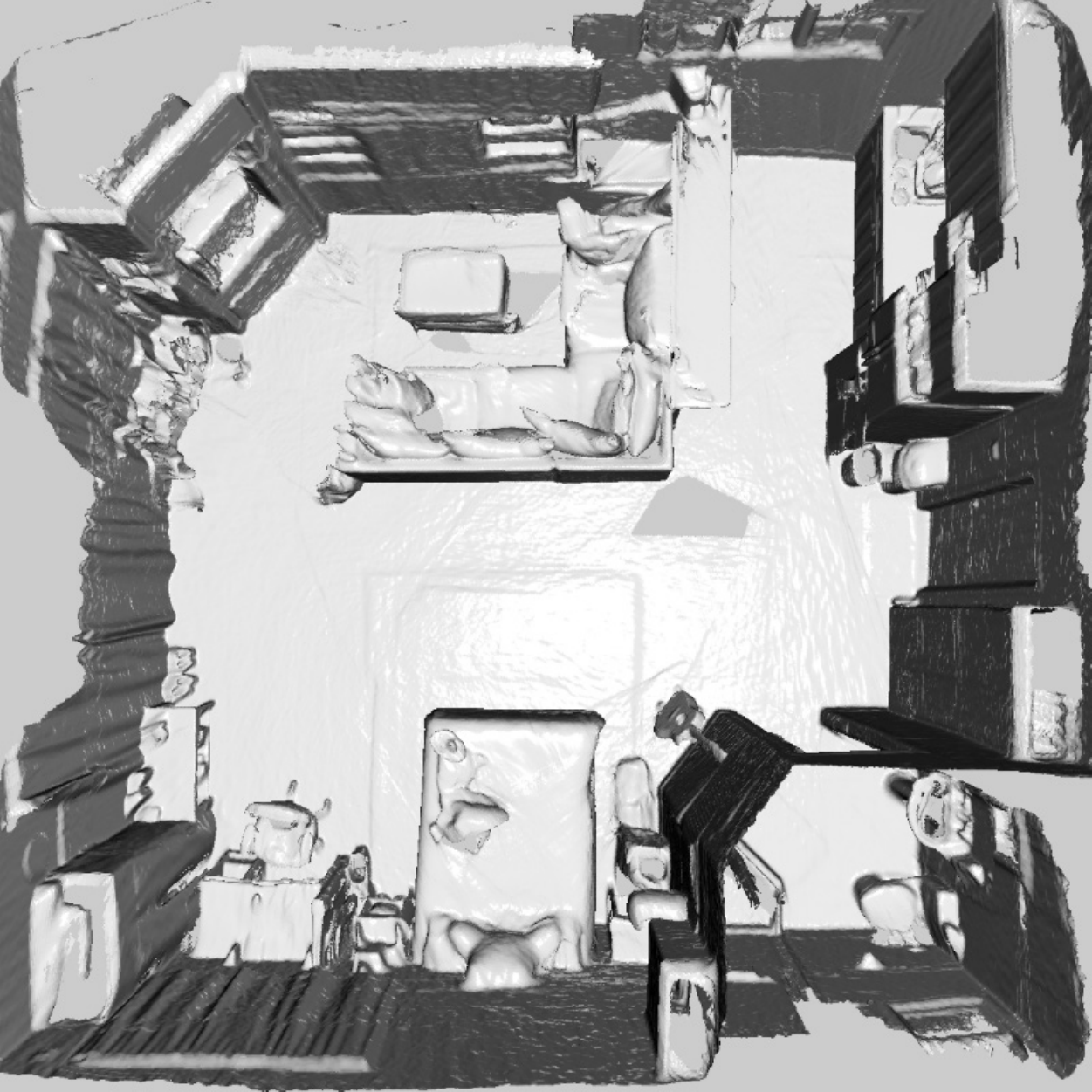}}       \\

                                                                                       &
    \makecell{(a)}                                                                     &
    \makecell{(b)}                                                                     &
    \makecell{(c)}                                                                       \\
  \end{tabular}
  \vspace{-1mm}
  \caption{Qualitative comparison on ScanNet\cite{dai2017scannet} \texttt{scene0000} with
    different shading mode. Our methods achieve accurate scene reconstruction without
    the need for scene size. In all the figures ground truth trajectory are shown in
    black and the estimated trajectory are shown in red.}
  \label{render_scannet}
  \vspace{-5pt}
\end{figure}

\section{Experiments}
\label{Experiment}

\subsection{Experimental Setup}
\subsubsection{Datasets}
NeB-SLAM is evaluated on four different datasets, each containing a
distinct set of scenes. Following iMAP\cite{2021imap}, NICE-SLAM\cite{2022nice},
and Co-SLAM\cite{wang2023co}, the reconstruction quality of 8 synthetic
scenes in Replica\cite{straub2019replica} are quantified. Additionally,
7 synthetic scenes from NeuralRGBD\cite{azinovic2022neural} are evaluated.
For the purpose of evaluating pose estimation, 6 scenes from
ScanNet\cite{dai2017scannet} are considered, where the ground-truth (GT) poses
were obtained with BundleFusion\cite{dai2017bundlefusion}, and
3 scenes from the TUM RGB-D dataset\cite{sturm2012benchmark} are evaluated,
where the GT poses were provided by a motion capture system. In addition to the
aforementioned datasets, we also evaluated the reconstruction quality
of our method on the apartment dataset\cite{2022nice}, which was collected by the
Nice-SLAM authors using Azure Kinect with a larger scene size than the
previous ones.

\subsubsection{Metric}
For fair comparisons, following \cite{wang2023co}, any unobserved regions
outside the camera
frustum and noise points within the camera frustum but outside the target
scene are removed. After mesh culling, The reconstruction quality is
evaluated using the following 2D and 3D metrics: \textbf{Depth L1} (cm),
\textbf{Acc}uracy (cm), \textbf{Comp}letion (cm), and \textbf{Comp}letion
\textbf{Rate} (\%) with a threshold of 5cm. For 2D metric, following
\cite{2022nice}, we sample N = 1000 virtual views from the GT
and reconstructed mesh. Any views with unobserved points were rejected
and resampled. The depth L1 is then defined as the average L1 difference
between the rendered GT depth and the reconstructed depth. For 3D metrics,
We proceed in accordance with the methodology outlined in \cite{2021imap},
initially sampling 200,000 points $\mathcal{G}$ and $\mathcal{R}$ from
the GT and reconstruction meshes
in a uniform manner. \textbf{Acc} is quantified as the average distance
between sampled points on the reconstructed mesh and the nearest point on
the GT mesh. \textbf{Comp} is evaluated as the average distance between
sampled points from the GT mesh and the nearest point on the reconstructed
mesh. Finally, \textbf{Comp} \textbf{Rate} is determined as the percentage
of points in the reconstructed mesh with a completion of less than 5 cm.
\begin{equation}
  \begin{split}
    &\textbf{Acc}=\sum_{\bm{g}\in\mathcal{G}}(\min_{\bm{r}\in\mathcal{R}}
    \left\lVert \bm{g}-\bm{r} \right\rVert )/\left\lvert \mathcal{G} \right\rvert\\
    &\textbf{Comp}=\sum_{\bm{r}\in\mathcal{R}}(\min_{\bm{g}\in\mathcal{G}}
    \left\lVert \bm{g}-\bm{r} \right\rVert )/\left\lvert \mathcal{R} \right\rvert\\
    &\textbf{Comp Rate}=\sum_{\bm{r}\in\mathcal{R}}(\min_{\bm{g}\in\mathcal{G}}
    \left\lVert \bm{g}-\bm{r} \right\rVert <0.05)/\left\lvert \mathcal{R} \right\rvert
  \end{split}
\end{equation}
In the context of camera tracking evaluation, the absolute trajectory
error (ATE) RMSE (cm)\cite{sturm2012benchmark} is employed.
Unless otherwise stated, the results are reported as the average of
five runs by default.

\begin{table}
  \centering
  \caption{ATE RMSE (cm) results on Replica dataset\cite{straub2019replica}. NeB-SLAM
    achieves better performance compared to baseline methods.}
  \footnotesize
  \setlength{\tabcolsep}{0.38em}
  \begin{tabular}{lccccccccc}
    \toprule
    Methods                       & \texttt{r0} & \texttt{r1} & \texttt{r2} & \texttt{o0} & \texttt{o1} & \texttt{o2} & \texttt{o3} & \texttt{o4} & Avg.      \\ \midrule
    iMap$^*$                      & 13.72       & 3.76        & 4.73        & 5.30        & 3.25        & 12.88       & 6.06        & 11.53       & 7.65      \\
    NICE-SLAM                     & 2.11        & 4.91        & 1.47        & 1.22        & 1.83        & 2.07        & 4.77        & 1.52        & 2.49      \\
    Vox-Fusion                    & \fs 0.27    & 1.33        & \rd 0.47    & 0.70        & 1.11        & \nd 0.46    & \fs 0.26    & \rd 0.58    & \rd 0.65  \\
    MIPS-Fusion                   & 1.10        & 1.20        & 1.10        & 0.70        & 0.80        & 1.30        & 2.20        & 1.10        & 1.19      \\
    SplaTAM                       & \nd 0.31    & \nd 0.40    & \fs 0.29    & \fs 0.47    & \fs 0.27    & \fs 0.29    & \nd 0.32    & \nd 0.55    & \fs 0.36  \\
    Co-SLAM                       & 0.63        & 1.20        & 0.99        & 0.56        & \rd{0.55}   & 2.08        & 1.61        & 0.69        & 1.04      \\
    \textbf{NeB-SLAM}             & \rd 0.42    & \fs{0.34}   & \nd{0.43}   & \nd{0.50}   & \nd{0.50}   & \rd{1.17}   & \rd{0.82}   & \fs{0.52}   & \nd{0.59} \\
    \textbf{NeB-SLAM}$^{\dagger}$ & 0.57        & \rd{0.45}   & 0.85        & \rd{0.58}   & 0.57        & 1.25        & 0.93        & 0.65        & 0.73      \\\bottomrule
  \end{tabular}
  \label{ate_replica}
\end{table}

\begin{table}
  \centering
  \caption{ATE RMSE (cm) results on Synthetic RGBD dataset\cite{azinovic2022neural}. Our method achieves
    the best tracking performance in every scene.}
  \footnotesize
  \setlength{\tabcolsep}{0.36em}
  \begin{tabular}{lcccccccc}
    \toprule
    Methods                       & \texttt{br} & \texttt{ck} & \texttt{gr} & \texttt{gwr} & \texttt{ma} & \texttt{tg} & \texttt{wr} & Avg.      \\ \midrule
    iMap$^*$                      & 9.21        & 30.57       & 21.23       & 15.73        & 218.60      & 117.14      & 268.45      & 97.28     \\
    NICE-SLAM                     & 3.60        & 6.50        & 2.75        & 3.07         & 1.83        & 52.07       & 3.70        & 10.50     \\
    Co-SLAM                       & \rd{1.95}   & \nd{1.88}   & \rd{1.24}   & \rd{1.29}    & \rd{0.74}   & \rd{2.29}   & \rd{1.84}   & \nd{1.25} \\
    \textbf{NeB-SLAM}             & \fs{0.71}   & \nd{1.42}   & \fs{0.96}   & \fs{0.85}    & \fs{0.37}   & \fs{0.58}   & \fs{1.02}   & \fs{0.75} \\
    \textbf{NeB-SLAM}$^{\dagger}$ & \nd{0.74}   & \fs{1.40}   & \nd{1.13}   & \nd{1.03}    & \nd{0.71}   & \nd{0.98}   & \nd{1.54}   & \rd{0.93} \\\bottomrule
  \end{tabular}
  \label{ate_synthetic}
\end{table}

\subsubsection{Baselines}
The present study compares the reconstruction quality and camera tracking
of the following different methods: iMAP\cite{2021imap}, NICE-SLAM\cite{2022nice},
Co-SLAM\cite{wang2023co}, Vox-Fusion\cite{yang2022vox}, MIPS-Fusion\cite{tang2023mips}
and SplaTAM\cite{keetha2024splatam}. iMAP$^*$ is the reimplemention released by
the authors of NICE-SLAM. Prior to the comparison, all methods implement
the mesh culling strategy previously described. It is important to note
that these methods are designed for known scenes, whereas our method is
intended for unknown scenes. This distinction is evident in the input
parameters of our method, which do not include the scene size, in contrast
to the aforementioned methods.

\subsubsection{Implementation Details}
These methods are executed on a desktop PC with an Intel Core i9-14900KF
CPU and NVIDIA RTX 4090 GPU.
In our method (NeB-SLAM), $\tau_{th}$ is set to 0.2. During camera tracking, a sample of $N_t=1024$ pixels
is taken and 10 iterations are performed to optimize the camera pose.
During mapping, we sample $N_m=2048$ and $N_a=512$ pixels, and utilize
200 iterations for the first frame mapping and 10 iterations for each
subsequent five frames of BA. For each ray, we uniformly sample 32
points and depth-guided sample 11 points.
For each NeB ($5\times5\times5$ m$^3$), a 16-level HashGrid from Rmin = 16
to Rmax is employed, with a maximum hash entries $T=2^{15}$ per level,
Rmax is determined by the voxel size of 2 cm, and OneBlob encodes 16 bins
per dimension. Additionally, we provide smaller
memory versions (NeB-SLAM$^{\dagger}$) that utilize an $T=2^{14}$ HashGrid
for each NeB. Two 2-layer shallow MLPs with 32
neurons are utilized to decode SDFs and colors. The geometric
feature $\bm{g}$ has a size of 15. We optimize the
camera pose using a learning rate of $1e-3$ in tracking and the feature grids,
decoders, and camera poses during mapping using learning rates of $1e-2$,
$1e-2$ and $1e-3$. The weights for each loss are $\lambda_c = 5$,
$\lambda_d = 0.1$, $\lambda_{sdf} = 1000$, $\lambda_{fs} = 10$, and
$\lambda_{reg} = 1e-6$. The truncation distance, $tr$, is set to 10 cm.
The learning rate of camera pose in tracking for TUM dataset is set to $1e-2$
and we set $tr$ and iterations of BA to 5 cm and 20 respectively. For
ScanNet, we uniformly sample 96 points and depth-guided sample 21 points.

\begin{figure}
  \centering
  \scriptsize
  \setlength{\tabcolsep}{0.5pt}
  \newcommand{\sz}{0.5}  %
  \begin{tabular}{cc}
    \makecell{\includegraphics[width=\sz\linewidth]{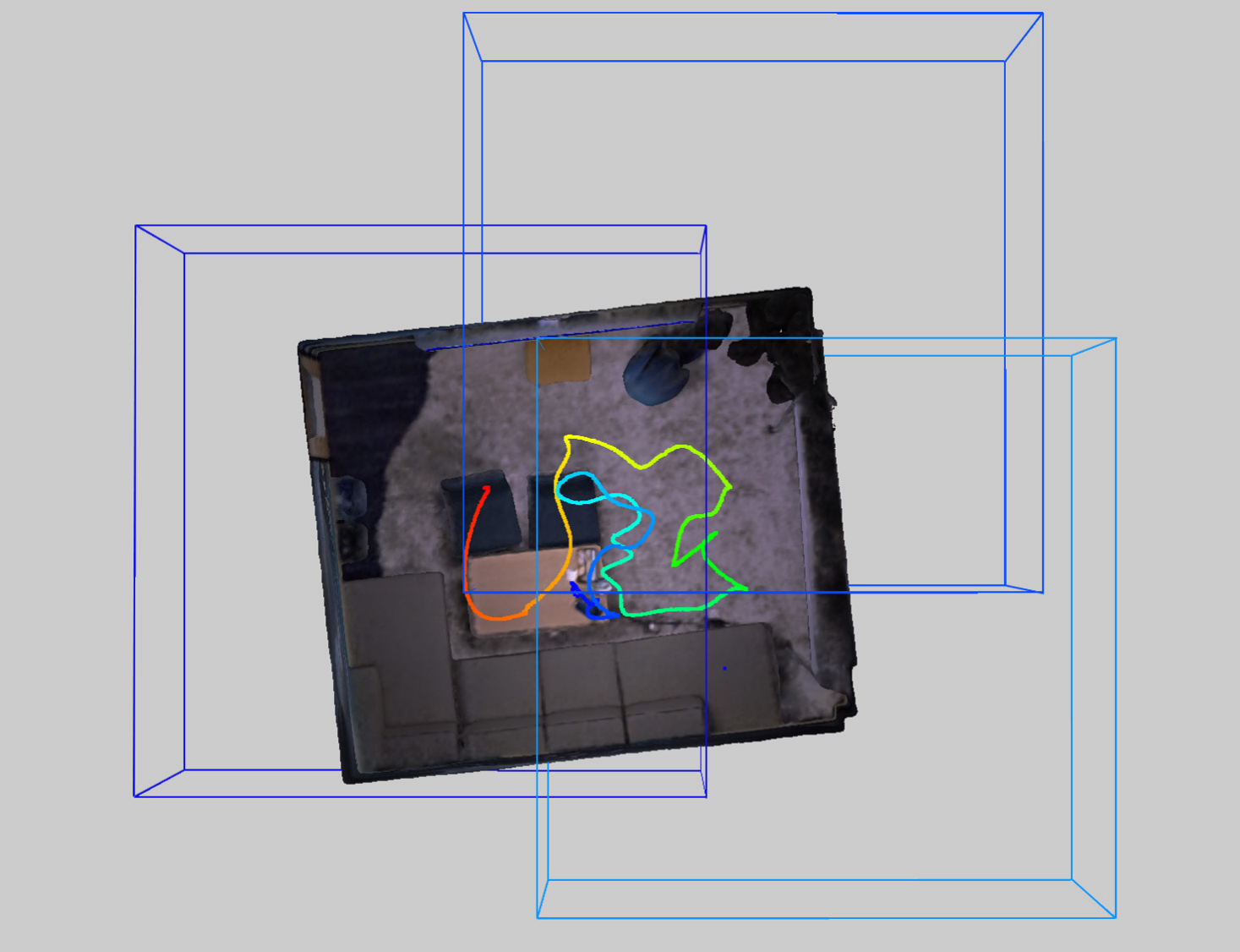}}   &
    \makecell{\includegraphics[width=\sz\linewidth]{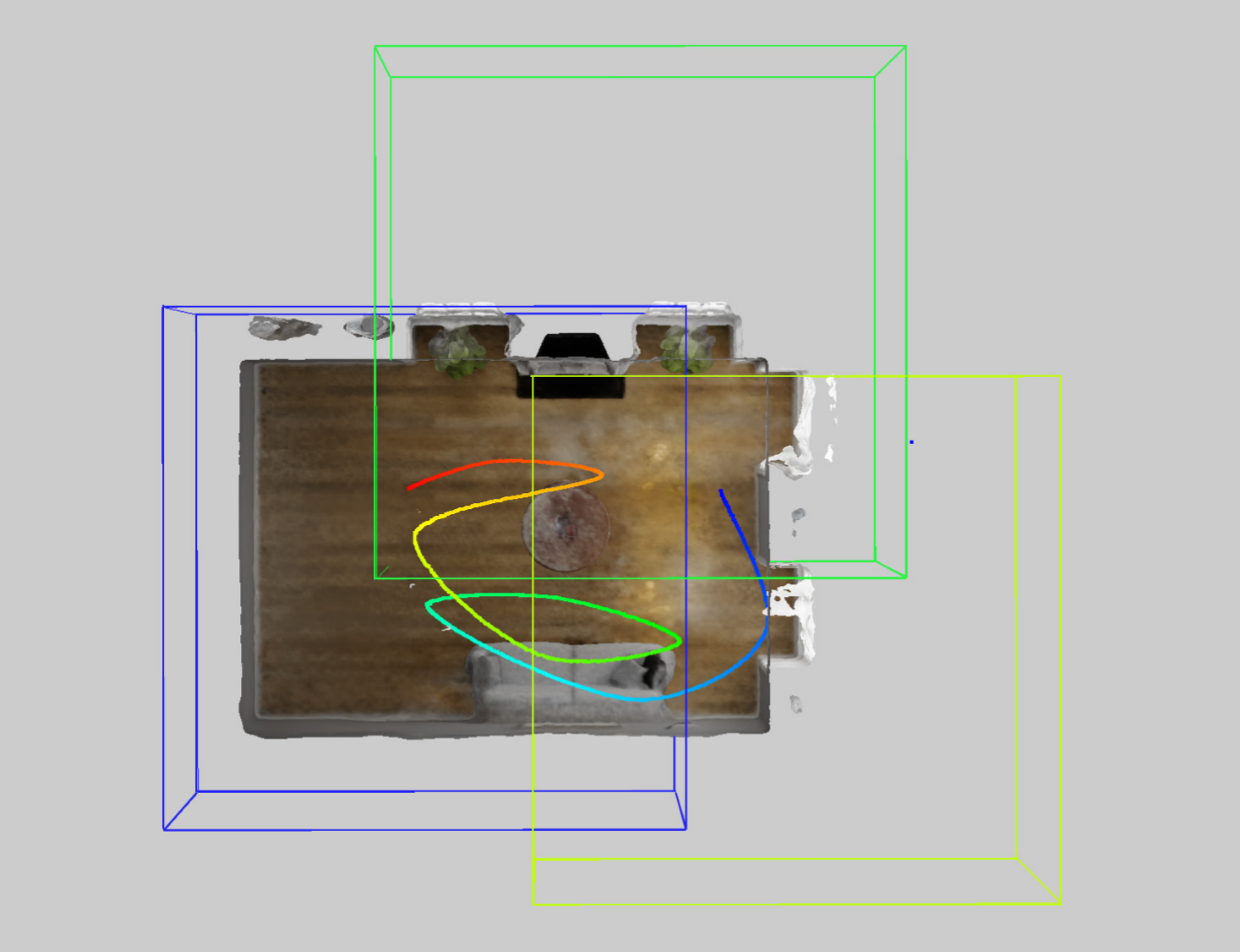}}    \\
    \vspace{1mm}
    \makecell{Replica-\texttt{office-0}}                                         &
    \makecell{Sythentic-\texttt{gwr}}                                                \\

    \makecell{\includegraphics[width=\sz\linewidth]{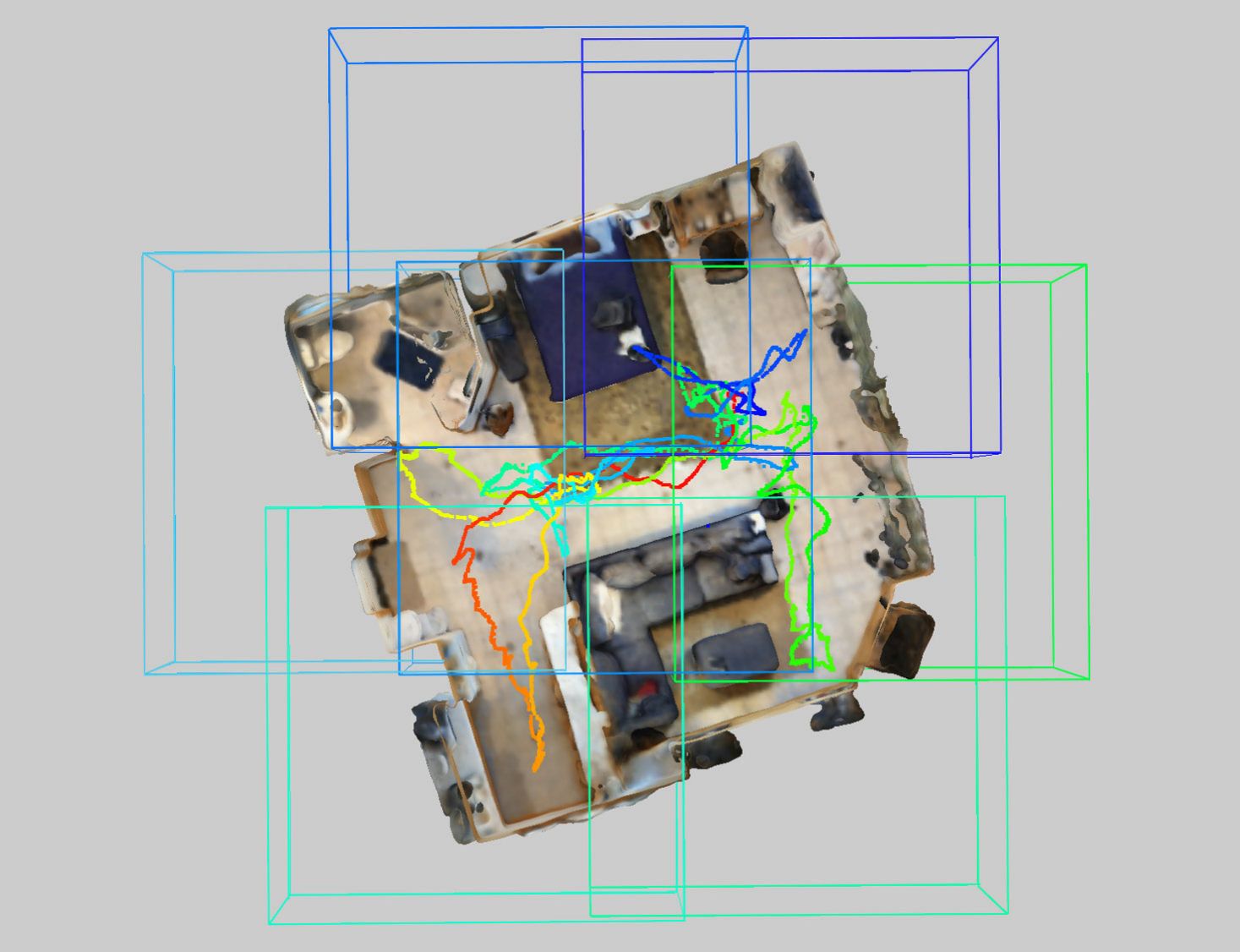}} &
    \makecell{\includegraphics[width=\sz\linewidth]{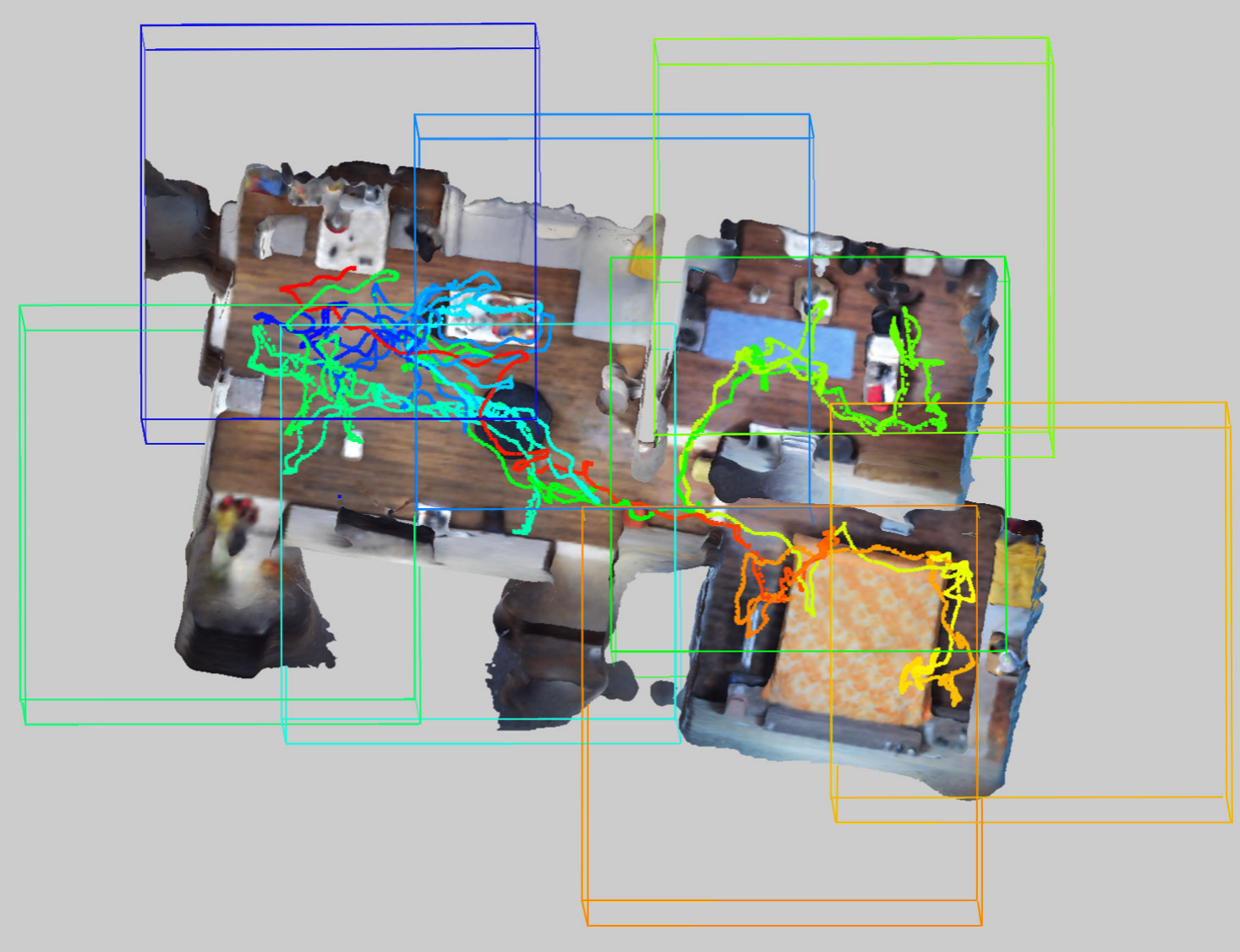}}         \\

    \makecell{Scannet-\texttt{scene0000}}                                        &
    \makecell{Apartment}                                                             \\

    \multicolumn{2}{l}{Frame ID}                                                     \\

    \multicolumn{2}{l}{\includegraphics[width=0.3\linewidth]{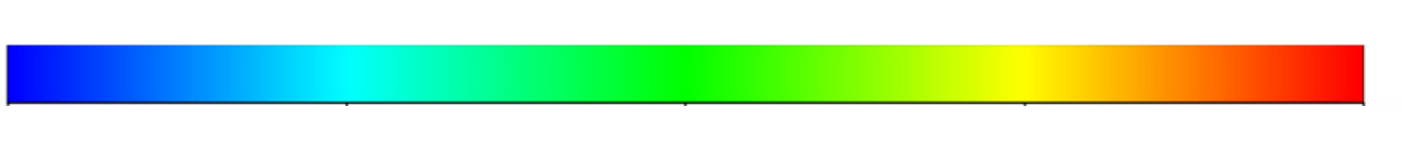}} \\
  \end{tabular}
  \vspace{-1mm}
  \caption{NeBs allocation of our method in different scenarios.}
  \label{block}
  \vspace{-5pt}
\end{figure}

\begin{figure*}
  \centering
  \scriptsize
  \setlength{\tabcolsep}{0.5pt}
  \newcommand{\sz}{0.245}  %
  \begin{tabular}{lcccc}
    \makecell{\rotatebox{90}{NICE-SLAM~\cite{2022nice}}}                               &
    \makecell{\includegraphics[width=\sz\linewidth]{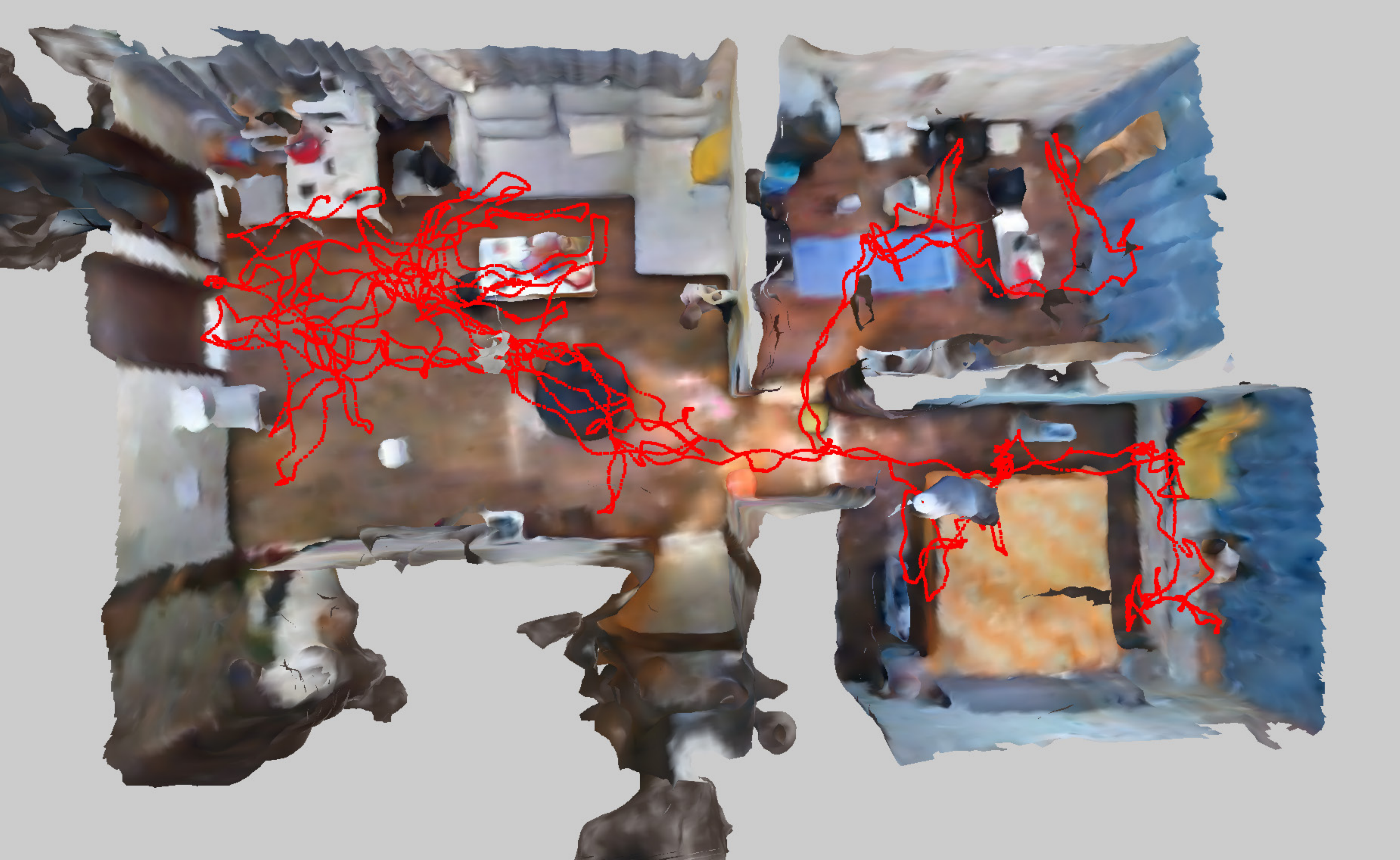}}        &
    \makecell{\includegraphics[width=\sz\linewidth]{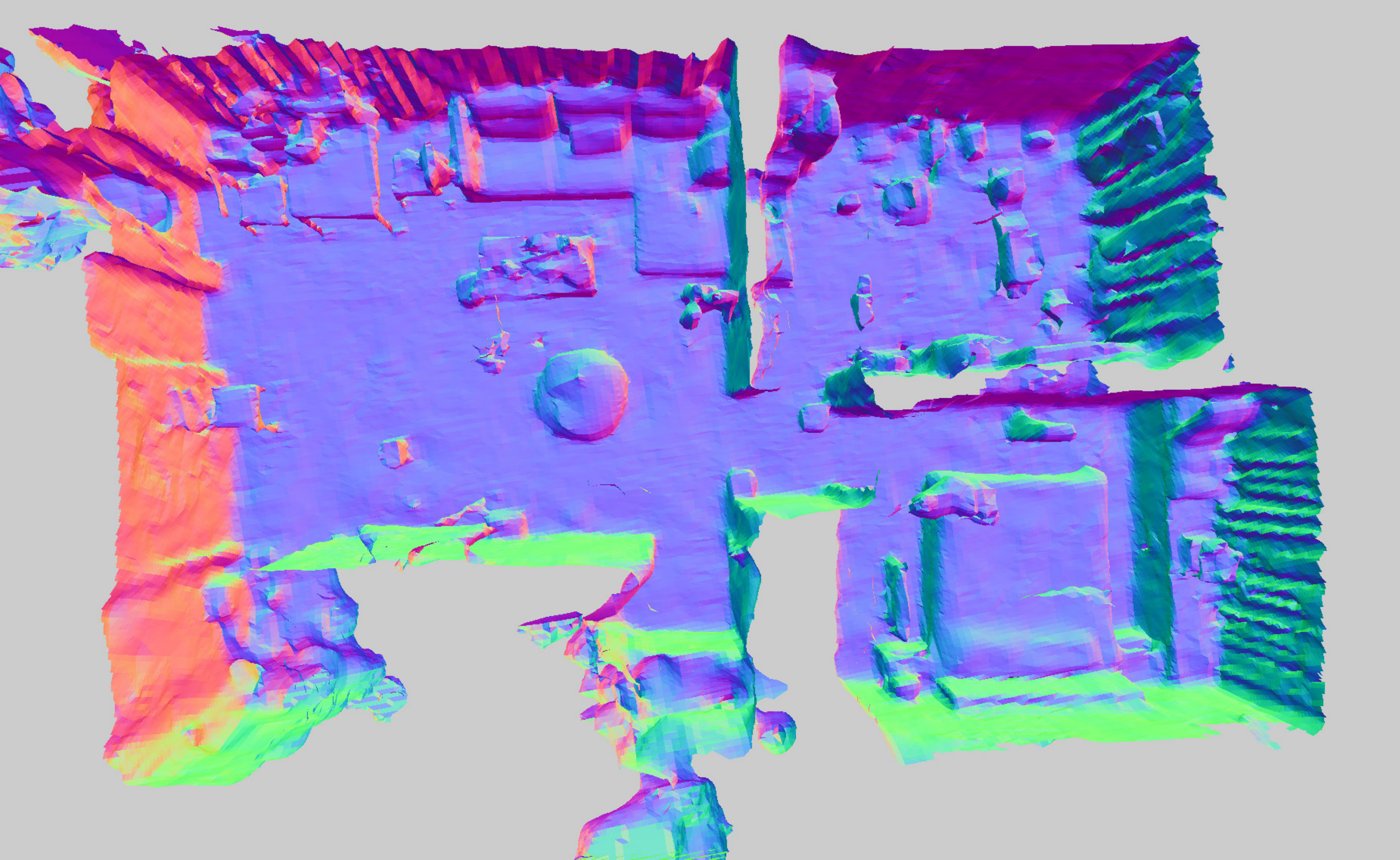}} &
    \makecell{\includegraphics[width=\sz\linewidth]{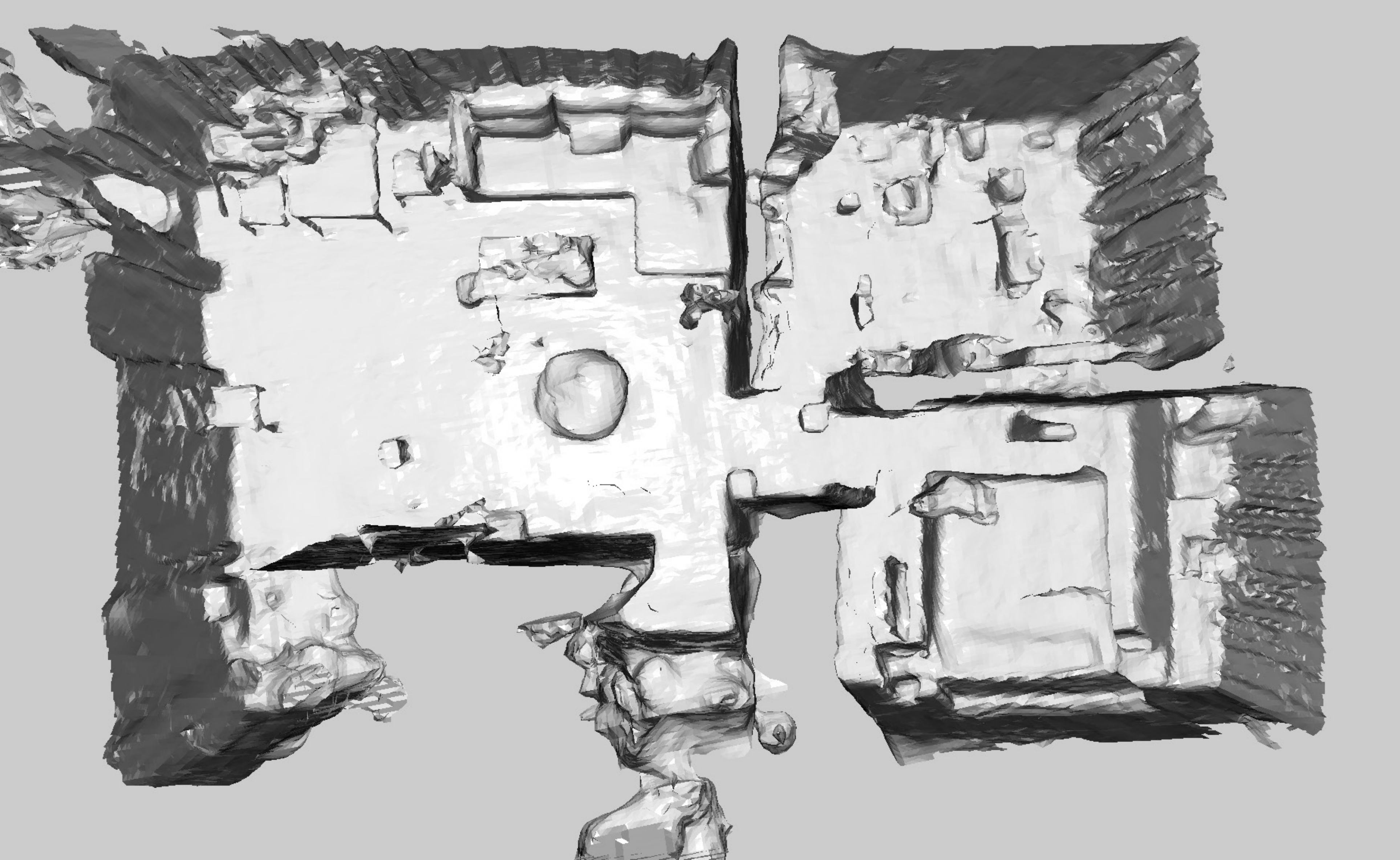}}   &
    \makecell{\includegraphics[width=\sz\linewidth]{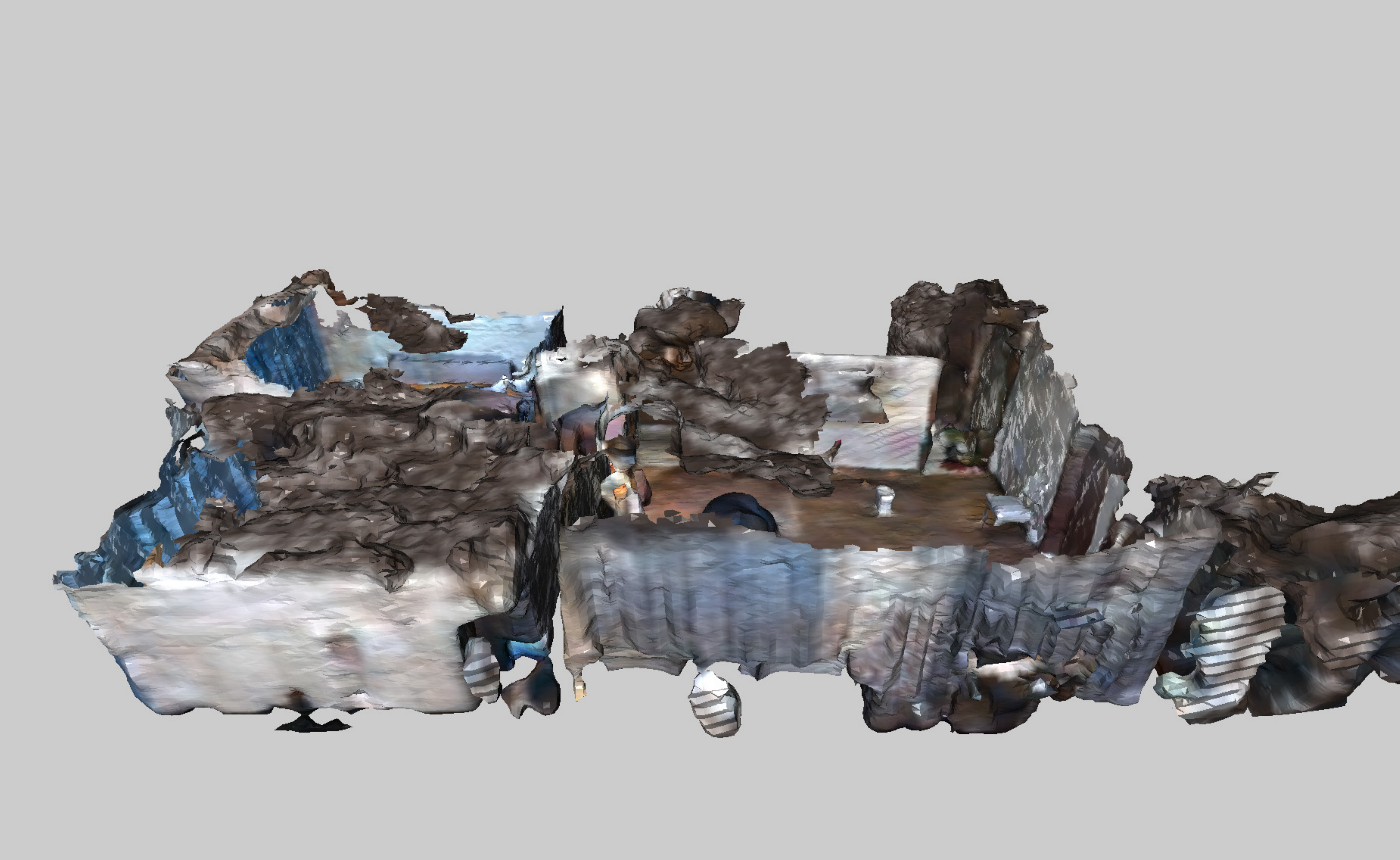}}       \\

    \makecell{\rotatebox{90}{Co-SLAM~\cite{wang2023co}}}                               &
    \makecell{\includegraphics[width=\sz\linewidth]{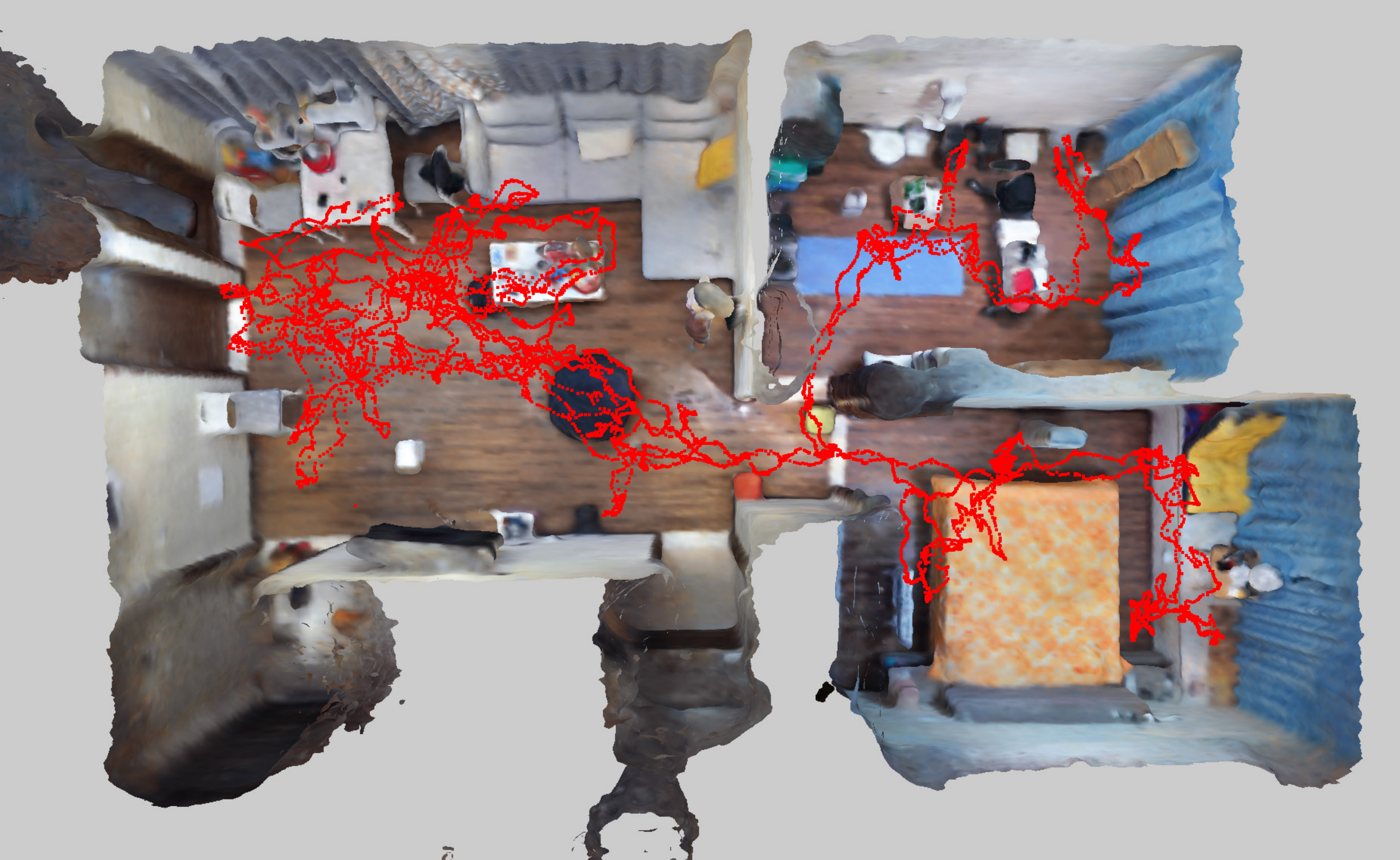}}          &
    \makecell{\includegraphics[width=\sz\linewidth]{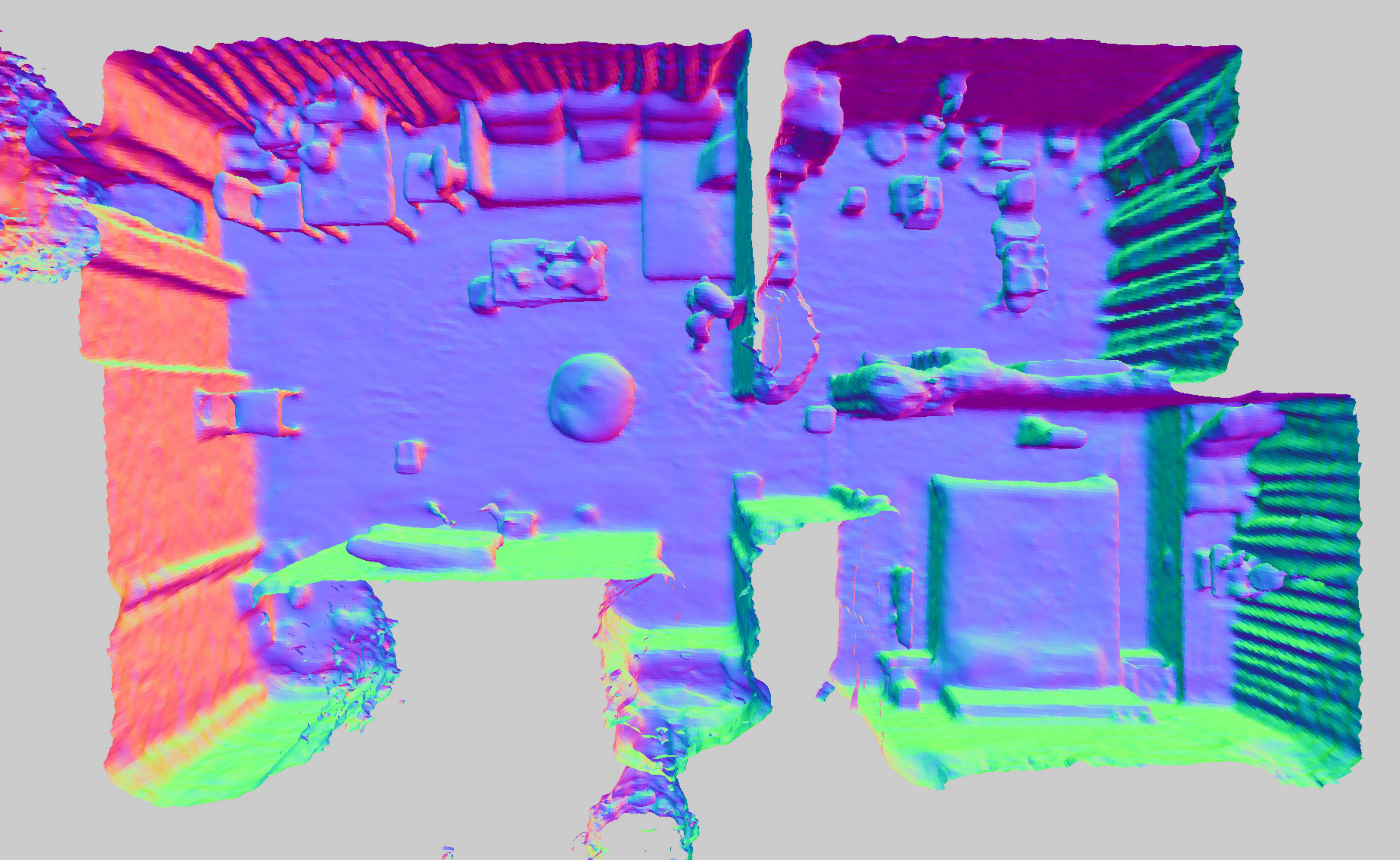}}   &
    \makecell{\includegraphics[width=\sz\linewidth]{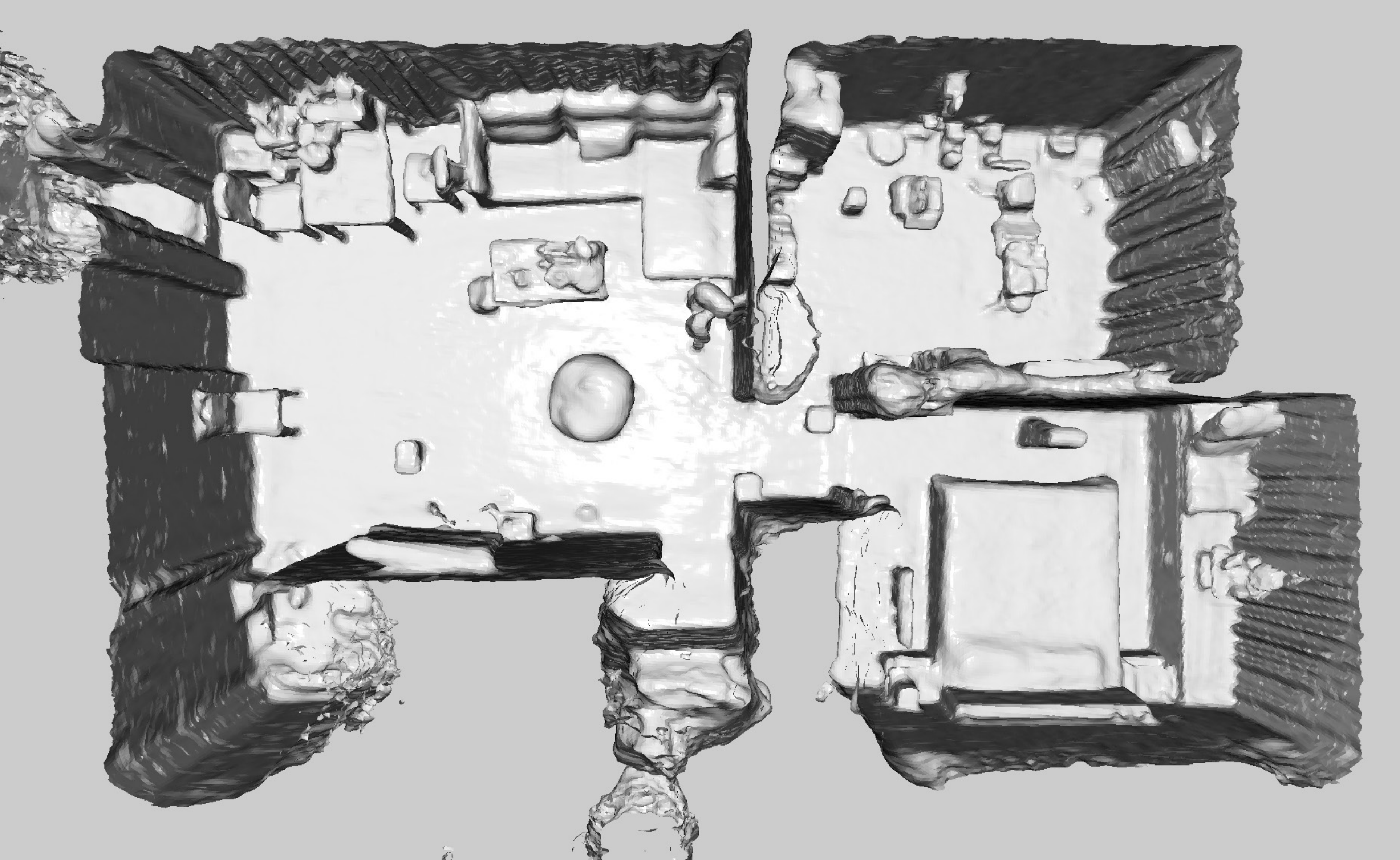}}     &
    \makecell{\includegraphics[width=\sz\linewidth]{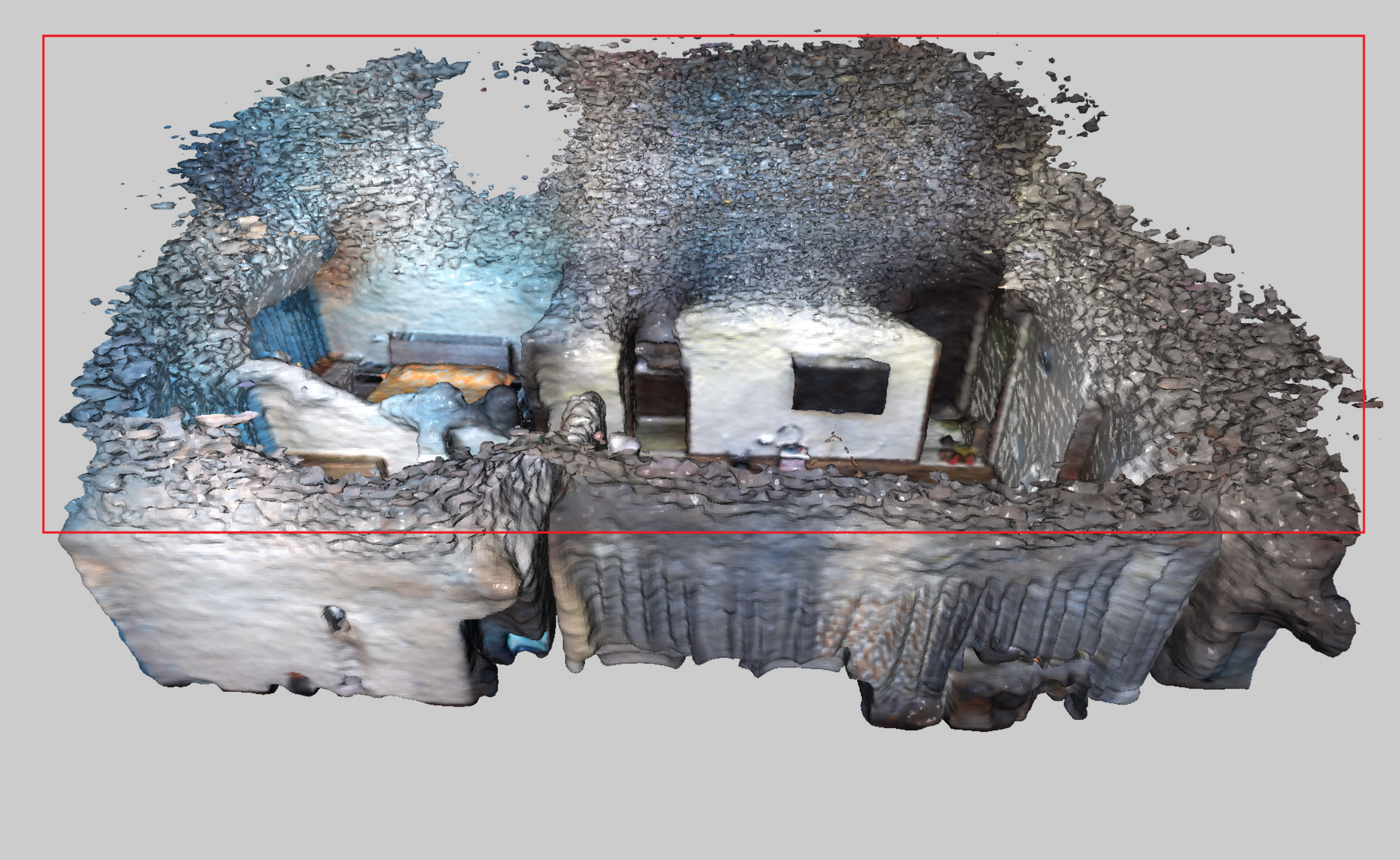}}         \\

    \makecell{\rotatebox{90}{NeB-SLAM}}                                                &
    \makecell{\includegraphics[width=\sz\linewidth]{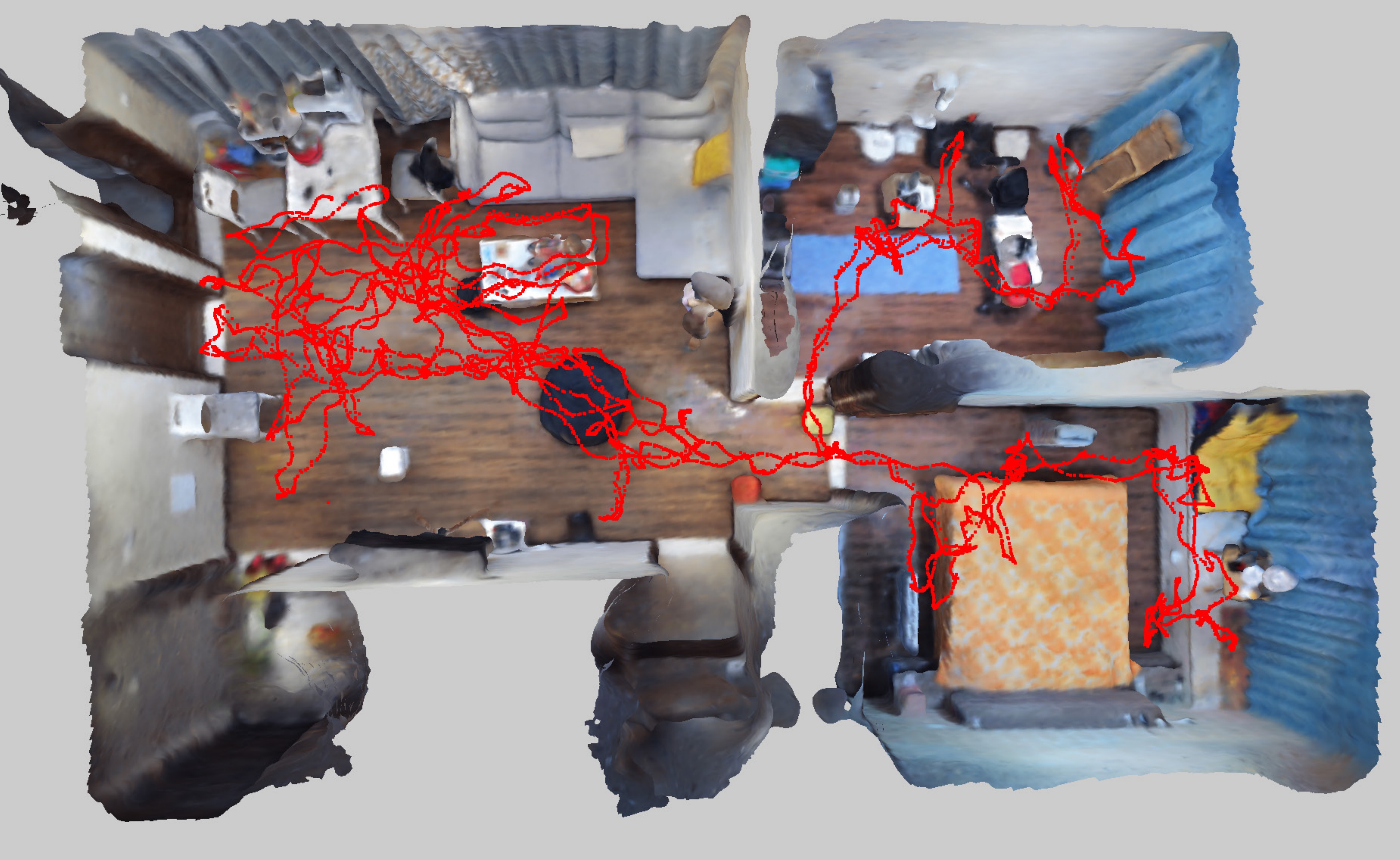}}         &
    \makecell{\includegraphics[width=\sz\linewidth]{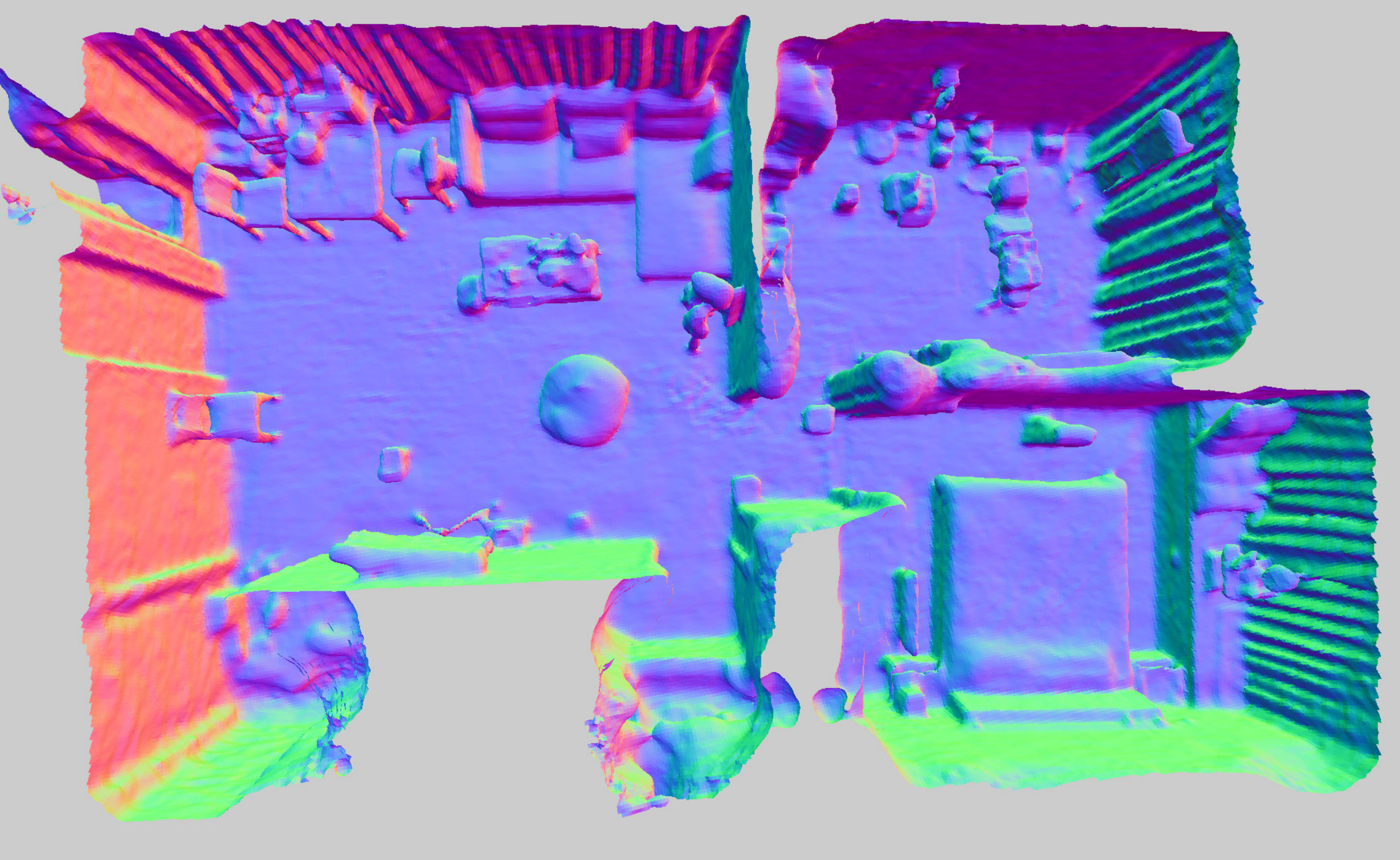}}  &
    \makecell{\includegraphics[width=\sz\linewidth]{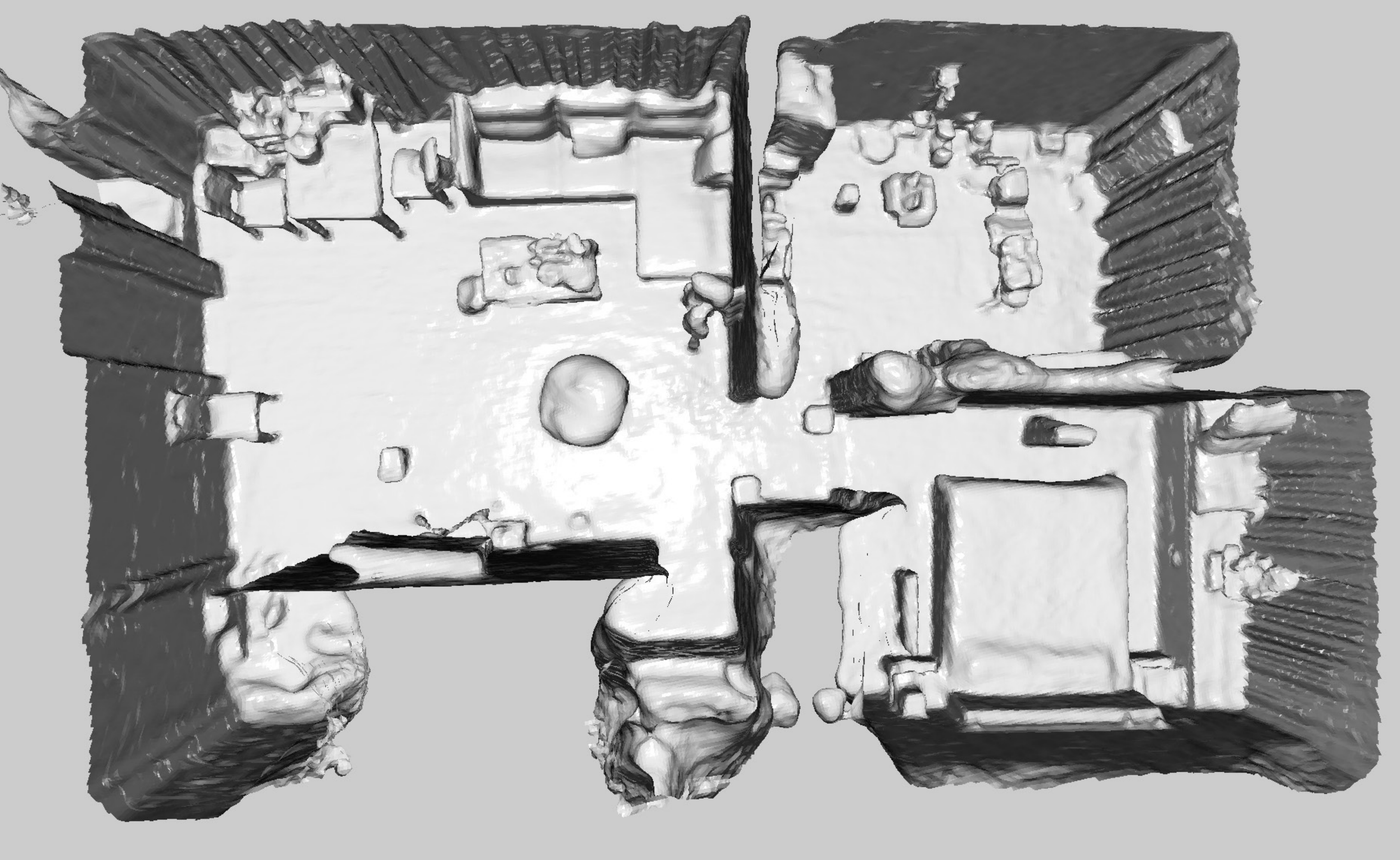}}    &
    \makecell{\includegraphics[width=\sz\linewidth]{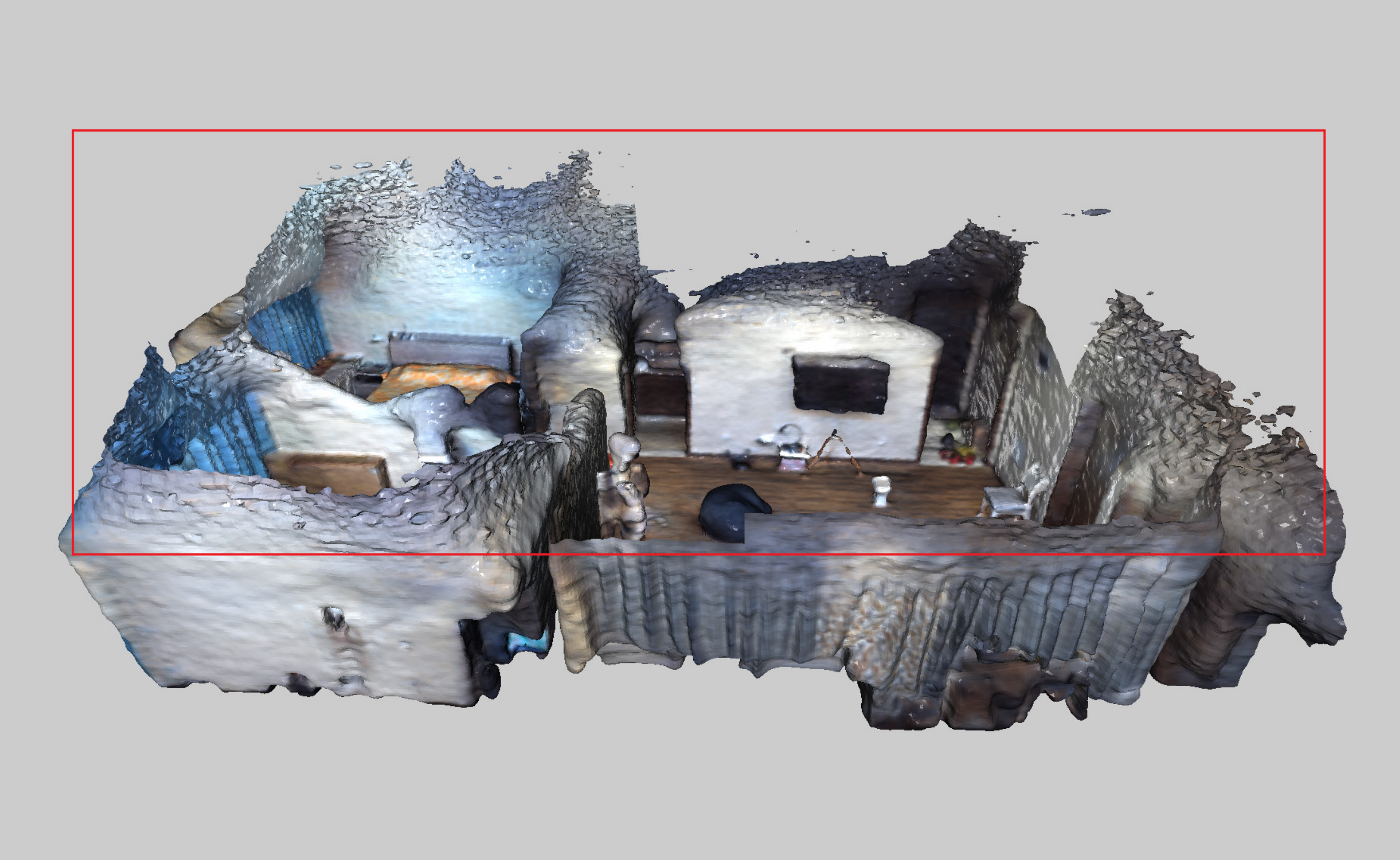}}        \\

                                                                                       &
    \makecell{(a)}                                                                     &
    \makecell{(b)}                                                                     &
    \makecell{(c)}                                                                     &
    \makecell{(d)}                                                                       \\
  \end{tabular}
  \vspace{-1mm}
  \caption{Qualitative comparison on NICE-SLAM\cite{2022nice} apartment sequence with
    different shading mode. In comparison to NICE-SLAM, our method produces more refined
    results with enhanced quality and accuracy. In contrast to Co-SLAM, our method generates
    less data for unobserved regions, as illustrated in (d). It is noteworthy
    that our method does not require an input scene size.}
  \label{render_apartment}
  \vspace{-5pt}
\end{figure*}

\subsection{Evaluation of Tracking and Mapping}
\subsubsection{Replica dataset} Tab \ref{recon_results} presents a comparative
analysis of the reconstruction accuracy of the proposed method (NeB-SLAM) with
that of the baseline method on Replica dataset\cite{straub2019replica}.
In the majority of instances, our method demonstrates superior performance.
In comparison to Co-SLAM, our method has demonstrated enhanced performance
in nearly all scenarios. It is observed that the less memory version
(NeB-SLAM$^{\dagger}$) yields better results in certain instances. It is
noteworthy that our approach does not require the size of the scene to be
inputted, and instead relies on the proposed adaptive map growth strategy
to gradually allocate NeBs to cover the entire scene. Consequently, in terms
of memory, our method is not dominant in some scenes, as shown in Tab. \ref{run_time}.
On this dataset, our method allocates 3.7 NeBs equally for a single scene.
Since each NeB represents only a localized scene, as illustrated in Fig. \ref{mapping},
our method is more expressive than Co-SLAM\cite{wang2023co}, which employs
a single hash-grid to represent the entire scene.
Furthermore, the absolute trajectory error of our method is evaluated on
this dataset, as illustrated in Tab. \ref{ate_replica}. Our method is
demonstrated to outperform other baseline methods, with the exception
of the 3D Gaussian Splatting-based SplaTAM\cite{keetha2024splatam}.

\begin{table}[]
  \centering
  \caption{ATE RMSE (cm) results on ScanNet dataset\cite{dai2017scannet}. NeB-SLAM achieves better
    or on-par performance compared to baseline method}
  \footnotesize
  \setlength{\tabcolsep}{0.38em}
  \begin{tabular}{lccccccc}
    \toprule
    Methods                     & \texttt{0000} & \texttt{0059} & \texttt{00106} & \texttt{00169} & \texttt{0181} & \texttt{0207} & Avg.      \\
    \midrule
    iMap$^*$                    & 47.71         & 45.90         & 30.33          & 58.60          & 21.04         & 25.95         & 38.26     \\
    NICE-SLAM                   & 12.29         & 12.25         & \nd{8.02}      & 20.23          & 13.41         & \fs{6.01}     & 12.04     \\
    Vox-Fusion                  & 17.61         & 35.53         & 8.85           & 20.12          & 19.44         & 7.59          & 18.19     \\
    MIPS-Fusion                 & 7.90          & \rd{10.70}    & 9.70           & 9.70           & 14.20         & 7.80          & 10.00     \\
    SplaTAM                     & 12.83         & \fs{10.10}    & 17.72          & 12.08          & \fs{11.10}    & 7.46          & 11.88     \\
    Co-SLAM                     & \rd{7.54}     & 12.17         & 8.85           & \rd{5.79}      & 12.89         & \rd{7.03}     & \rd{9.05} \\
    \textbf{NeB-SLAM}           & \fs{5.17}     & \nd{10.48}    & \fs{7.81}      & \fs{4.54}      & \fs{10.43}    & \nd{6.33}     & \fs{7.46} \\
    \textbf{NeB-SLAM}$^\dagger$ & \nd{6.34}     & 11.79         & \rd 8.82       & \nd{5.32}      & \nd{10.73}    & 7.69          & \nd{8.45} \\

    \bottomrule
  \end{tabular}
  \label{ate_scannet}
\end{table}

\begin{table}[]
  \centering
  \caption{ATE RMSE (cm) results on TUM RGBD dataset\cite{sturm2012benchmark}. Overall,
    our approach is better.}
  \footnotesize
  \begin{tabular}{lcccc}
    \toprule
    Methods                     & \texttt{fr1/desk} & \texttt{fr2/xyz} & \texttt{fr3/office} & Avg.      \\
    \midrule
    iMap$^*$                    & 4.07              & \nd{1.97}        & 5.20                & 3.75      \\
    NICE-SLAM                   & 3.01              & 2.11             & \rd{2.64}           & \rd{2.59} \\
    MIPS-Fusion                 & 3.00              & 1.40             & 4.60                & 3.00      \\
    SplaTAM                     & 3.35              & \rd 1.24         & 5.16                & 3.31      \\
    Co-SLAM                     & \rd{2.94}         & 2.07             & 3.06                & 2.69      \\
    \textbf{NeB-SLAM}           & \fs 1.78          & \fs{0.85}        & \fs{1.39}           & \fs{1.34} \\
    \textbf{NeB-SLAM}$^\dagger$ & \nd{1.85}         & \nd{0.98}        & \nd{1.42}           & \nd{1.42} \\

    \bottomrule
  \end{tabular}
  \label{ate_tum}
\end{table}

\subsubsection{Synthetic dataset}
The similar evaluations are conducted on the Synthetic \cite{azinovic2022neural}
dataset to assess the reconstruction results and absolute accuracy errors of our
method. The findings are presented in Tab. \ref{recon_results} and
Tab. \ref{ate_synthetic}. The proposed methodology demonstrates high reconstruction
accuracy with the lowest ATE, despite the absence of scene dimensions as input.

\subsubsection{ScanNet and TUM datasets}
Following \cite{2022nice} and \cite{wang2023co}, we also evaluate the tracking
accuracy of our method on ScanNet\cite{dai2017scannet} and TUM\cite{sturm2012benchmark}
datasets. The results are presented in Tab. \ref{ate_scannet} and Tab. \ref{ate_tum}.
For ScanNet dataset, 6 scenes are evaluated and compared with the baseline methods,
resulting in the highest tracking accuracy in most scenes, with an average tracking
frame rate of approximately 15.8 FPS. On this dataset, our method allocates an
average of 5 NeBs per scene, which is less advantageous in terms of the number
of parameters compared to Co-SLAM\cite{wang2023co}. However, when targeting unknown scenes, exploring
the scene using a small number of redundant parameters is unavoidable.
Fig. \ref{render_scannet} shows the qualitative comparison on ScanNet \texttt{scene0000}
with different shading mode. Our methods achieve accurate scene reconstruction
without the need for scene size. For the TUM
dataset, we evaluate 3 of the scenes with an average tracking frame rate of
approximately 17.2 FPS. On average, 1.6 NeBs are allocated to each scene, and the
number of parameters is less than that of Co-SLAM.
It is noteworthy that both datasets are based on real-world scenes. The
tracking accuracy of our method on both datasets is superior to that of SplaTAM.
Moreover, both Vox-Fusion\cite{yang2022vox} and MIPS-Fusion\cite{tang2023mips}
can be employed in situations where the specific circumstances are not yet known.
However, Vox-Fusion lacks the functionality of loop closure detection, which
impedes the correction of cumulative drift. MIPS-Fusion employs a submap overlap
approach to detect loop closure, which is only capable of correcting smaller drifts.

\begin{table}[]
  \centering
  \caption{Run-time and memory comparison on Replica, Synthetic RGBD,
    ScanNet, TUM RGBD and Apartment datasets with respective settings.
    Run-time is reported in \texttt{ms/frame / \#iter}. For our method,
    the Enc. is reported in \texttt{\#para / average number of NEBs}.}
  \footnotesize
  \setlength{\tabcolsep}{0.46em}
  \begin{tabular}{c|ccccc}
    \toprule
    \multirow{2}{*}{Datasets} & \multirow{2}{*}{Methods}    & \multirow{2}{*}{Track. (ms)$\downarrow$} & \multirow{2}{*}{Map. (ms)$\downarrow$} & \multicolumn{2}{c}{\#param. (MB)$\downarrow$}        \\
    \cmidrule(lr){5-6}
                              &                             &                                          &                                        & Enc.                                          & Dec. \\

    \midrule
    \multirow{5}{*}{\rotatebox{90}{Replica}}
                              & iMap$^*$                    & 1204.3/50                                & 10738.8/300                            & /                                             & 0.85 \\
                              & NICE-SLAM                   & 58.1/10                                  & 1986.5/60                              & 66.13                                         & 0.22 \\
                              & Co-SLAM                     & 41.8/10                                  & 74.2/10                                & 6.33                                          & 0.02 \\
                              & \textbf{NeB-SLAM}           & 48.3/10                                  & 81.4/10                                & 12.75/3.7                                     & 0.02 \\
                              & \textbf{NeB-SLAM}$^\dagger$ & 43.3/10                                  & 72.4/10                                & 6.83/3.7                                      & 0.02 \\

    \midrule
    \multirow{5}{*}{\rotatebox{90}{Synthetic}}
                              & iMap$^*$                    & 1259.6/50                                & 10698.8/300                            & /                                             & 0.85 \\
                              & NICE-SLAM                   & 57.2/10                                  & 1398.1/60                              & 7.90                                          & 0.22 \\
                              & Co-SLAM                     & 42.5/10                                  & 75.5/10                                & 6.52                                          & 0.02 \\
                              & \textbf{NeB-SLAM}           & 50.8/10                                  & 92.7/10                                & 8.74/2.5                                      & 0.02 \\
                              & \textbf{NeB-SLAM}$^\dagger$ & 44.8/10                                  & 80.7/10                                & 4.68/2.5                                      & 0.02 \\

    \midrule
    \multirow{5}{*}{\rotatebox{90}{ScanNet}}
                              & iMap$^*$                    & 1249/50                                  & 10477/300                              & /                                             & 0.85 \\
                              & NICE-SLAM                   & 391.2/50                                 & 2856.7/60                              & 38.91                                         & 0.22 \\
                              & Co-SLAM                     & 58.3/10                                  & 138.2/10                               & 3.01                                          & 0.02 \\
                              & \textbf{NeB-SLAM}           & 63.1/10                                  & 176.1/10                               & 17.00/5.0                                     & 0.02 \\
                              & \textbf{NeB-SLAM}$^\dagger$ & 59.1/10                                  & 148.1/10                               & 9.10/5.0                                      & 0.02 \\

    \midrule
    \multirow{5}{*}{\rotatebox{90}{TUM}}
                              & iMap$^*$                    & 4765.7/200                               & 10451.2/300                            & /                                             & 0.85 \\
                              & NICE-SLAM                   & 5653.3/200                               & 7279.7/60                              & 387.41                                        & 0.22 \\
                              & Co-SLAM                     & 48.7/10                                  & 262.3/20                               & 6.40                                          & 0.02 \\
                              & \textbf{NeB-SLAM}           & 58.2/10                                  & 288.6/20                               & 5.68/1.6                                      & 0.02 \\
                              & \textbf{NeB-SLAM}$^\dagger$ & 52.2/10                                  & 271.6/20                               & 3.04/1.6                                      & 0.02 \\

    \midrule
    \multirow{5}{*}{\rotatebox{90}{Apartment}}
                              & iMap$^*$                    & 1247.4/50                                & 15516.9/300                            & /                                             & 0.85 \\
                              & NICE-SLAM                   & 268.7/50                                 & 2657.5/60                              & 119.09                                        & 0.22 \\
                              & Co-SLAM                     & 45.3/10                                  & 118.1/10                               & 41.85                                         & 0.02 \\
                              & \textbf{NeB-SLAM}           & 53.8/10                                  & 132.4/10                               & 27.2/8.0                                      & 0.02 \\
                              & \textbf{NeB-SLAM}$^\dagger$ & 50.2/10                                  & 127.8/10                               & 6.19/8.0                                      & 0.02 \\

    \bottomrule
  \end{tabular}
  \label{run_time}
\end{table}

\subsubsection{Apartment dataset}
We evaluate our method on an apartment dataset collected by \cite{2022nice}.
The dataset consists of 12595 images with a scene larger than those in the previous
datasets. As illustrated in Fig. \ref{render_apartment}, Our method produces more refined
results with enhanced quality and accuracy compared to NICE-SLAM\cite{2022nice}. In
contrast to Co-SLAM\cite{wang2023co}, our method generates less data for unobserved
regions, as illustrated in Fig. \ref{render_apartment} (d). It is noteworthy that our
method does not require an input scene size. Furthermore, our method demonstrates a
notable superiority with respect to the number of parameters as illustrated in Tab.
\ref{run_time}. For scenes of greater complexity with a larger size, the number of
parameters of our method increases in a manner that is nearly linear, while that of
the baseline methods approach a geometric increase. Fig. \ref{block} depicts the
allocation of NeBs across various scenarios. Our method allocates NeBs sequentially
along the trajectory and progressively covers the entire unknown scene.

\subsection{Performance Analysis}
On a desktop PC with an Intel Core i9-14900KF CPU and NVIDIA RTX 4090 GPU, our
method (NeB-SLAM) achieves a tracking frame rate of 20 fps when utilising the default
settings. For datasets that present greater challenges, such as those from Scannet
and TUM, 15 FPS can still be achieved as shown in Tab. \ref{run_time}.
In comparison to Co-SLAM, our method does not exhibit superior processing efficiency
or a smaller number of parameters. Nevertheless, we are capable of constructing a
comprehensive map of uncharted environments with a minimal increase in parameters,
while ensuring high-precision tracking. This is a capability that is not available
with all baseline methods. Furthermore, the computational complexity of
our method remains relatively constant as the number of NeBs increases, a consequence
of our approach, which considers only the NeBs within the current view frustum.

\begin{table}[]
  \centering
  \caption{Ablation study on loop closure on Replica dataset.}
  \footnotesize
  \begin{tabular}{c|ccccc}
    \toprule
    loop closure & ATE$\downarrow$ & Depth L1$\downarrow$ & Acc.$\downarrow$ & Comp.$\downarrow$ & Comp. Ratio$\uparrow$ \\
    \midrule
    \XSolidBrush & 0.84            & 1.43                 & 1.97             & 2.07              & 93.61                 \\
    \Checkmark   & \bf{0.59}       & \bf{1.38}            & \bf{1.92}        & \bf{1.94}         & \bf{93.85}            \\
    \bottomrule
  \end{tabular}
  \label{loop}
\end{table}

\subsection{Ablation Study}
We evaluate two sizes of hash tables, as shown in Tab. \ref{run_time},
NeB-SLAM$^\dagger$ ($T$=14) is more advantageous than NeB-SLAM ($T$=15) in terms
of processing efficiency and the number of parameters in each dataset. However,
NeB-SLAM yields superior results in terms of reconstruction accuracy and pose
estimation. A larger hash table size was not tested since the number of parameters
would be significantly higher.

Furthermore, the impact of loop closure detection on tracking accuracy and
reconstruction quality is evaluated on Replica dataset\cite{straub2019replica},
as illustrated in Tab. \ref{loop}. The implementation of global pose
correction through the use of loop closure detection has been demonstrated
to enhance the precision of camera tracking. The attainment of high-accuracy
camera poses has been shown to facilitate greater global consistency in
the mapping process, which in turn leads to an improvement in the quality
of the reconstructed map.

\section{Conclusion}
The proposed NeB-SLAM is designed to address the challenge of constructing
dense maps for unknown scenes. Our approach involves a divide-and-conquer
strategy, whereby the unknown scene is divided into multiple NeBs of fixed
size. These NeBs are adaptively allocated during camera tracking, gradually
covering the entire unknown scene. The BoW model is also employed for global
loop closure detection with the objective of correcting the cumulative error.
This results in enhanced camera tracking accuracy and global map consistency.
Furthermore, when confronted with larger scenes, our method ensures the
linear growth of model parameters, rather than geometric growth, while
maintaining the scene representation capability of the model.

\textbf{Limitations.}
At present, our method is only capable of adaptively assigning fixed-size
NeBs. In our future work, we intend to pursue the adaptation of NeB sizes
with the objective of achieving a more efficient representation of 3D scenes.

\bibliographystyle{IEEEtran}
\bibliography{egbib}



\end{document}